\RequirePackage{fix-cm}
\documentclass[twocolumn]{svjour}         
\smartqed  						
\usepackage{graphicx}
\usepackage{mathptmx}
\usepackage{amsmath,amsfonts}
\usepackage{color}
\usepackage{array}

\usepackage{setspace}

\usepackage[margin=2cm]{geometry}
\usepackage{float}
\usepackage{subcaption}
\usepackage{bbm}
\usepackage{mathrsfs} 
\usepackage{hyperref}
\usepackage[round]{natbib}
\usepackage[labelfont=bf]{caption}
\usepackage[ruled]{algorithm2e}
\usepackage{algorithmic}
\usepackage[labelsep=space]{caption}
\usepackage{dirtytalk}
\usepackage{graphicx}

\newcommand{\R}{\mathbb{R}}
\newcommand{\N}{\mathbb{N}}

\newcommand{\eps}{\varepsilon}

\setcitestyle{aysep={}}

\begin{document}
\graphicspath{{figures/}}

\title{Segmentation of Scanning Tunneling Microscopy Images Using Variational Methods and Empirical Wavelets}

\author{Kevin Bui \and Jacob Fauman \and David Kes \and Leticia Torres Mandiola \and Adina Ciomaga \and Ricardo Salazar \and  Andrea L. Bertozzi \and J\'er\^ome Gilles \and Andrew I. Guttentag \and Paul S. Weiss}

\institute{K. Bui \at
	      Department of Industrial Engineering and Management Sciences, Northwestern University,
	      2145 Sheridan Road, Evanston, IL 60208, United States.
              \email{kbui1993@gmail.com}
          \and J. Fauman \at
	      Department of Applied Mathematics, 526 UCB, ECOT 225,
	      University of Colorado Boulder,
	      Boulder, Colorado 80309-0526, United States.
	      \email{jacob.fauman@colorado.edu}
	  \and D. Kes \at 
	      College of Arts and Sciences, University of San Francisco, USF Downtown Campus, 
	      101 Howard Street, San Francisco, CA 94105, United States.
	      \email{ddkes@dons.usfca.edu}
	  \and L. Torres Mandiola \at 
	      Department of Mathematics, South Kensington Campus, Imperial College London, 
	      London, United Kingdom SW7 2AZ.
	      \email{leticia.torres-mandiola16@imperial.co.uk}
	  \and A. Ciomaga \at
	      Univ. Paris Diderot, Sorbonne Paris Cit\'e, Laboratoire Jacques-Louis Lions, 
	      UMR 7598, UPMC, CNRS, F-75205 Paris, France.
	      \email{adina@ljll.univ-paris-diderot.fr}
	  \and R. Salazar \at
	      Department of Mathematics, University of California, Los Angeles, 520 Portola Plaza, 
	      Los Angeles, CA 90095-1555, United States.
	      \email{rsalazar@math.ucla.edu}
	  \and A.L. Bertozzi \at
	      Department of Mathematics, University of California, Los Angeles, 520 Portola Plaza, 
	      Los Angeles, CA 90095-1555, United States.
	      \email{bertozzi@math.ucla.edu}
	  \and J. Gilles \at
	      Department of Mathematics \& Statistics, San Diego State University, 5500 Campanile Dr, 
	      San Diego, CA 92181-7720, United States.
	      \email{jgilles@mail.sdsu.edu}
	  \and A.I. Guttentag \at
	      {Department of Chemistry and Biochemistry, University of California, Los Angeles, 
	      Los Angeles, California 90095, United States. \and
    	      California NanoSystems Institute, University of California, Los Angeles, 
	      Los Angeles, California 90095, United States.
	      \email{aguttentag@ucla.edu}}
	  \and P.S. Weiss \at
	      {Department of Chemistry and Biochemistry, University of California, Los Angeles, 
	      Los Angeles, California 90095, United States. \and
	      Department of Materials Science and Engineering, University of California, Los Angeles, 
	      Los Angeles, California 90095, United States. \and
	      California NanoSystems Institute, University of California, Los Angeles, 
	      Los Angeles, California 90095, United States.
	      \email{psw@cnsi.ucla.edu}}
}

\maketitle

\begin{abstract}
In the fields of nanoscience and nanotechnology, it is important to be able to functionalize surfaces chemically for a wide variety of applications. 
Scanning 
tunneling microscopes (STMs) are important instruments in this area used to measure the surface structure and chemistry with better than molecular resolution. Self-assembly is frequently used to create monolayers that redefine the surface chemistry in just a single-molecule-thick layer \citep{nuzzo1983adsorption,smith2004patterning,love2005self}. Indeed, STM images reveal rich information 
about the structure of self-assembled monolayers since they convey chemical and physical properties of the studied material.

In order to assist in and to enhance the analysis of STM and other images \citep{thomas2015defect,thomas2016mapping}, we propose and demonstrate an image-processing framework that produces two image segmentations: one is based on
intensities (apparent heights in STM images)
and the other is based on textural patterns. The proposed framework begins with a cartoon+texture decomposition, which separates an image into its cartoon and texture components. 
Afterward, 
the cartoon image is segmented by a modified multiphase version of the local Chan-Vese model \citep{wang2010efficient}, while the texture image is segmented by a combination of 
2D empirical wavelet transform and a clustering algorithm. Overall, our proposed framework contains several new features, specifically in 
presenting a new application of cartoon+texture decomposition and of the empirical wavelet transforms and in developing a specialized framework to segment STM images and other data. To demonstrate the 
potential of our 
approach, we apply it to actual STM images of cyanide monolayers on Au\{111\} and present their corresponding segmentation results.   
\end{abstract}

\section{Introduction}\label{sec:intro}
\begin{figure}
\centering
\includegraphics[width=0.45\textwidth]{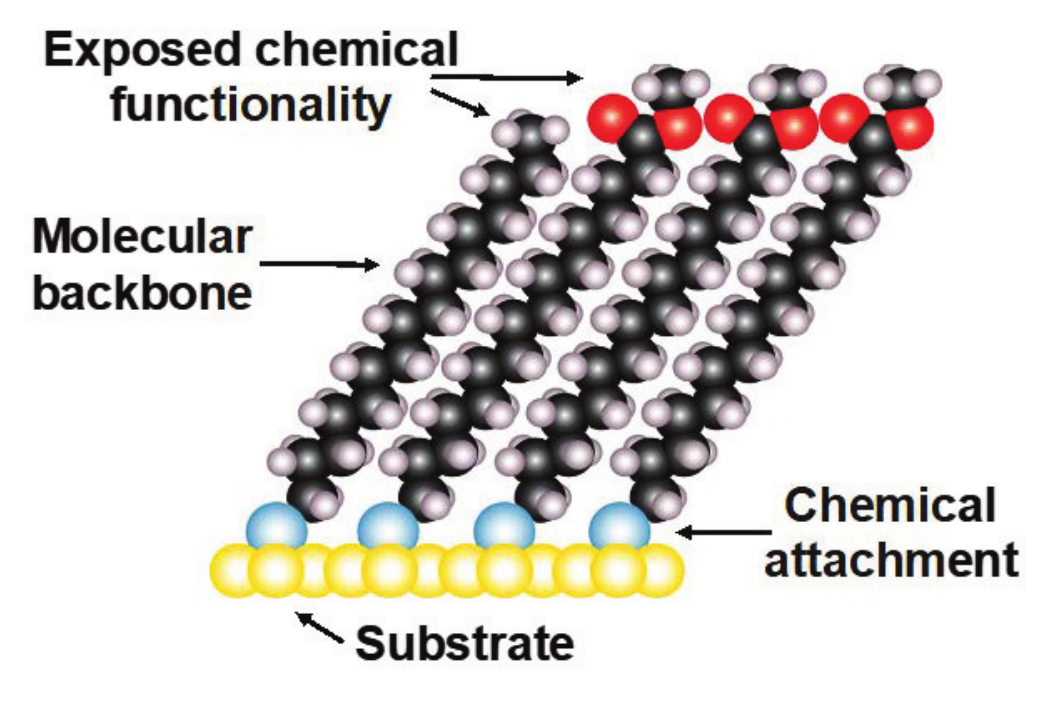}
\caption{In self-assembled monolayers, a single layer of molecules is chemically bound to a solid or liquid substrate. The wide range of substrates (\textit{e.g.}, metals, semiconductors, insulators, glasses, superconductors, nanoparticles) that can be used call for complementary chemistries of attachment of the molecular layers. The exposed functional group at the ends of the molecules typically dominates the interactions of the substrate with the surrounding chemical, physical, and biological environment}
\label{fig:SAM}
\end{figure}

Self-assembled monolayers (SAMs) have been extensively studied and applied in nanoscience, nanotechnology, and beyond \citep{poirier1997characterization, gooding2003self,smith2004patterning, love2005self}. These SAMs are formed by mole\-cules that have a head 
group 
(\textit{e.g.}, sulfur, selenium, carboxylate, phosphonate) that is chemically bound to a substrate (\textit{e.g.}, gold, silver, copper, platinum, germanium), and often form two-dimensional crystalline lattices (see Figure~\ref{fig:SAM}). The strong affinity between the head 
group and the substrate and the intermolecular interactions between the molecular backbones and tail groups lead to self-assembly of the monolayers via exposure of the substrate to the mole\-cules in solution, in vapor, or by contact with a supporting structure such as a polymer stamp. The organization of the monolayer structure depends on the chemical properties and structures of the molecular monolayer components \citep{claridge2013bottom}. Controlling the design of a SAM requires properly tuning the chemical and physical properties of the assembled mole\-cules 
\citep{CHEM}. As a result, control of basic parameters and external stimuli (\textit{e.g.}, deposition conditions, temperature, electrochemical potential, and illumination) on SAMs  has been examined to target specific assemblies for nanotechnology research and applications \citep{guttentag2016hexagons}. One method of analyzing the chemical and physical 
properties 
of SAM is by examining molecular resolution images obtained by scanning tunneling microscopy. 

\begin{figure*}
\begin{center}
\includegraphics[width=0.65\textwidth]{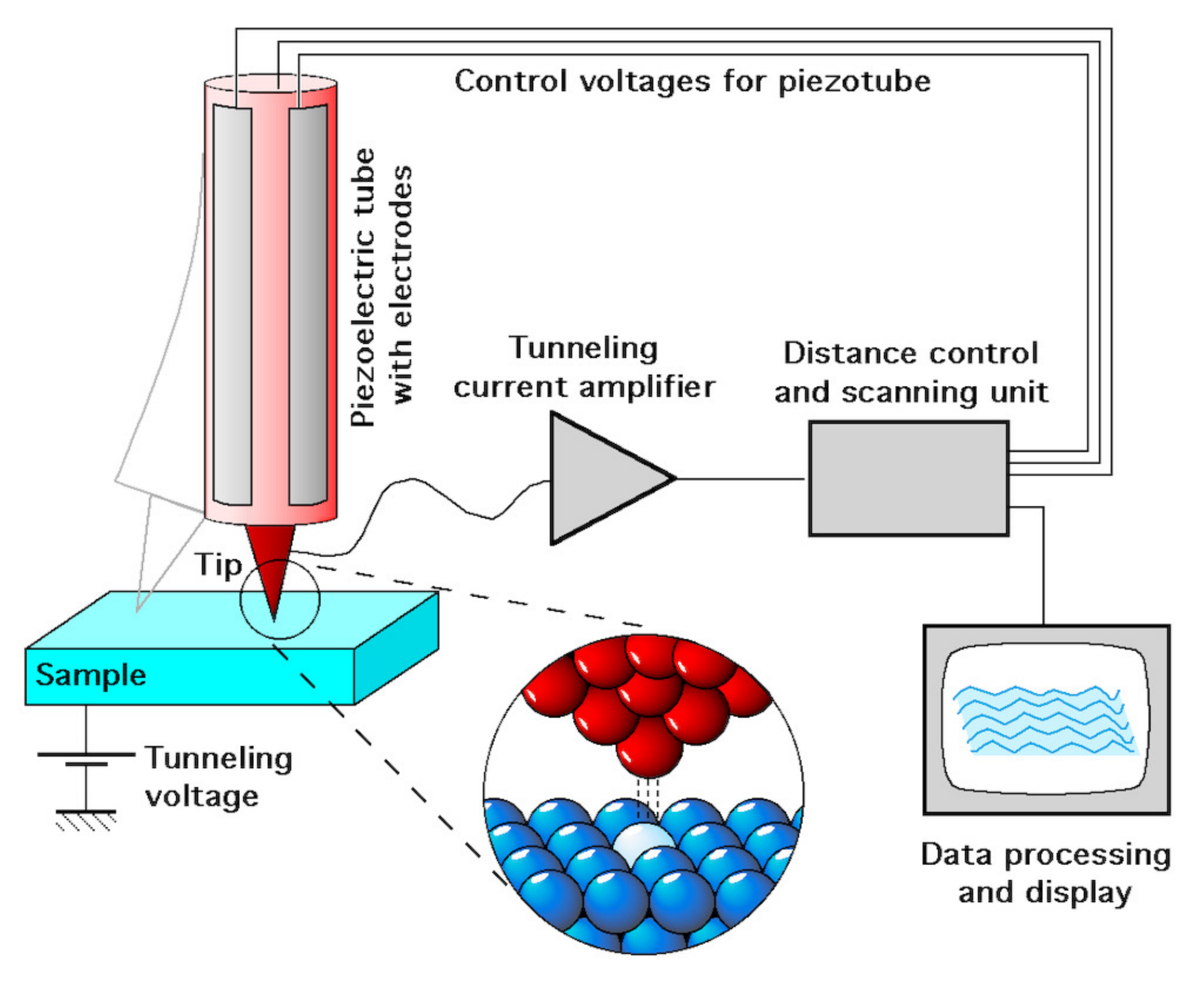}
\end{center}
   \caption{To obtain scanning tunneling microscope images, constant current is held as the tip moves across the surface, experiencing voltage drops over bumps. Figure provided by Michael Schmid, TU Wien, at \url{https://commons.wikimedia.org/w/index.php?curid=180388}}
\label{fig:STMprocess}
\end{figure*}

\begin{figure*}[ht]
\centering
 \begin{subfigure}[t]{0.25\textwidth}
  \centering
  \includegraphics[scale=0.35]{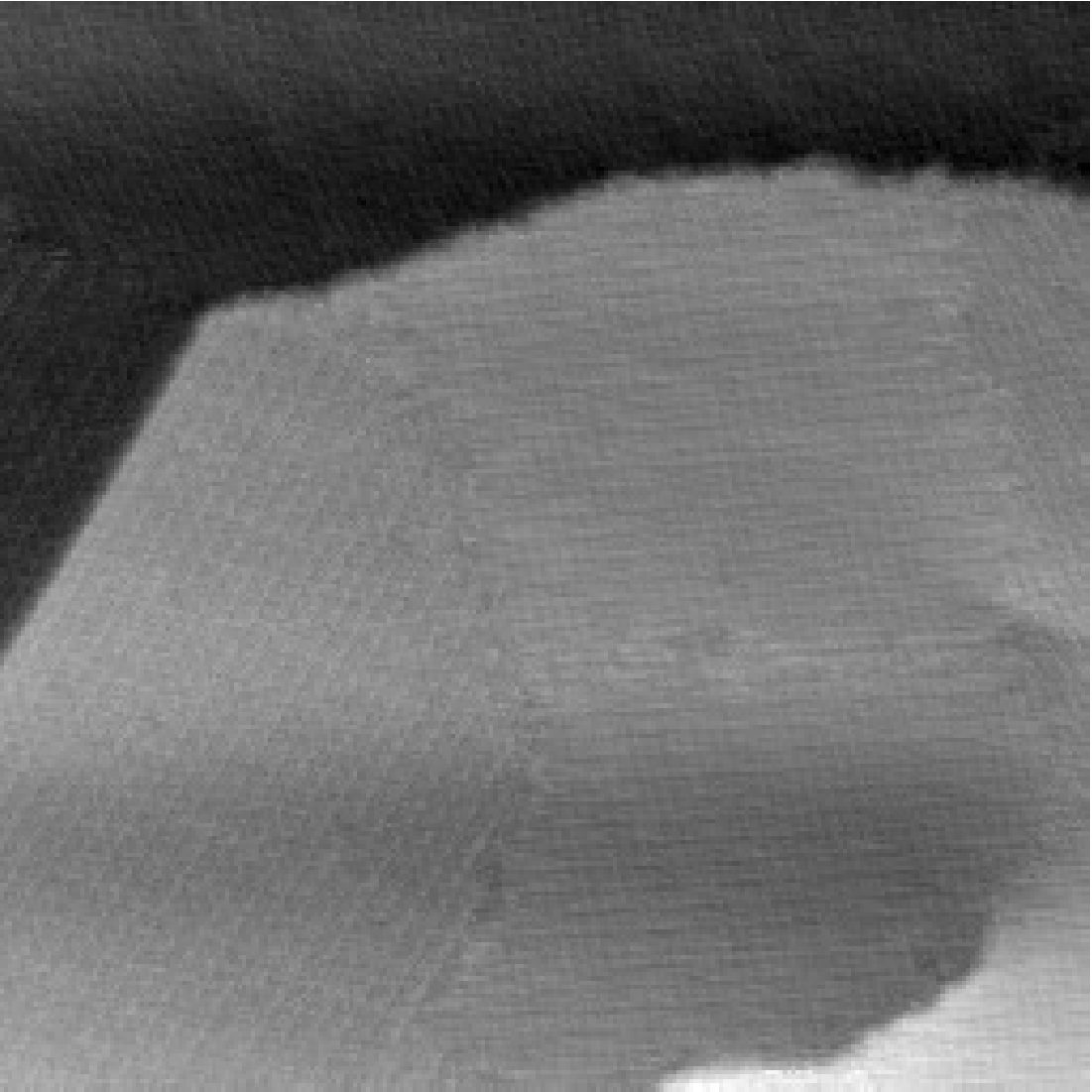}
  \end{subfigure}%
 \begin{subfigure}[t]{0.25\textwidth}
  \centering
  \includegraphics[scale=0.35]{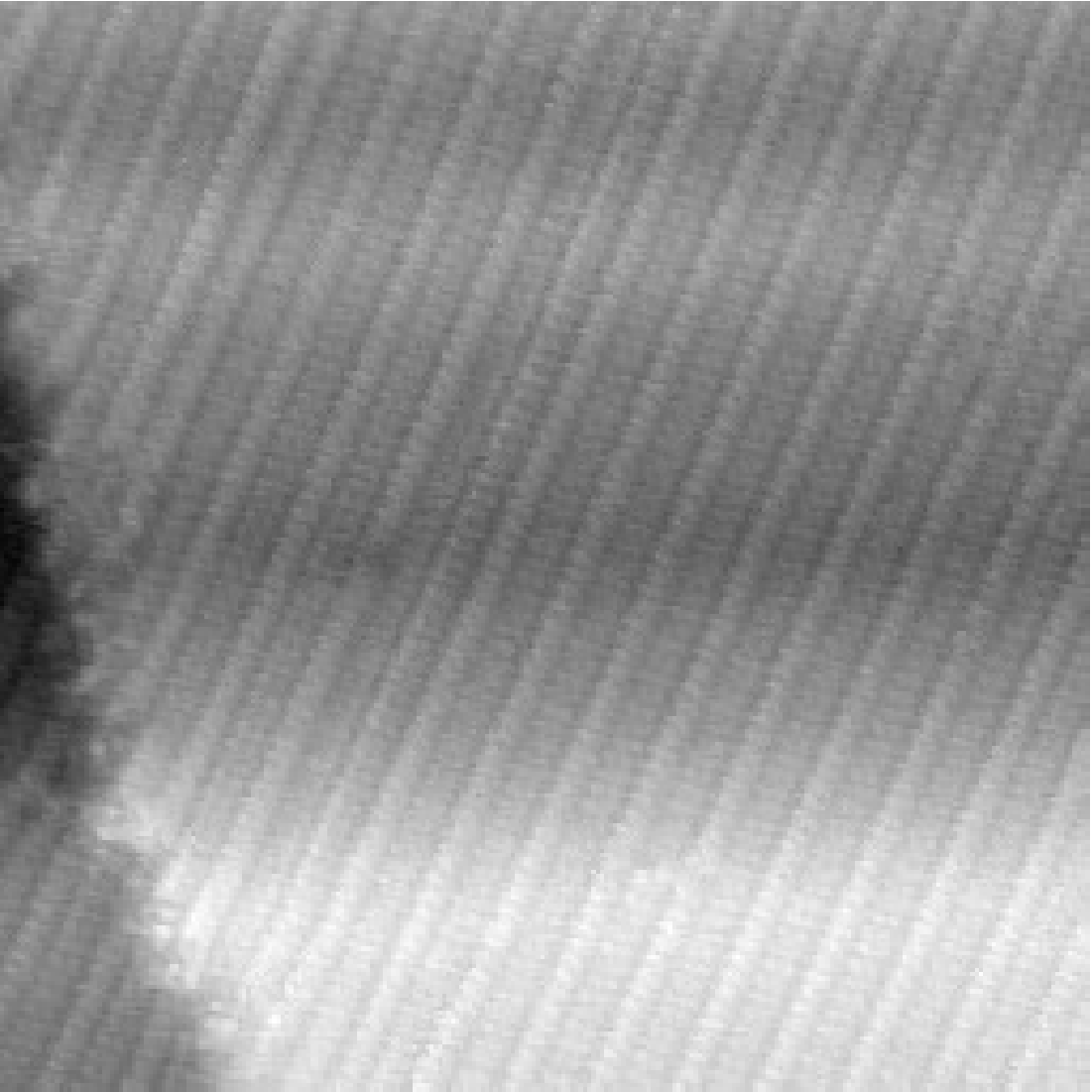}
 \end{subfigure}%
 \begin{subfigure}[t]{0.25\textwidth}
  \centering
  \includegraphics[scale=0.35]{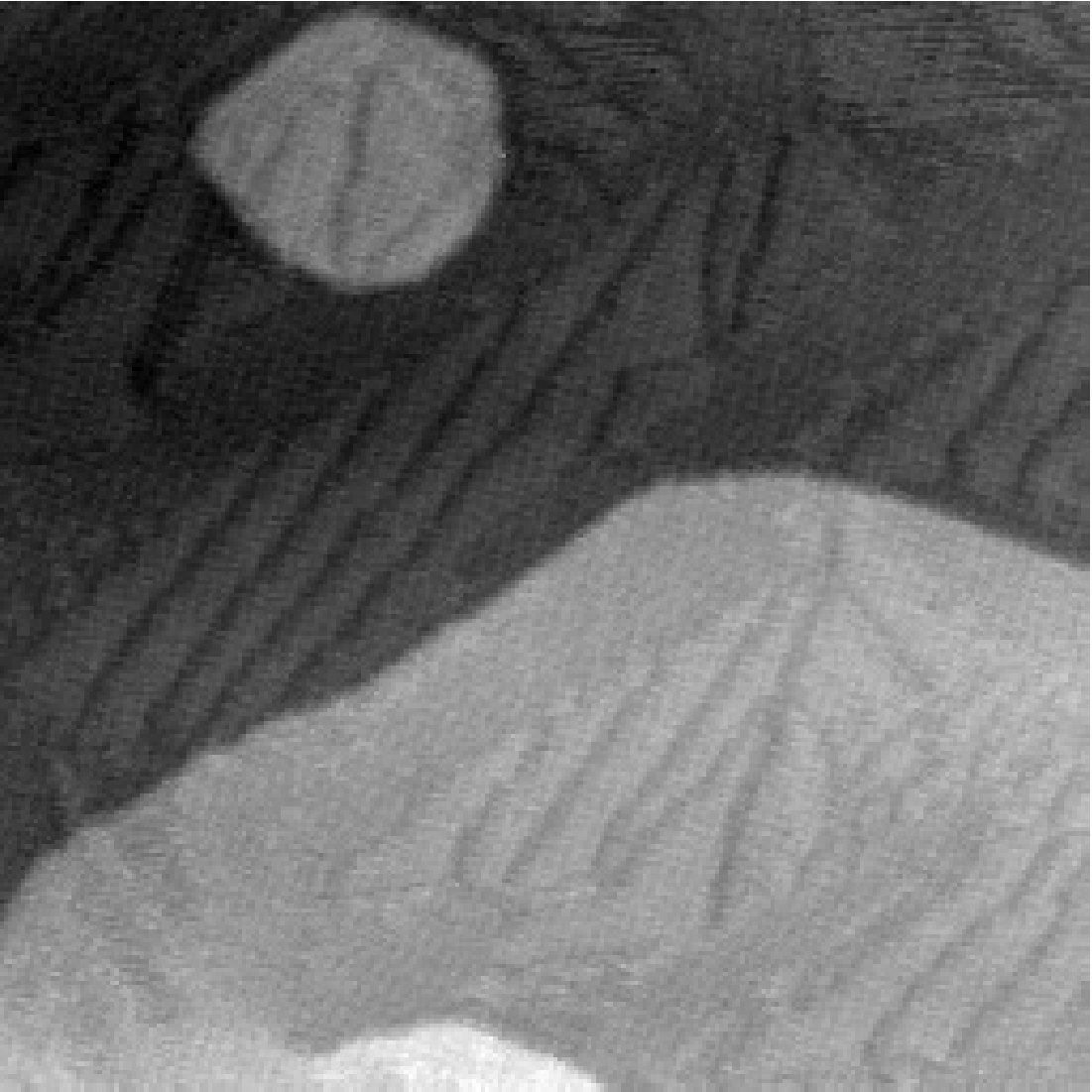}
 \end{subfigure}%
 \begin{subfigure}[t]{0.25\textwidth}
  \centering
  \includegraphics[scale=0.35]{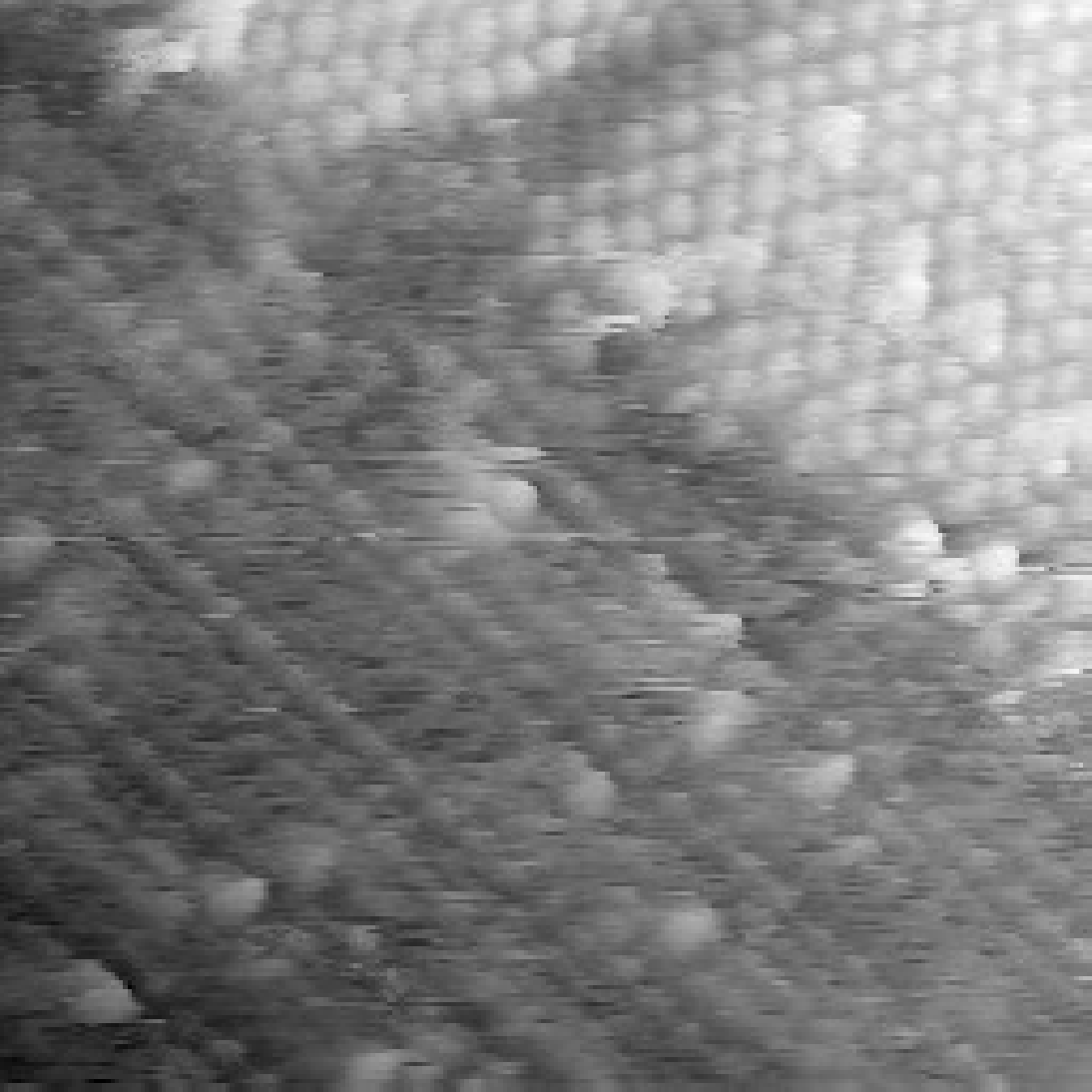}
 \end{subfigure}
 \vskip 1mm

 \begin{subfigure}[t]{0.25\textwidth}
  \centering
  \includegraphics[scale=0.35]{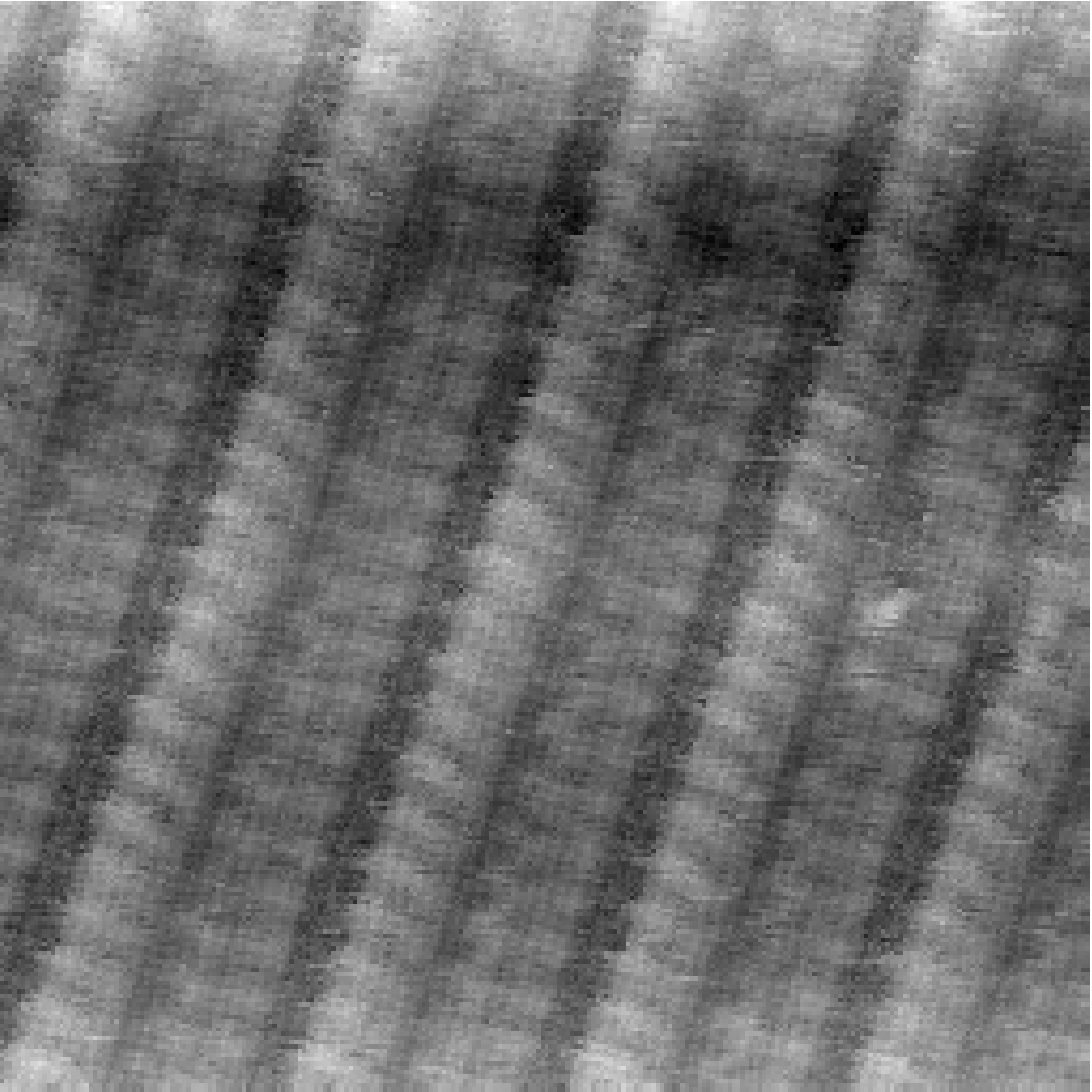}
 \end{subfigure}%
  \begin{subfigure}[t]{0.25\textwidth}
  \centering
  \includegraphics[scale=0.35]{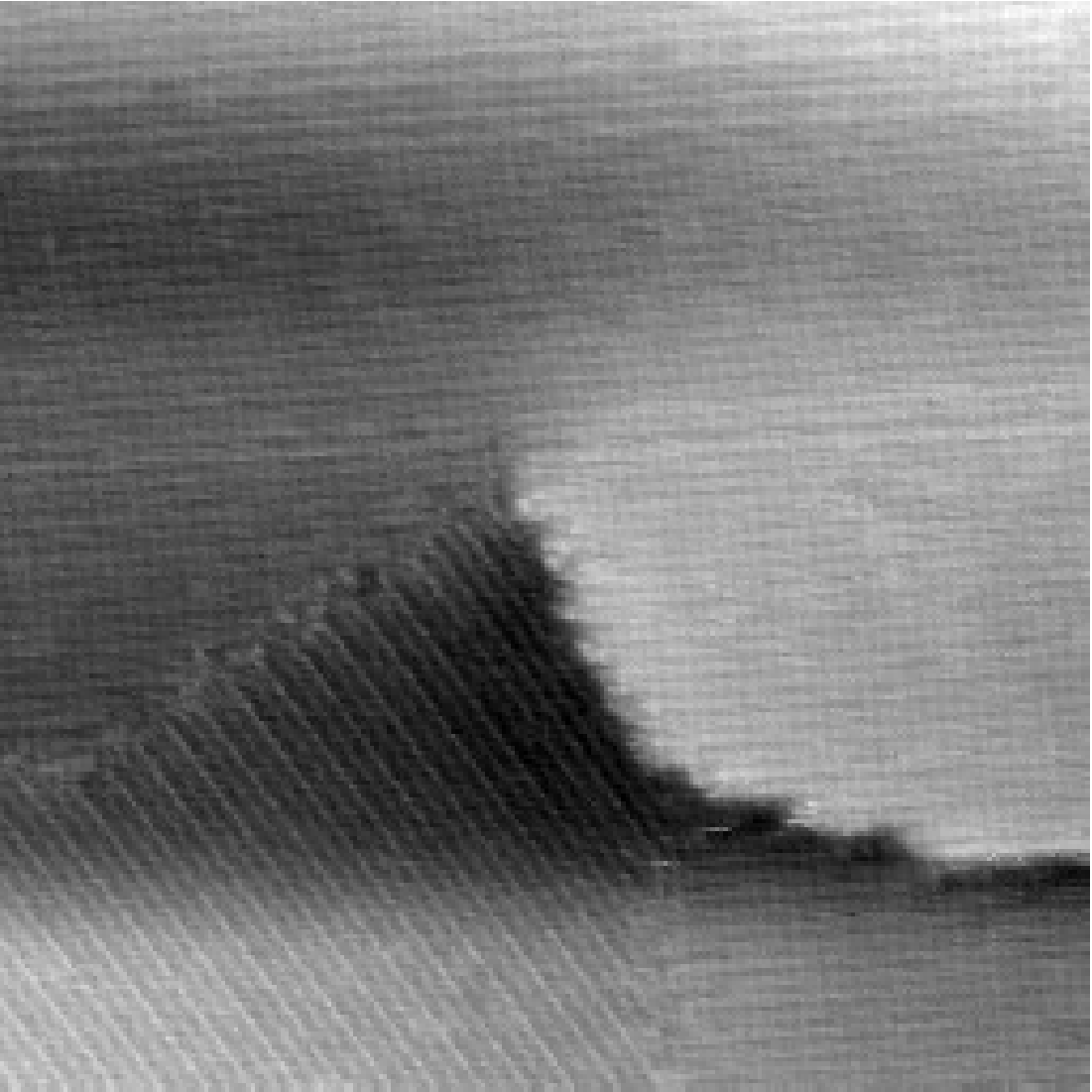}
 \end{subfigure}%
  \begin{subfigure}[t]{0.25\textwidth}
  \centering
  \includegraphics[scale=0.35]{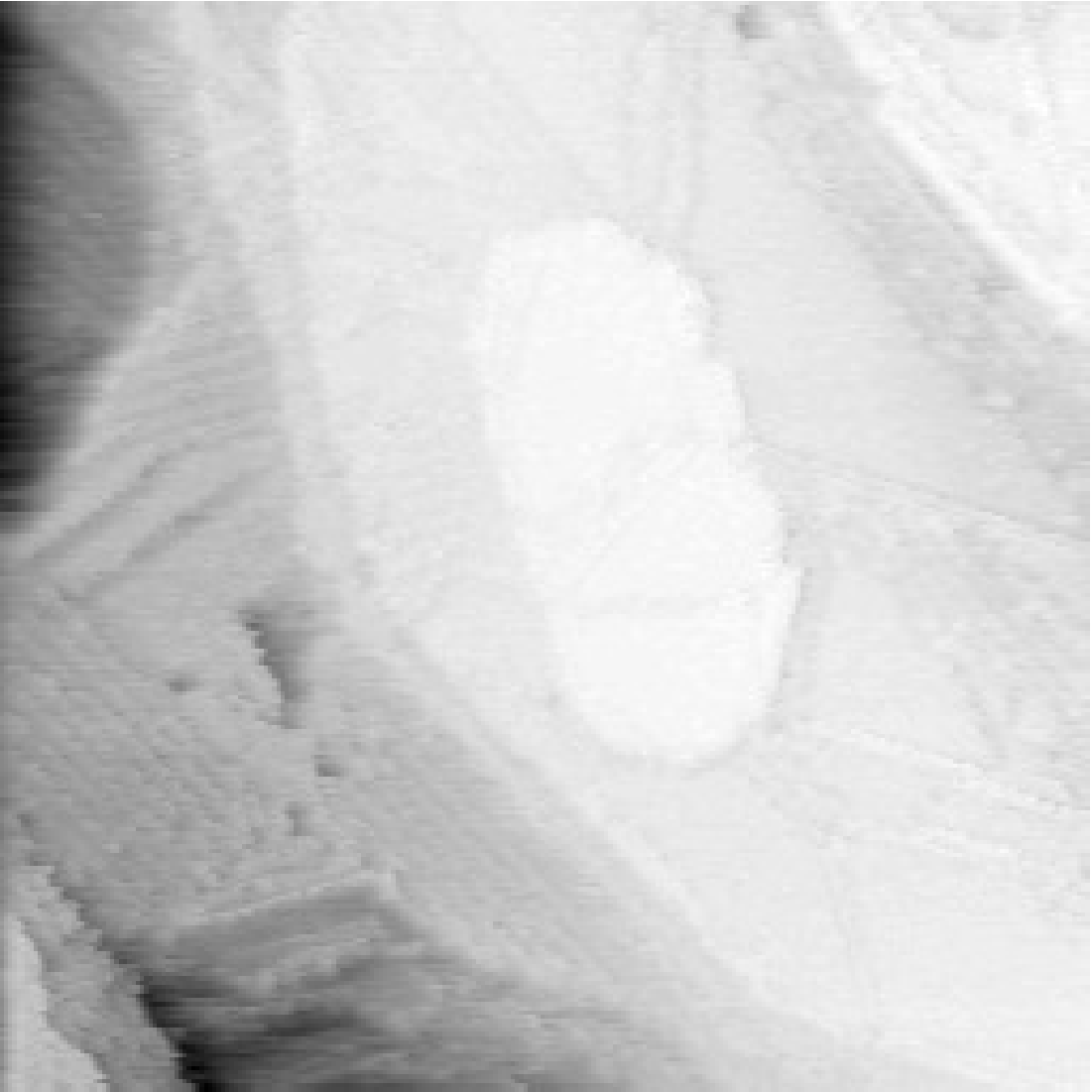}
 \end{subfigure}%
  \begin{subfigure}[t]{0.25\textwidth}
  \centering
  \includegraphics[scale=0.35]{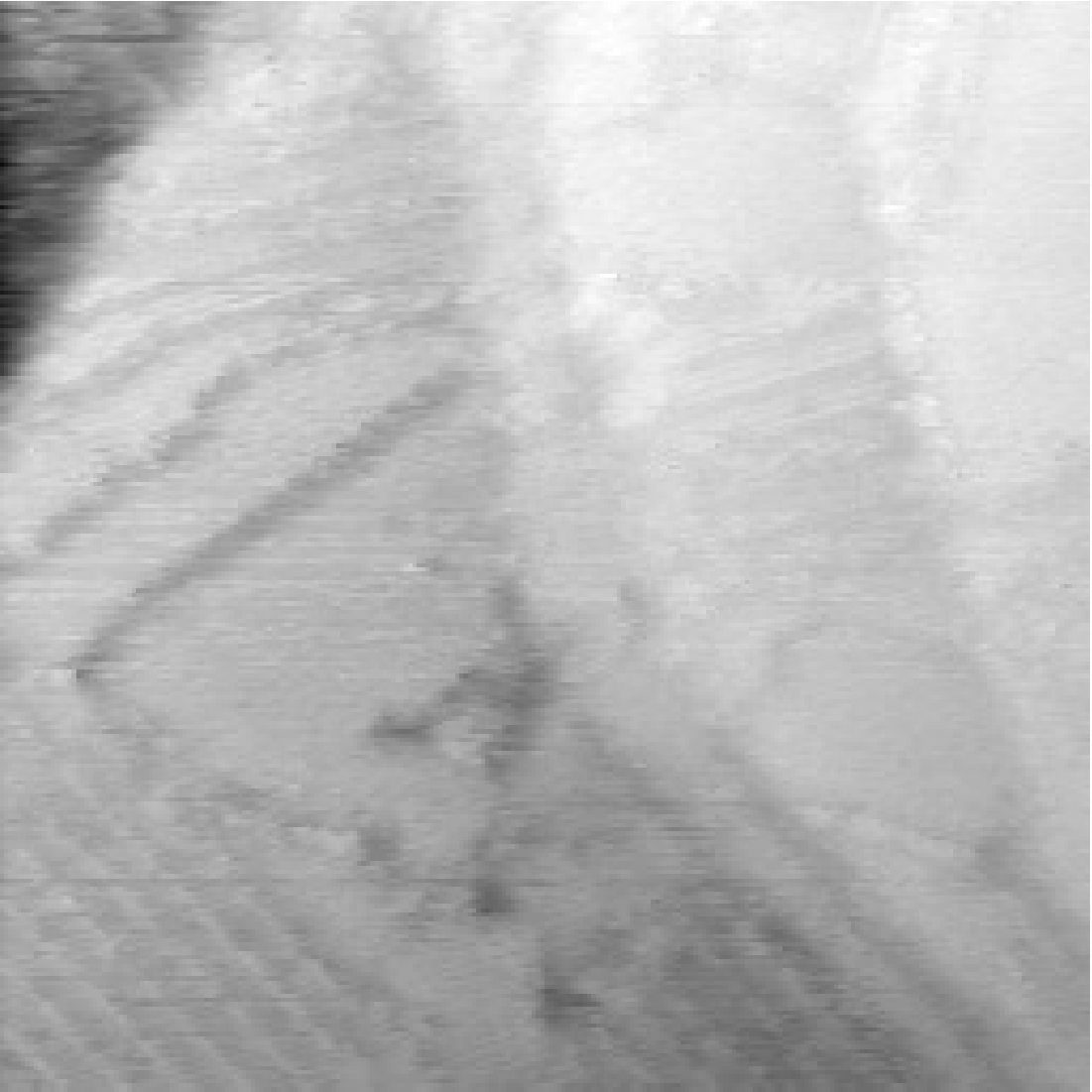}
 \end{subfigure}
\caption{Raw scanning tunneling microscope images of cyanide on Au\{111\}, reproduced from \cite{guttentag2016hexagons} with permissions. Images copyright American Chemical Society}
\label{fig:STMimages}
\end{figure*}

In order to obtain a scanning tunneling microsopy image of a SAM, an atomically sharp conducting probe tip is brought within one or two atomic diameters of the surface of the sample so that electrons can tunnel from the surface to the tip. A voltage bias is applied between the two and the tip-sample separation is typically adjusted while scanning to maintain a constant tunneling current of electrons. Since the current is extremely sensitive to the tip-sample separation, better than atomic resolution is often obtained and apparent height differences across the surface are recorded, thereby acquiring nanoscale images with molecular features. The scanning procedure is shown in Figure~\ref{fig:STMprocess} and examples of STM 
images of cyanide (CN) monolayers on Au\{111\} are shown in Figure~\ref{fig:STMimages}.
These images show the varying textures and different apparent heights (displayed as intensities) as a result of the structure and chemical properties of the SAM. Partitioning the images according 
to the apparent heights and texture patterns would help facilitate the understanding and analyses of SAMs and other surfaces studied. We note that not only are the ordered regions important but also are the boundaries between them \citep{poirier1997characterization} since these domain boundaries determine access of other molecules to the substrate and can be used to isolate single molecules, or pairs, lines, or clusters of molecules \citep{bumm1996single, kim2011creating, claridge2013bottom}.

Here, we propose a novel framework to analyze STM images that produces segmentation based on intensities and segmentation based on texture features. The proposed 
meth\-od 
consists of three main steps. The first step performs a cartoon+texture decomposition of the STM image. Its cartoon and texture components are then analyzed 
separately. The cartoon image is the component containing only the edges or boundaries of homogeneous regions in the images and it is devoid of any oscillatory patterns. 
On the other hand, the texture image consists of oscillatory patterns. We propose two parallel steps: the cartoon component is segmented using a 
variant of the Chan-Vese model \citep{chan2000active} and the texture component is segmented by feeding some classifier with features based on empirical wavelets 
\citep{G1,GTO}.

First, we propose a novel multiphase version of the 
local Chan-Vese model \citep{wang2010efficient} and develop an efficient algorithm based on the MBO scheme \citep{merriman1992diffusion,merriman1994motion} to solve it. The 
proposed 
model creates a more accurate segmentation result that is robust against intensity inhomogeneities compared to the original multiphase Chan-Vese model \citep{CT1}. 
Secondly, we improve the 2D empirical wavelet transform, specifically the empirical curvelet transform, so that it provides a filter bank of empirical curvelets extracting 
meaningful textural information. From the filter bank outputs, we design a texture feature matrix to feed a clustering algorithm in 
order to identify regions of different textural patterns on the STM image.

The paper outline is as follows. In Section~\ref{sec:CTdecom}, we describe the nonlinear decomposition algorithm we use in our approach to decompose 
an STM image into its cartoon and textures components. In Section~\ref{sec:cartoon}, we propose a modified version of the local multiphase model and 
determine its diffuse interface 
approximation in order to develop an algorithm based on the MBO scheme to solve it. In Section~\ref{sec:texture}, we review the empirical wavelet transform (particularly the 
empirical curvelet transform) and propose modifications on how the curvelets' supports are detected. We then explain the design of the texture feature matrix based on the 
empirical curvelet coefficients. In 
Section~\ref{sec:experiments}, our proposed framework is applied to the images in Figure~\ref{fig:STMimages} in order to evaluate its efficacy in partitioning STM images by 
intensities and texture patterns. Finally, in Section~\ref{sec:conclusion}, we summarize our results and discuss possible research directions to further improve the proposed 
framework.

\section{The Cartoon+Texture Decomposition}\label{sec:CTdecom}
As discussed in the introduction, the first step of our algorithm involves the decomposition of an image into its cartoon and textural components.

The cartoon+texture decomposition can be posed as an inverse problem and it consists  of decomposing an image $f: \Omega \rightarrow \mathbb{R}$, where $\Omega$ is the image domain, into 
\[ f = u + v \]
where $u$ and $v$ are the cartoon and the texture components, respectively.
The cartoon image carries broad information about the image and is usually modeled by a function of bounded 
variation 
(piecewise smooth with possibly a discontinuity set). The texture image contains oscillatory information and is thus usually modeled by oscillating functions. 

One of the earliest variational models that inspired cartoon+texture decomposition is the Rudin-Osher-Fatemi total variation minimization model \citep{ROF92}
\begin{equation}\label{eq:rof}
	\inf_{\substack{u\in BV(\Omega)\\v\in L^2(\Omega)}} 
	\Big\{  \sigma \|u\|_{TV}+ \|v\|_{L^2}^2, \; f = u+v \big\},
\end{equation}
where $\|u\|_{TV}$  is the total variation of $u$ and $\sigma$ is a tuning parameter that controls the regularization strength. The model was originally used for denoising purposes because of the functional spaces that $u$ and $v$ belong to. The function $u$ belongs to 
the space of functions of bounded variations 
\[ {BV}(\Omega) = \left\{u \in L^1(\Omega): \int_{\Omega} |\nabla u | < \infty \right\},\] 
which penalizes oscillations such as noise and textures but  allows for piecewise-smooth functions made of homogeneous regions with sharp boundaries. However,  the decomposition is well-posed only in a multiresolution setup since image features can be considered as texture in one scale and cartoon at a different scale.
One of the most popular models in cartoon+texture decomposition is the $TV-L^1$ model proposed by \cite{CE05},  for which \cite{chambolle2004algorithm} provided a fast projection 
algorithm.

Although models, such as \eqref{eq:rof} and the above mentioned, are able to perform cartoon + texture decomposition, \cite{Me01} argued that the texture image extracted  by these models does not fully characterize the oscillatory patterns of the original image. As a result, he proposed to replace the $L^2$ norm in \eqref{eq:rof} by weaker norms  (associated to spaces of oscillatory distributions) in order to better capture the oscillatory patterns. In practice, some of these norms are difficult to compute. To remedy  this drawback,  \cite{buades2011cartoon} developed a nonlinear version of the linearized Mey\-er's model. 

The linearized Meyer's model is
\begin{equation}\label{eq:H1-model}
	\inf_{\substack{ u \in H^{1}(\Omega)\\ v\in  H^{-1}(\Omega)}} 
	\Big \{ \sigma^4 \| \nabla u\|_{L^2}^2 + \|v\|_{H^{-1}}^2, \; f = u + v \Big \},
\end{equation}
where 
$H^1(\Omega) = \{ u \in L^2(\Omega): \nabla u \in L^2(\Omega)\}$ 
and $H^{-1}(\Omega)$ is the dual space of the homogeneous 
version of $H^1(\Omega)$. They are defined in the Fourier domain by
\begin{align*}
&H^1(\Omega) \\
 &=\{ u \in L^2(\Omega): \int_{\Omega} [1+(2\pi|\xi|)^2] |\mathcal{F}_2({u})(\xi)|^2 d\xi < \infty \}
 \end{align*}
 and
 \begin{align*}
 &H^{-1}(\Omega) \\
 &= \{ v \in L^2(\Omega): \int_{\Omega} [1+(2\pi|\xi|)^2]^{-1} |\mathcal{F}_2({v})(\xi)|^2 d\xi < \infty \},
\end{align*}
where $\mathcal{F}_2(\cdot)$ is the 2D Fourier Transform. 
Minimizing the quadratic functional \eqref{eq:H1-model} yields the solution 
\[ u = L_\sigma \ast f \text{ with } \mathcal{F}_2(L_{\sigma})(\xi) = \frac{1}{1 + (2 \pi \sigma |\xi|)^4},\]
where the convolution kernel is given in the Fourier domain. Since $\mathcal{F}_2(L_{\sigma})(\xi)$ defines a low-pass filter, any frequency $\xi$ that is significantly smaller than 
$\frac{1}{2 \pi \sigma}$ is kept in $u$. Otherwise, it is kept in $v$. Thus, the cartoon + texture decomposition is performed by applying a low-pass/high-pass filter onto $f$,
\[ (u,v) = (L_{\sigma} \ast f, (Id - L_{\sigma}) \ast f).\]
The drawback of the low-pass filter is that  sharp edges are altered, while they should  be preserved as much as possible since they are necessary for cartoon segmentation.  

The nonlinear version of \eqref{eq:H1-model} relies on a classifier  which determines whether a pixel of the original image belongs to  the cartoon or the texture component \citep{buades2011cartoon}. The idea consists of measuring the rate of change of the local total variation (LTV) between the original image and its lowpass filtered version, defined by
\begin{equation}\label{eq: reduction}
	\lambda_{\sigma}(x) = \frac{LTV_{\sigma}(f)(x) - LTV_{\sigma}(L_{\sigma} \ast f)(x)}{LTV_{\sigma}(f)(x)},
\end{equation}
where
\begin{equation}
	LTV_{\sigma}(f)(x) = L_{\sigma} \ast |\nabla f|(x).
\end{equation} 
If a neighbourhood of a pixel $x$ does not contain any textures, then $f$ and $L_\sigma\ast f$ will be similar within the neighbourhood, so $\lambda_{\sigma}(x)$ is close to $0$. If some textures are\\ present within the neighborhood, then the total variation of the neighborhood in the filtered image will be smaller and\\ $\lambda_\sigma(x)$ is close to $1$. Therefore, the cartoon component $u$ is computed as the weighted average between the original image $f$ and the filtered image $L_{\sigma} \ast f$ depending on $\lambda_{\sigma}$, i.e.
\[ u (x) = w\big(\lambda_{\sigma}(x)\big)(L_{\sigma} \ast f)(x)+ \big(1-w(\lambda_{\sigma}(x))\big)f (x),\]
with the soft threshold function $w: [0,1] \rightarrow [0,1]$ given by
\begin{equation*}
 w(x)=\begin{cases}
       0 &\text{if}\quad x< a_1\\
       (x-a_1)/(a_2-a_1)\quad &\text{if}\quad a_1\leq x\leq a_2\\
       1 &\text{if}\quad a_2< x
      \end{cases},
\end{equation*}
where $0<a_1 < a_2<1$. For our experiments in Section~\ref{sec:experiments}, as suggested by  \cite{buades2011cartoon}, we fixed $a_1 = 0.25$ and $a_2 = 0.50$. The texture component is easily obtained by computing the difference between the original image and its cartoon component. The algorithm for the nonlinear version is summarized in Algorithm~\ref{alg:NonLinear_CTD}.

\begin{algorithm}[!t]
\caption{Cartoon+Texture Decomposition} \smallskip
\KwIn{Original image $f$, parameter $\sigma >0$}
{
  \begin{algorithmic}[1]
    \STATE Compute the LTV reduction rate using \eqref{eq: reduction} at each pixel.
    \STATE Obtain the cartoon component: 
   		\begin{align*}
   		\begin{split}
   		    u(x) &\gets w\big(\lambda_{\sigma}(x)\big)(L_{\sigma} \ast f)(x)\\&+ \big(1-w(\lambda_{\sigma}(x))\big)f(x)
   		    \end{split}
   		\end{align*}
    \STATE Obtain the texture component:
   		$$v(x) \gets f(x)-u(x)$$
  \end{algorithmic}
  }
\KwOut{Cartoon and texture components $u$, $v$}
\label{alg:NonLinear_CTD}
\end{algorithm}

\begin{figure}[ht]
 \centering
  \begin{subfigure}[t]{0.5\columnwidth}
  \centering
  \includegraphics[width=0.9\columnwidth]{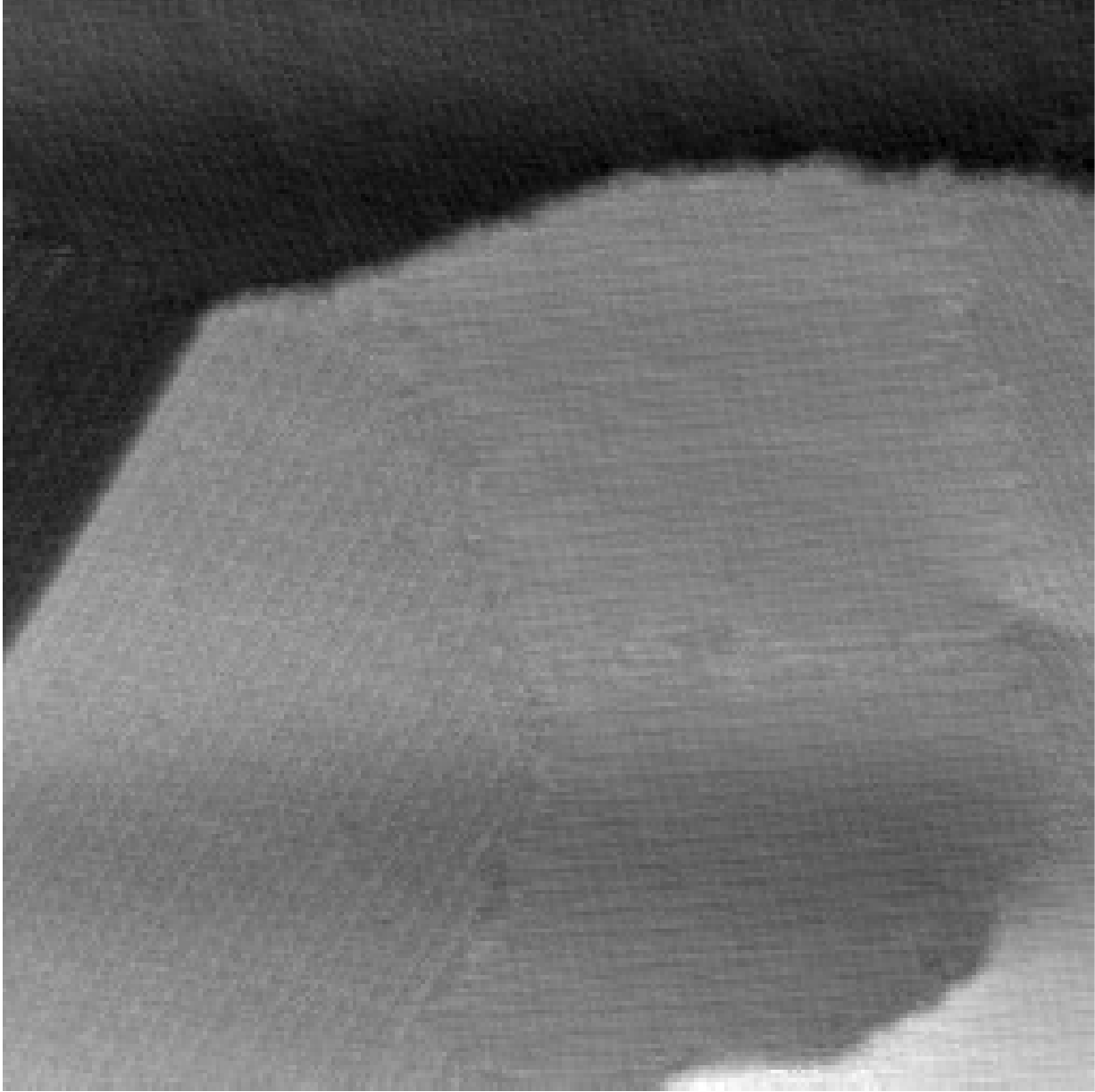}
  \caption{Original STM image}
  \label{fig:dec.a}
  \end{subfigure}%
  \vskip 1mm

  \begin{subfigure}[t]{0.5\columnwidth}
  \centering
  \includegraphics[width=0.9\columnwidth]{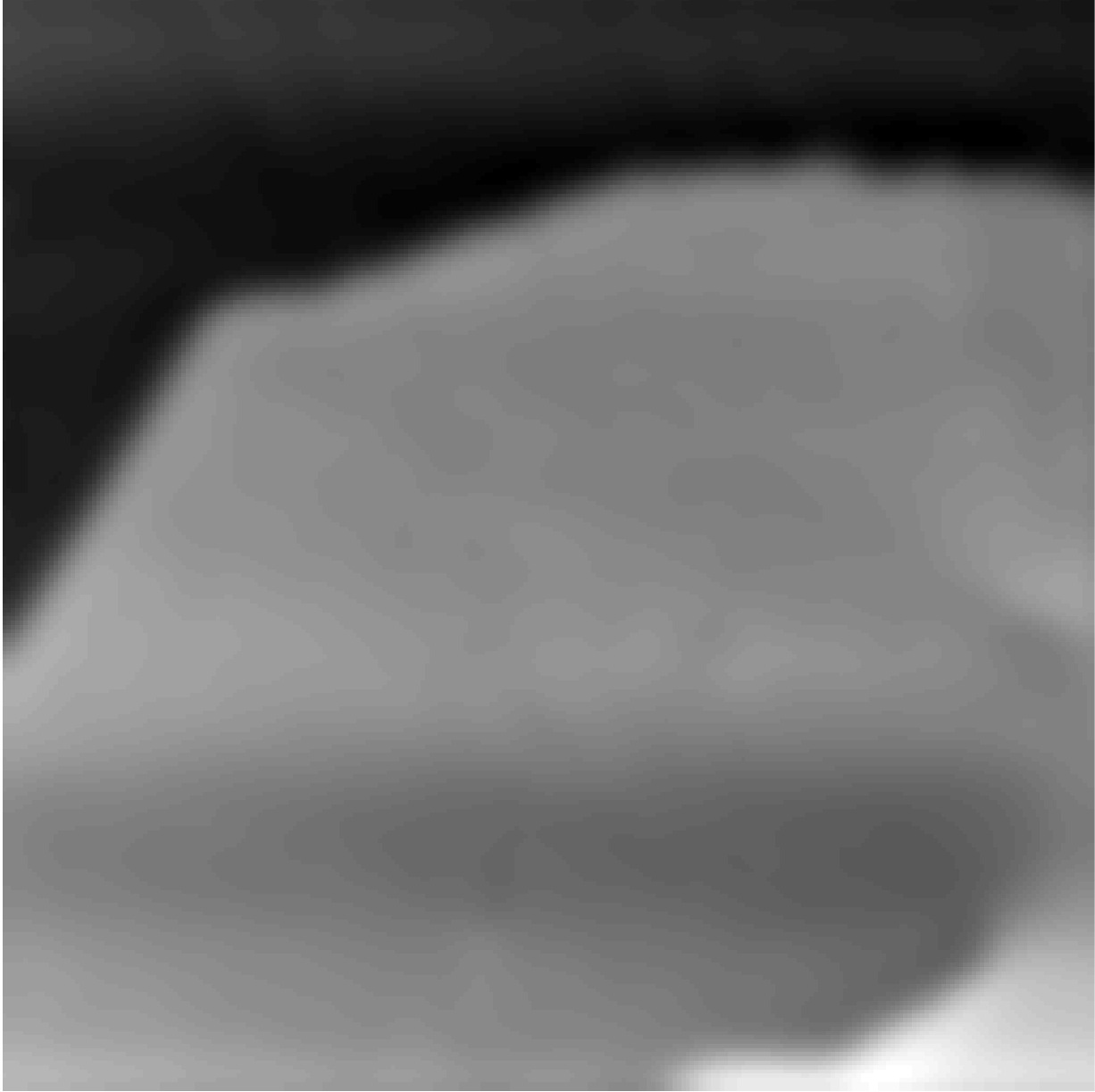}
  \caption{Linear filtering: cartoon}
  \label{fig:dec.b}
  \end{subfigure}%
  \begin{subfigure}[t]{0.5\columnwidth}
  \centering
  \includegraphics[width=0.9\columnwidth]{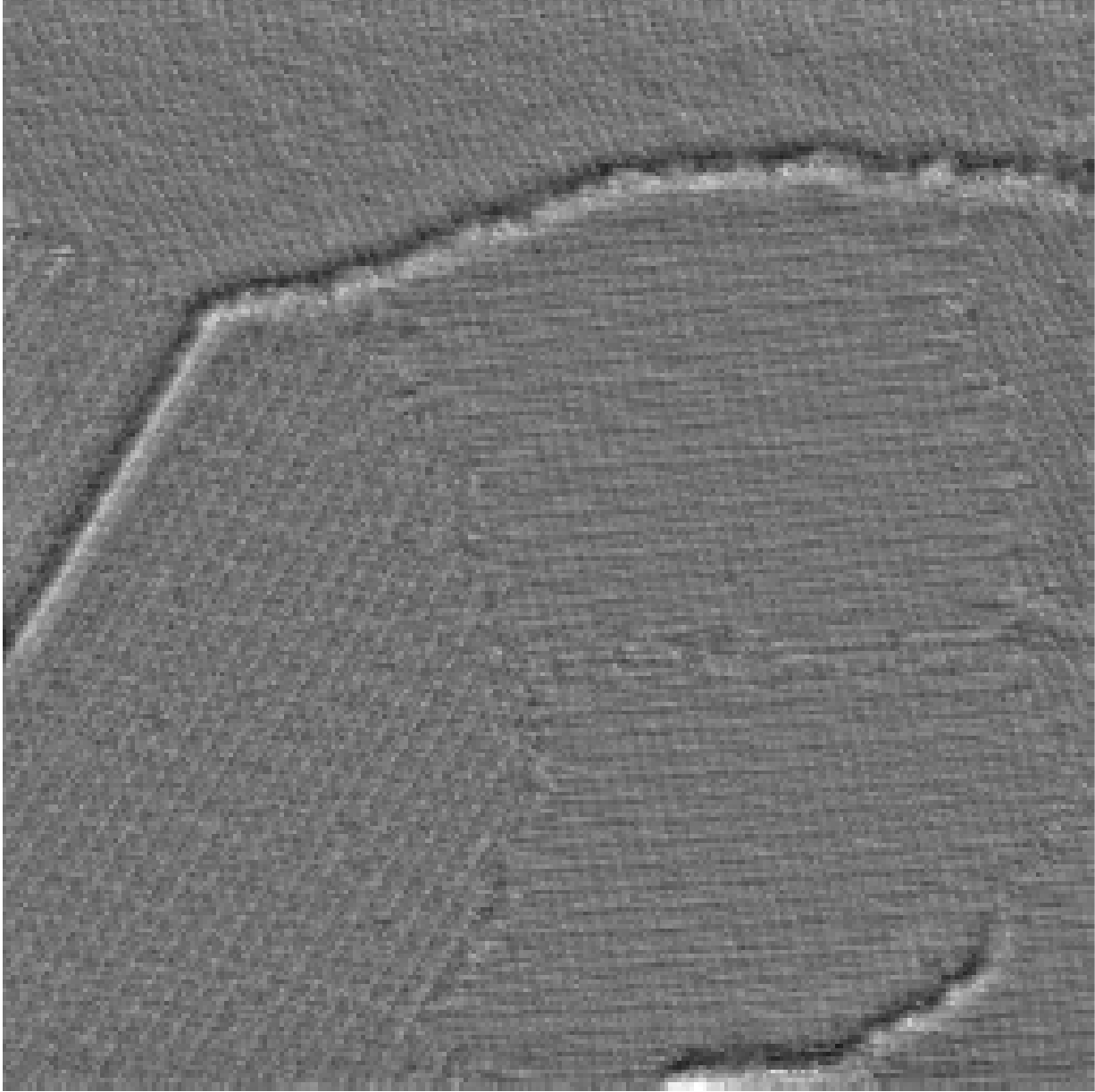}
  \caption{Linear filtering: texture}
  \label{fig:dec.c}
  \end{subfigure}%
  \vskip 1mm
  
  \begin{subfigure}[t]{0.5\columnwidth}
  \centering
  \includegraphics[width=0.9\columnwidth]{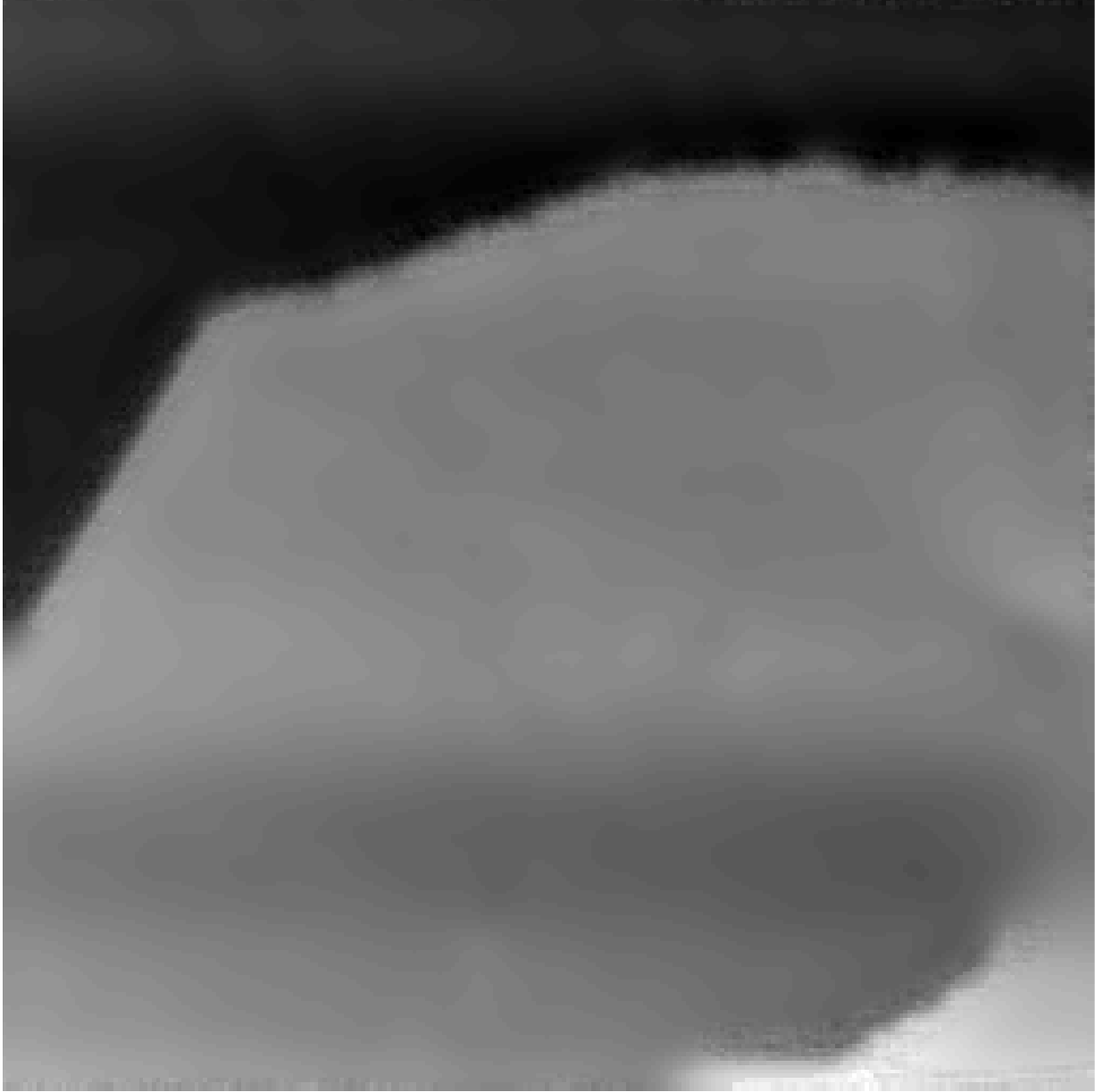}
  \caption{Nonlinear filtering: cartoon}
  \label{fig:dec.d}
  \end{subfigure}%
  \begin{subfigure}[t]{0.5\columnwidth}
  \centering
  \includegraphics[width=0.9\columnwidth]{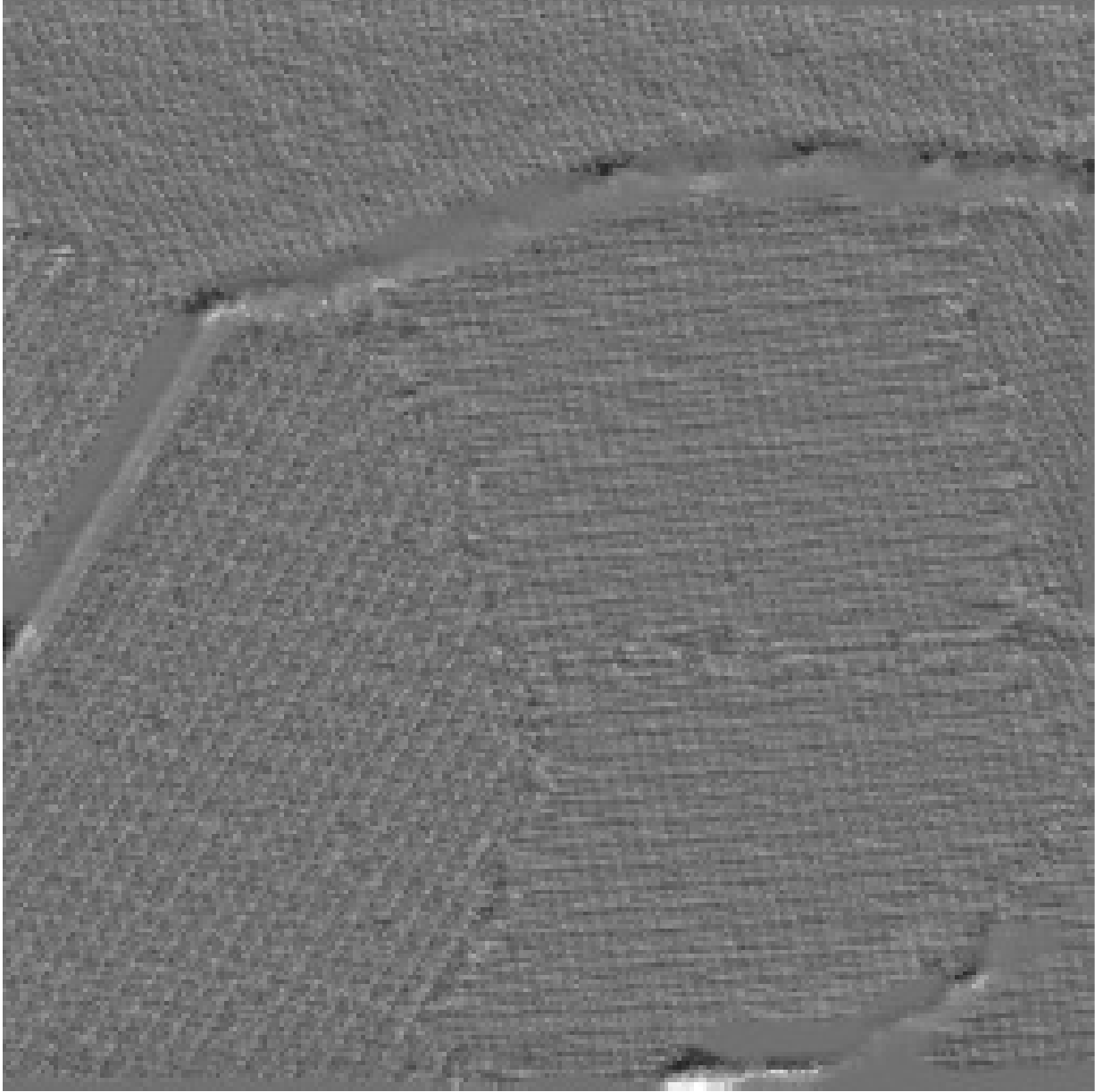}
  \caption{Nonlinear filtering: texture}
  \label{fig:dec.e}
  \end{subfigure}
  
  \caption{Results of linear and nonlinear cartoon texture decompositions with $\sigma = 3$, reproduced from \cite{guttentag2016hexagons} with permission. Image in (a) copyright American Chemical Society}
  \label{fig:CTdecomp}
\end{figure}

In Figure~\ref{fig:CTdecomp}, we present results for both linear and nonlinear decompositions for one of the images in Figure~\ref{fig:STMimages} to illustrate why choosing the nonlinear version is more interesting. Although both decompositions succeed in separating the image into its cartoon and texture components, as shown in Figures~\ref{fig:dec.b} and \ref{fig:dec.d}, the nonlinear decomposition gives sharper edges in the cartoon component than does the linear decomposition. Since in general these edges do not correspond to textures associated with molecular orientations but rather with topographic transitions of the molecular layers, the nonlinear version is therefore preferable.

\section{Segmentation of the Cartoon Image}\label{sec:cartoon}

The cartoon component of an STM image provides information about the stratification  of scanned molecules. The lighter the gray intensity is, the greater the apparent height. Here, the largest variations are due to monoatomic steps in the substrate; a single layer of molecules is present everywhere on the surface.   Identifying different gray level regions is therefore crucial for finding the topographic properties of the substrate and chemical layers.

We propose to use a level-set approach based on the multiphase Chan-Vese (multiphase CV) model to perform the segmentation of the cartoon component. The Chan-Vese model is inspired by the pioneering work of \cite{MS89}, who suggested performing image segmentation by solving the minimization problem
\begin{equation}\label{eq::MS}
	\inf_{u,\Gamma} \left\{
		\int_{\Omega \backslash \Gamma} |\nabla u|^2 +
		\lambda\int_\Omega (u - u_0)^2 +		
		\mu \int_{\Gamma}ds\right\}, 
\end{equation}
where $u_0:\Omega\to\R$ is the cartoon image to be segmented and $\mu$ and $\lambda$ are weighing parameters. The function $u$ is a piecewise  smooth  approximation of $u_0$, which is allowed to have jumps across $\Gamma$ -- a closed subset of $\Omega$ given by a finite union of rectifiable curves. 

Although several algorithms have been proposed to compute the solution \citep{ambrosio1990approximation}, its computation \citep{luminita1997reduced,march1992visual} is relatively complicated and numerically expensive. To  overcome this drawback, 
simplifications of the energy functional \eqref{eq::MS} to piecewise constant functions have been proposed.

Based on the level-set approach \citep{osher1988fronts}, the multiphase CV segmentation model \citep{CT1} remains one of the most popular models. 
In this setup, the function $u$ is allowed to have only four values $c_1,c_2,c_3$, and $c_4$, one for each of the four distinct regions. Furthermore,  using two level-set functions $\phi_1$ and $\phi_2$, thus producing four phases $\{\phi_1>0, \phi_2>0\}, \{\phi_1>0, \phi_2<0\}, \{\phi_1<0, \phi_2>0\}$, and  $\{\phi_1<0, \phi_2>0\}$, one can show it is sufficient to generate a partition of the domain $\Omega$ into regions of different intensities having triple junctions or $T$-junctions such that
$\Gamma = \{\phi_1=0\} \cup \{\phi_2 = 0\}$ (See \cite{CT1} for details).  Let $\mathbf{u} = (u_1,u_2)$ be a vector-valued function and $\mathbf{c} = (c_1,c_2,c_3,c_4)$ be a vector of constants. The fidelity term is defined as
\begin{equation} \label{eq:fidelity}
\begin{split}
	\mathcal E_{fid}(\mathbf{c}, \mathbf{u}) = 
				   \int_{\Omega} (c_1 - u_0)^2 u_1 u_2 \\
				+ \int_{\Omega} (c_2 - u_0)^2 u_1 (1-u_2) \\
				+ \int_{\Omega} (c_3 - u_0)^2(1-u_1) u_2 \\
				+ \int_{\Omega}(c_4 - u_0)^2(1-u_1) (1-u_2).
\end{split}
\end{equation}
The perimeter term is defined by
\begin{equation} \label{eq:perimeter}
	\mathcal E_{per}(\mathbf{u}) = \int_{\Omega} |\nabla  u_1| + \int_{\Omega} |\nabla u_2|.
\end{equation}

Using the standard Heaviside function $H$ with the notations $\Phi=(\phi_1,\phi_2)$  and $H(\Phi)=(H(\phi_1),H(\phi_2))$, the four-phase piecewise constant Mumford-Shah model can be written in terms of the level set functions $\phi_1$ and $\phi_2$ as
\begin{equation} \label{eq:CV}
\begin{split}
	\mathcal E_{CV}(\mathbf{c}, H(\Phi)) = 
	\lambda \mathcal E_{fid}(\mathbf{c}, H(\Phi)) + \mu \mathcal E_{per}(H(\Phi)).
\end{split}
\end{equation}

\subsection{Local Multiphase Chan-Vese Model}

The model \eqref{eq:CV} is, however, not robust against illumination bias, such as shadows, or intensity inhomo\-gene\-ities, such as weak edges, usually present in STM images. For 
example,  if a region of an image  is partially overlapped by a shadow, it might be segmented into two regions. Or, on the contrary, regions with inhomo\-geneous gray-level intensities, may not be segmented at all, even though they might provide valuable information.

\cite{wang2010efficient} proposed a local term that can be added to \eqref{eq:CV} to counteract the lighting issues. We define, as before, a local term 
\begin{align}\label{eq:local}
\begin{split}
	\mathcal E_{loc}(\mathbf{d}, \mathbf{u}) = 
	\int_{\Omega} (g_k\ast u_0-u_0-d_1)^2  u_1 u_2 \\ 
			+ \int_{\Omega}  (g_k\ast u_0-u_0-d_2)^2 u_1(1-u_2) \\
			+ \int_{\Omega}  (g_k\ast u_0-u_0-d_3)^2 (1-u_1) u_2\\
			+ \int_{\Omega}  (g_k\ast u_0-u_0-d_4)^2(1- u_1)(1-u_2),
\end{split}
\end{align}
where $g_k$ is a convolution kernel with $(k \times k)$-size window and $\mathbf{d} = (d_1, d_2, d_3,d_4)$ is a vector-valued function. The local multiphase Chan-Vese model (local MCV) requires minimizing the energy functional
\begin{equation} \label{eq:CVloc}
	\mathcal E_{CVloc} (\mathbf{c}, \mathbf{d}, H(\Phi)) = 
	\mathcal E_{CV}(\mathbf{c}, H(\Phi)) + \beta \mathcal E_{loc}(\mathbf{d}, H(\Phi)).
\end{equation}

With $g_k$ as a low-pass filter, the image difference between the  filtered cartoon image and the original cartoon image will have its edges properly identified and the areas with slowly varying intensities disregarded.  By incorporating the  image  difference, the model would take into account weak edges and illumination bias.

The effectiveness of incorporating \eqref{eq:local} depends on the choice of the convolution kernel $g_k$. Since $g_k$ needs to be a low-pass filter, one issue to be aware of  is over-smoothing of the edges. For the purpose of our paper, we use a Gaussian filter, which weighs more pixels close to the center than pixels  distant from it. This property effectively preserves the edges of the image as the image is being\\ smoothed. Moreover, the choice of standard deviation for the Gaussian filter enables greater control in the amount of smoothing.

\subsection{Ginzburg-Landau Approximation for the Local MCV Model}

Computing the minimizer for the energy functional \eqref{eq:CVloc} can be computationally expensive. For example, see the work of \cite{getreuer2012chan}. Instead, one could alternatively obtain an approximate solution by threshold dynamics, suggested by  \cite{ET}. To this end, we  approximate the energy functional  \eqref{eq:CVloc} by a sequence of energies 
\begin{equation} \label{eq:CVloc-eps}
	\mathcal E^\eps_{CVloc} (\mathbf{c}, \mathbf{d}, \mathbf{u}) = \lambda\mathcal E_{fid}(\mathbf{c}, \mathbf{u}) + \mu\mathcal{E}^{\eps}_{GL}(\mathbf{u}) +
	\beta \mathcal E_{loc}(\mathbf{d}, \mathbf{u}),
\end{equation}
where the perimeter term $\mathcal E_{per}$ in the multiphase CV model \eqref{eq:CV} is replaced by the Ginzburg-Landau functional
\begin{equation}\nonumber
 	\mathcal E^\eps_{GL} (\mathbf{u}) =   
	\epsilon \int_{\Omega} (|\nabla u_1|^2 +|\nabla u_2|^2) + 
	\frac{1}{\epsilon} \int_{\Omega} (W(u_1)+W(u_2)),
\end{equation}
where $W(u) = u^2(1-u)^2$. Applying the results of \cite{modica1977esempio,modica1987gradient}, it can be shown that $\mathcal E^\eps_{GL} (\mathbf{u})$ $\Gamma$-converges to $\mathcal E_{per} (\mathbf{u})$ as $\epsilon \rightarrow 0^+$.
Note that everything in the new energy functional is now expressed in terms of the new functions $u_i$ instead of $H(\phi_i)$ for $i=1,2$.

By calculus of variations, keeping $\mathbf{d}$ and $\mathbf{u}$ fixed and minimizing with respect to $\mathbf{c}$, we obtain
\begin{align} \begin{split}\label{eq:c}
 	c_1 &= \mathcal A[u_0,u_1,u_2]\\
	c_2 &= \mathcal A[u_0,u_1,1-u_2]\\
	c_3 &= \mathcal A[u_0,1-u_1,u_2]\\
 	c_4 &= \mathcal A[u_0,1-u_1,1-u_2],
\end{split}\end{align}
where 
\[\mathcal A[u_0,u_1,u_2] = \frac{\int_{\Omega} u_0 u_1  u_2}{\int_{\Omega} u_1 u_2}.\]
Similarly, keeping $\mathbf{c}$ and $\mathbf{u}$ fixed and minimizing with respect to $\mathbf{d}$, we obtain
\begin{align} \begin{split}\label{eq:d}
d_1 &= \mathcal A[g_k\ast u_0-u_0,u_1,u_2]\\
d_2 &= \mathcal A[g_k\ast u_0-u_0,u_1,1-u_2]\\
d_3 &= \mathcal A[g_k\ast u_0-u_0,1-u_1,u_2]\\
d_4 &= \mathcal A[g_k\ast u_0-u_0,1-u_1,1-u_2].
\end{split}\end{align}
 Keeping now $\mathbf{c}$ and  $\mathbf{d}$ fixed and minimizing $\mathcal E^\eps_{CVloc}$ with respect to  $\mathbf{u}$, we deduce the Euler-Lagrange equations for $\mathbf{u}$. It is a common technique to parametrize the descent direction by an artificial time variable $t\geq 0$ and initialize $\mathbf{u}(x,0)$. The equations for  $\mathbf{u}(x,t) = (u_1(x,t), u_2(x,t))$ are
\begin{align}
	\begin{split}\label{eq:u1}
	\frac{\partial{u_1}}{\partial{t}} = 
		& - \lambda \mathcal{L}(u_0,\mathbf{c},u_2)[u_1] \\
		& - \beta \mathcal{L}(g_k\ast u_0-u_0,\mathbf{c},u_2)[u_1]\\
		& +\mu\left(2\eps\Delta u_1 - \frac{1}{\eps} W'(u_1) \right)
	\end{split}
\end{align}
\begin{align}
	\begin{split}\label{eq:u2}
	\frac{\partial{u_2}}{\partial{t}} = 
		& - \lambda \mathcal{L}(u_0,\mathbf{\bar c},u_1)[u_2] \\
		& - \beta  \mathcal{L}(g_k\ast u_0-u_0,\mathbf{\bar c},u_1)[u_2]\\
		& +\mu\left(2\eps\Delta u_2 - \frac{1}{\eps} W'(u_2)\right),
	\end{split}
\end{align}
where $\mathbf{\bar c} = (c_1,c_3,c_2,c_4)$ and the operator  $\mathcal{L}(u_0,\mathbf{c},u_2)[u_1]$ 
is constant with respect to $u_1$, is linear in  $u_2$, and is given by
\[ \begin{split} 
	\mathcal{L}(u_0,\mathbf{c},u_2)[u_1] =  
	& (c_1-u_0)^2 u_2 +(c_2-u_0)^2  (1-u_2) \\
	& -(c_3-u_0)^2 u_2 - (c_4-u_0)^2(1-u_2).
\end{split} \]
At $t=0$, we initialize $u_1$ and $u_2$ as the checkerboard functions 
\begin{align}
\begin{split}\label{eq:u-ini}
u_1(x,0) & = \mathbbm{1}_{\{\sin{\frac{\pi x_1}{3}}\sin{\frac{\pi x_2}{3}} > 0 \}} \\
u_2(x,0) &= \mathbbm{1}_{\{\sin{\frac{\pi x_1}{10}}\sin{\frac{\pi x_2}{10}}>0\}},
\end{split}\end{align}
where $\mathbbm{1}$ is the characteristic function, since initialization of the checkerboard function was observed to have faster convergence to the solution for the two-phase model \citep{getreuer2012chan}.
\subsection{MBO scheme for Solving the Local MCV}

In order to solve the above system of parabolic PDEs, we implement the MBO scheme \citep{merriman1992diffusion,merriman1994motion}. The underlying idea is to solve the PDE in three steps: solve the linear ODE 
\begin{align}\label{eq:ode}
 \frac{\partial{w}}{\partial{t}} = 
 	 - \lambda \mathcal{L}(u_0,\mathbf{c},u)[w]  
 	 - \beta  \mathcal{L}(g_k\ast u_0 - u_0,\mathbf{c},u)[w],
\end{align} 
solve the heat equation,
\begin{align}\label{eq:heat} 
	\frac{\partial{v}}{\partial{t}} = 2\mu \epsilon \Delta v 
\end{align} 
and apply thresholding, which corresponds to solving the nonlinear ODE
\begin{align}\label{eq:thre} 
	\frac{\partial u}{\partial t} = -\frac{\mu}{\eps}W'(u).
\end{align}	
We use the fact that as $\eps\to 0$, the solution is approaching one of the steady states ($u=0$ and $u=1$).
Since we are dealing with a system of equations, we need to alternate the steps between the two functions $u_1$ and $u_2$.

Using the MBO scheme, we develop an iterative algorithm that computes a sequence of solutions 
$ \left\{ \mathbf{c}^n, \mathbf{d}^n, \mathbf{u}^n\right\}.$
After initialization of $\mathbf{u}^0 = (u_1^0, u_2^0)$ as the checkerboard functions \eqref{eq:u-ini} and the computation of
\begin{align*}
    \tilde{u}_0 = g_k\ast u_0 - u_0,
\end{align*}
for each iteration $n\in\N$ we proceed as follows:
\begin{enumerate}
	\item Compute the average intensities
		\begin{align} \begin{split}\label{eq:cn}
 		c_1^n &= \mathcal A[u_0,u_1^n,u_2^n]\\
		c_2^n &= \mathcal A[u_0,u_1^n,1-u_2^n]\\
		c_3^n &= \mathcal A[u_0,1-u_1^n,u_2^n]\\
 		c_4^n &= \mathcal A[u_0,1-u_1^n,1-u_2^n].
		\end{split}\end{align}
		and average differences of intensities
		\begin{align} \begin{split}\label{eq:dn}
 		d_1^n &= \mathcal A[\tilde{u}_0,u_1^n,u_2^n]\\
		d_2^n &= \mathcal A[\tilde{u}_0,u_1^n,1-u_2^n]\\
		d_3^n &= \mathcal A[\tilde{u}_0,1-u_1^n,u_2^n]\\
 		d_4^n &= \mathcal A[\tilde{u}_0,1-u_1^n,1-u_2^n].
		\end{split}\end{align}

	\item Let $v_1^{n+1} $ be the solution of the ODE \eqref{eq:ode}, with initial data $u_1^n$, 
		computed at time $\;dt$ for the operator $\mathcal{L}(\cdot,\mathbf{c},u_2^n).$
		This can be easily solved via a finite difference scheme of the form
	 	\begin{equation}\label{eq:ode1-n}
		\begin{split} 
		\frac{ v_1^{n+1} - u_1^n}{\;dt} = 
			&  -\lambda \mathcal{L}(u_0,\mathbf{c},u_2^n)[u_1^n]  \\
 			& -\beta  \mathcal{L}(\tilde{u}_0,\mathbf{c},u_2^n)[u_1^n].
		\end{split}
		\end{equation}	
	\item Let $w_1^{n+1}$ be the solution of the heat equation \eqref{eq:heat}, with initial data $v_1^{n+1}$, 
		computed at time $\;dt$. The equation can be solved in the Fourier domain as
		\begin{equation}\label{eq:heat1-n}
		\mathcal{F}_2({w}_1^{n+1}) = \frac{1}{1+2\mu\;dt |\xi|^2} \mathcal{F}_2({v}_1^{n+1}).
		\end{equation}
	\item Threshold to approach steady-state solutions of  equation \eqref{eq:thre} as follows:
		\begin{align}\label{eq:thre1-n}
		u_1^{n+1} = 
			\begin{cases}  
    				0 \quad&\textrm{if}\quad w_1^{n+1} \in (-\infty, \frac{1}{2}]\\
    				1 \quad&\textrm{if}\quad w_1^{n+1} \in (\frac{1}{2}, +\infty).
			\end{cases}
		\end{align}
	\item Set $v_2^{n+1} $ to be the solution of the ODE \eqref{eq:ode}, with initial data $u_2^n$, 
		computed at time $\;dt$, for the operator\\ $\mathcal{L}(\cdot,\mathbf{\bar c},u_1^{n+1}). $
		This can be  solved, as before, via a finite difference scheme of the form
	 	\begin{equation}\label{eq:ode2-n}
		\begin{split} 
		\frac{ v_2^{n+1} - u_2^n}{\;dt} = 
			& -\lambda \mathcal{L}(u_0,\mathbf{\bar c},u_1^{n+1})[u_2^n]  \\
 			& -\beta  \mathcal{L}(\tilde{u}_0,\mathbf{\bar c},u_1^{n+1})[u_2^n].
		\end{split}
		\end{equation}
	\item Let $w_2^{n+1}$ be the solution of the heat equation \eqref{eq:heat}, with initial data $v_2^{n+1}$, 
		computed at time $\;dt$. As before, we solve the equation in the Fourier domain as
		\begin{equation}\label{eq:heat2-n}
		\mathcal{F}_2({w}_2^{n+1}) = \frac{1}{1+2\mu\;dt |\xi|^2} \mathcal{F}_2({v}_2^{n+1}).
		\end{equation}		
	\item	Threshold to approach steady-state solutions of  equation \eqref{eq:thre} as follows:
		\begin{align}\label{eq:thre2-n}
		u_2^{n+1} = 
			\begin{cases}  
    				0 \quad&\textrm{if}\quad w_2^{n+1} \in (-\infty, \frac{1}{2}]\\
    				1 \quad&\textrm{if}\quad w_2^{n+1} \in (\frac{1}{2}, +\infty).
			\end{cases}
		\end{align}
\end{enumerate}

In order to obtain the segmented result $\tilde{u}$, we multiply $u_2$ by two and add it to $u_1$ to form at most four segmented regions. If we simply add $u_1$ and $u_2$  together,  we would have at most three regions. The algorithm for minimizing \eqref{eq:CVloc} is summarized in Algorithm~\ref{alg:LMCV}. Results are shown and discussed in Section \ref{sec:experiments}.

\begin{algorithm}[t]
\caption{MBO scheme for local MCV}\smallskip
\KwIn{Image $u_0$, parameters $\lambda,\mu,\beta,\;dt$}
  \begin{algorithmic}[1]
    \STATE Compute $$\tilde u_0 = g_k \ast u_0 - u_0$$ using the Gaussian filter, preferably imgaussfilt in MATLAB.
    \STATE Initialize $u_1$ and $u_2$ as in \eqref{eq:u-ini}.
   	 \FOR{$i = 1$ to $n$}
    		\STATE Compute the average intensities $\mathbf{c}^i$ as in \eqref{eq:cn}.
        		\STATE Compute the average intensities $\mathbf{d}^i$  as in \eqref{eq:dn}.
       	 	\STATE Compute the solution $\mathbf{u}^i$ using equations\\ \eqref{eq:ode1-n}-\eqref{eq:thre2-n}.
   	\ENDFOR
     \STATE Combine $u_1^n$ and $u_2^n$ to obtain multiphase image $\tilde{u}^n$,\\ i.e. $$\tilde{u}^n = u_1^n + 2u_2^n$$
  \end{algorithmic}
\KwOut{Segmented Image $\tilde{u}^n$}
\label{alg:LMCV}
\end{algorithm}

\section{Texture segmentation using Empirical Wavelet Transform}\label{sec:texture}
STM images may be comprised of various texture patterns, as shown in Figure~\ref{fig:STMimages}.  These patterns are not in general simple waves but rather a combination of several simple oscillations. The goal of this section is to partition the image textures into several components, each grouping together pixels belonging to a similar pattern.  Texture segmentation is in general a difficult  task because it needs to take into account all types of variability within textures. Such difficulties can be leveraged in the case of particular textures, as for STM images, since the textures correspond to periodic patterns at different frequencies and with different orientations.

Directional image decomposition methods involve a decomposition of the Fourier spectrum into basis elements. These methods include Gabor filters \citep{jain1991unsupervised, dunn1993optimal, dunn1994texture, dunn1995optimal,weldon1996efficient}, wavelets \citep{strang1993wavelet, arivazhagan2003texture, unser1995texture}, curvelets \citep{arivazhagan2006texture, candes2005continuous, candes2006fast, shen2009texture} and shearlets \citep{guo2006sparse}. However, they are not adaptive and may result in incoherent partitions for STM-type images. Adaptive methods, on the other hand, provide better image decomposition since the basis elements are generated by the information contained in the image itself. Among these adaptive methods, the 2D variational mode decomposition (VMD) is a non-recursive, fully adaptive algorithm that sparsely decomposes signals/images into ensembles of constituent modes \citep{dragomiretskiy2014variational, dragomiretskiy2015two,zosso2017two}. By minimizing an energy functional, this method simultaneously retrieves a given number of modes (texture patterns) together with their supports and  the frequencies around which they are band-limited. To accommodate microscopy images, where texture patterns are combinations of simple modes, the energy functional is adapted to lattices by coupling several modes with a single support function. The solution to the energy functional is optimized using alternating direction method of multipliers (ADMM) \citep{gabay1976dual,glowinski1975approximation} and the MBO scheme, requiring several parameters to tune for the fidelity, penalty, and regularization terms involved as well as the convergence rate of ADMM \citep{boyd2011distributed}. The VMD is an effective segmentation model, which has been demonstrated to work well on a broad variety of images. However, for our purpose, it presents two major inconveniences: (1) the large number of parameters to be tuned, which can become time consuming when trying to achieve the best texture segmentation result, and (2) the explicit number of active modes required in the decomposition, which is restrictive.

An alternative method is the empirical wavelet transform (EWT) recently proposed by \cite{G1}. It is very well adapted to patterns specific to STM image textures, and it automatically finds the number of modes while requiring very few parameters to tune. Using EWT, we propose a texture segmentation algorithm adapted to microscopy image textures, which consists of three steps: (1) perform the EWT on the texture component, (2) construct a feature matrix based on the processed EWT coefficients, and (3) apply a clustering algorithm to the feature matrix in order to obtain the final segmentation. Detailed below, the main contributions in the EWT algorithm are the improvement of boundary detection and partitioning the Fourier spectrum and the selection of texture features based on their local energy. 

\subsection{The Empirical Wavelet Transform}

The EWT was originally proposed as a signal decomposition method that detects and separates the signals' principal harmonic modes. The principal modes are modeled as amplitude modulated-frequency modulated (AM-FM) signals with compact support in the Fourier domain \citep{G1}. The EWT consists of two steps: 
(1) it partitions the Fourier spectrum into $N$ supports and it builds in the Fourier domain a filter bank, where each wavelet filter corresponds to a support,
and (2) it filters the input signal with the obtained filter bank to produce the different components.  The filter bank consists of $N$ wavelet filters: one low-pass  filter corresponding to the approximation component and $N-1$ bandpass filters corresponding to the details components.

\begin{figure}[!t]
	\centering
	\includegraphics[scale=.25]{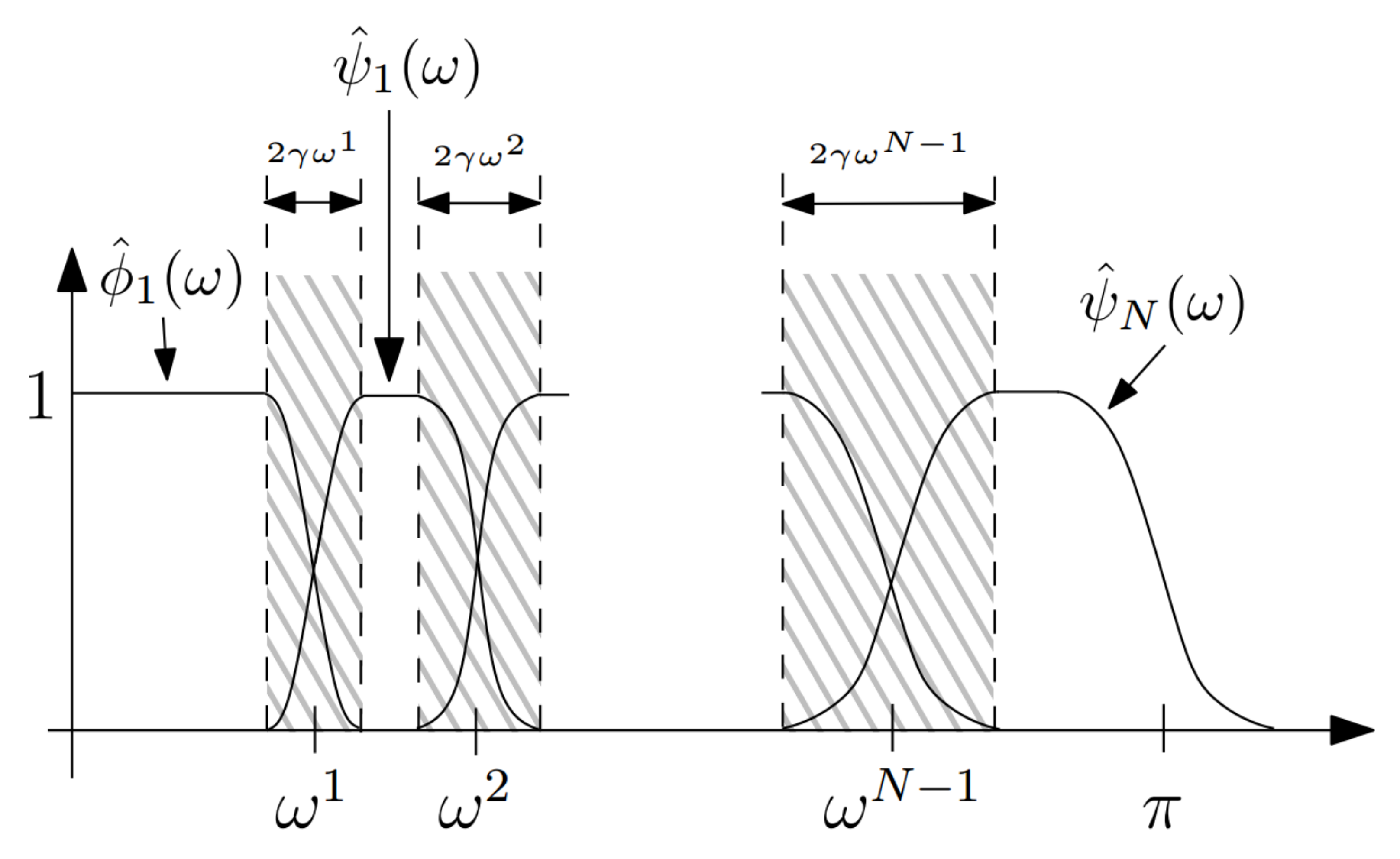}
	\caption{Partitioning of the Fourier spectrum of a 1D signal}
	\label{fig:filterbank}
\end{figure}

The partitioning of the Fourier spectrum is as important as building the adaptive wavelets since it provides information about the principal harmonic components. Several approaches to perform the boundaries detection in the Fourier domain were investigated by \cite{G1,SSHS,GTO}. In particular, \cite{SSHS} proposed a fully automatic  algorithm based on a combination of a scale-space representation and Otsu's\\ method. The advantage of this approach is in that it automatically finds the number $N$ of expected  modes as well as detects the boundaries of the Fourier supports. We used this approach in all experiments. 

Assume that the Fourier spectrum is partitioned into $N$ contiguous segments with boundaries
$\{\omega_n\}_{n=0}^N$, where $\omega_0 = 0$ and $\omega_N = \pi$ (see  Figure~\ref{fig:filterbank}). 
Then, based on Meyer's wavelet formulation, we construct a filter bank of wavelets
\[ \{\phi_1(x), \{\psi_n(x)\}_{n=1}^{N-1}\},\]
on the corresponding segments $\left[\omega_{n-1}, \omega_n\right]$.
 The Fourier\\ transform (denoted as $\mathcal{F}_1$ in the 1D case) 
of the empirical scaling function is given by
\begin{align}\label{eq:lowpass}
    \mathcal{F}_1 (\phi_1) (\omega) =
    \begin{cases}
    	1 \qquad \text{if } |\omega| \leq (1-\gamma)\omega_1\\
   	 \cos{\left[\frac{\pi}{2} \mathcal{B} \left(\frac{1}{2\gamma \omega_1} \left(|\omega| - 
	 (1-\gamma)\omega_1 \right) \right) \right ]}\\ 
	 \enspace \qquad \text{if } (1-\gamma) \omega_1 \leq |\omega| \leq (1+\gamma) \omega_1 \\
    	0 \qquad \text{if otherwise}
    \end{cases},
\end{align}
whereas the Fourier transforms of the empirical wavelets are given by
\begin{align}\label{eq:bandpass_scale}\nonumber
    &\mathcal{F}_1(\psi_n)(\omega) = \\
	&\begin{cases}
		1 \qquad \text{if } (1 + \gamma)\omega_n \leq |\omega| \leq (1-\gamma)\omega_{n+1}\\
		\cos{\left[\frac{\pi}{2}\mathcal{B}\left(\frac{1}{2\gamma \omega_{n+1}}\left( |\omega| - 
		(1-\gamma)(\omega_{n+1}\right)\right)\right]}\\
		\enspace \qquad \text{if }(1-\gamma)\omega_{n+1} \leq |\omega| \leq (1+\gamma) \omega_{n+1} \\
		\sin{\left[\frac{\pi}{2} \mathcal{B} \left(\frac{1}{2 \gamma \omega_n}\left (|\omega| - 
		(1-\gamma)\omega_n\right)\right)\right]} \\
		\enspace \qquad \text{if } (1-\gamma)\omega_n \leq |\omega| \leq (1+\gamma) \omega_n \\
		0 \qquad \text{if otherwise}
	\end{cases},
\end{align}
for $n = 1, \ldots, N$.  The function $\mathcal{B}$ is an arbitrary $\mathcal{C}^k([0,1])$ function satisfying the properties that $\mathcal{B}(t) = 0$ if $t \leq 0$,  $\mathcal{B}(t) = 1$ if $t \geq 1$, and $\mathcal{B}(t) + \mathcal{B}(1-t) = 1$ and $\mathcal{B}(t) \in (0,1)$ for all $t\in [0,1]$. The parameter $\gamma$ is chosen to ensure  that two consecutive transition areas (shown as dashed areas in Figure~\ref{fig:filterbank}) do not overlap. As shown by \cite{G1}, a proper selection of $\gamma$  guarantees that the filter bank\\ $\{\phi_1, \{\psi_n\}_{n=1}^{N-1}\}$ is a tight frame in $L^2(\mathbb{R})$. Then, the EWT is defined in the same way as the classical Wavelet Transform. For the signal function $f$, the details coefficients are given as 
 \begin{equation} \label{eq:waveletdetail}
	 \mathcal{W}_f^{\mathcal{E}}(n,x) = 
	 \mathcal{F}_1^* \left(\mathcal{F}_1(f)(\omega)\overline{\mathcal{F}_1(\psi_{n})(\omega)} \right)(x),
 \end{equation}
 and the approximation coefficient as
 \begin{equation} \label{eq:waveletapproximation}
 	\mathcal{W}_f^{\mathcal{E}}(0,x) = 
	\mathcal{F}_1^* \left(\mathcal{F}_1(f)(\omega) \overline{\mathcal{F}_1(\phi_1)(\omega)}\right)(x),
 \end{equation}
 where $\mathcal F_1^*$ stands for the inverse 1D Fourier Transform.

The EWT was later generalized to 2D images for various kinds of wavelet transform, specifically tensor wavelets, Littlewood-Paley wavelet transform, the ridgelet transform, and  the  curvelet transform \citep{GTO}. 

\subsection{The Empirical Curvelet Transform}

Textures in STM images can be seen as oscillatory patterns with multiple orientations.  Among all of the above mentioned variants, curvelets are wavelets that take into account various orientations \citep{candes2005continuous}. Therefore, its empirical counterpart, the Empirical Cur\-velet Transform (ECT), is  the most appropriate adaptation of EWT to partition texture images. Similar to the EWT, the ECT builds a filter bank in the Fourier domain where each filter has its support on a polar wedge.

As shown in Figure \ref{fig:spec.b}, the Fourier domain is partitioned in this case into a disk centered at the origin, which contains the low frequencies, and concentric annuli and angular sectors (polar wedges), which contain high frequencies. In order to build the filter bank adaptively, the ECT needs to detect the boundaries of the polar wedges empirically, which correspond to finding the scales for the angle and the radii.
Following \cite{GTO}, this step can be achieved by considering the pseudo-polar Fourier transform  \citep{averbuch2006fast,averbuch2008framework} and performing the previously described 1D detection to 1D spectra corresponding to averaging with respect to the frequency magnitude  and orientation,  respectively. \cite{GTO} proposed three different cases: (1) scales and angles are detected independently, (2) scales are detected first and angles are  detected per each scale, and (3) angles are detected first and scales are detected per each angular sector. In this work, we will consider the second option because STM images contain texture patterns with different main orientations having varying frequency magnitudes. 

In the following, we will denote $\omega=(\omega_{x_1},\omega_{x_2})$ the frequency coordinates in the Fourier plane, $|\omega|$ its magnitude and $\theta$ its angle. Otsu's boundary detection method will then provide $N_{\theta}$ number of angles and $N_s^m$ number of scales per each angular sector $m=1,\ldots, N_{\theta}$. This is  equivalent to obtaining the set of angular  boundaries $\{\theta_m\}_{m=1}^{N_{\theta}}$ and the set of scale boundaries  $\{\omega_n^m\}_{n=1}^{N_s^m}$ per angular sector for $m=1,\ldots,N_{\theta}$. Note that $\omega_1^1 = \ldots= \omega_1^{N_\theta}$ because all together they form the disk centered at the origin of the Fourier domain. The corresponding curvelet filters are then defined in the Fourier 
domain (we denote $\mathcal{F}_2$ the 2D Fourier transform) in the following way. The purely radial lowpass filter $\phi_1$ is given by
\begin{equation}\label{eq:lowpasscurvelet}
    \mathcal{F}_2(\phi_1)(\omega, \theta)= 
    \begin{cases}
    	1 \qquad \text{if}\quad |\omega| \leq (1-\gamma)\omega_1,\\
    	\cos{\left[\frac{\pi}{2} \mathcal{B} \left(\frac{1}{2\gamma \omega_1} \left(|\omega| - 
	(1-\gamma)\omega_1 \right) \right) \right ]}\\ 
	\enspace \qquad \text{if}\quad (1-\gamma) \omega_1 \leq |\omega| \leq (1+\gamma) \omega_1, \\
    	0 \qquad \text{if otherwise},
    \end{cases}
\end{equation} 
where $\omega_1 = \omega_1^1 = \ldots = \omega_1^{N_\theta}$. 
The polar curvelet associated to the polar wedge $I_n^m : = [\theta_m,\theta_{m+1}]\times[\omega_n^m,\omega_{n+1}^m]$ can be written as
\begin{equation}\label{eq:polarwedge}
	\mathcal{F}_2(\psi_{m,n})(\omega, \theta) =  W_n^m(|\omega|)V_m(\theta),
\end{equation}
where the radial window $W_n^m$ is 
\begin{equation}\label{eq:radial1}
	W_n^m(|\omega|) = 
	\begin{cases}
		1 \qquad\text{if}\quad (1 + \gamma)\omega_n^m \leq |\omega| \leq (1-\gamma)\omega_{n+1}^m,\\
		\cos\bigg[\frac{\pi}{2} \mathcal{B}\left(\frac{1}{2\gamma\omega_{n+1}^m}\left(|\omega|-
		(1-\gamma)\omega_{n+1}^m\right)\right)\bigg] \\
		\qquad\;\;\text{if}\quad (1-\gamma)\omega_{n+1}^m \leq |\omega| \leq (1+\gamma) \omega_{n+1}^m, \\
		\sin\bigg[\frac{\pi}{2} \mathcal{B} \left(\frac{1}{2 \gamma \omega_n^m}\left (|\omega|- 
		(1-\gamma)\omega_n^m\right)\right)\bigg] \\
		\qquad\;\;\text{if}\quad (1-\gamma)\omega_n^m \leq |\omega| \leq (1+\gamma) \omega_n^m, \\
		0 \qquad\text{if otherwise},
	\end{cases}
\end{equation}
for $n \neq N_s^m-1$ and 
\begin{align}\notag
	W	&_{N_s^m-1}^m (|\omega|)=\\ 
		&\begin{cases}
			1 \qquad \text{if}\quad (1+\gamma)\omega_{N_s^m-1}^m \leq |\omega|, \\
			\sin{\left[\frac{\pi}{2} \mathcal{B} \left(\frac{1}{2 \gamma \omega_{N_s^m-1}^m} 
			\left(|\omega| - (1-\gamma)\omega_{N_s^m-1}^m\right) \right)\right]}\\
			\enspace \qquad \text{if}\quad (1-\gamma)\omega_{N_s^m-1}^m \leq
			 |\omega| \leq (1+\gamma) \omega_{N_s^m-1}^m,\\
			0 \qquad \text{if otherwise},
		\end{cases}
\end{align}
for $n = N_s^m-1$, while the angular window $V_m$ is 
\begin{align}\label{eq:angle}
	V_m(\theta)=
		\begin{cases}
			1 \qquad \text{if}\quad \theta_m + \Delta \theta \leq \theta \leq \theta_{m+1} - \Delta \theta, \\
			\cos{\left[\frac{\pi}{2} \mathcal{B} \left(\frac{1}{2\Delta \theta} 
			\left(\theta - \theta_{m+1}+\Delta \theta \right) \right) \right]} \\
			\qquad\;\; \text{if}\quad \theta_{m+1} - \Delta \theta \leq \theta \leq \theta_{m+1} + \Delta \theta, \\
			\sin{\left[\frac{\pi}{2} \mathcal{B} \left( \frac{1}{2\Delta \theta} 
			\left(\theta - \theta_m + \Delta \theta\right)\right)\right]} \\
			\qquad\;\; \text{if}\quad \theta_m - \Delta \theta \leq \theta \leq \theta_m + \Delta \theta, \\
			0 \qquad \text{if otherwise}. 
	\end{cases}
\end{align}
The parameters $\gamma$ and $\Delta \theta$ are chosen in order to guarantee the tight frame property (see \cite{GTO} for details). This leads to the construction of the filter  bank of empirical curvelets 
\begin{equation}\label{eq:filterbank}
	\left\{ \phi_1(x), \{\psi_{m,n}(x)\}_{\substack{m=1,\ldots,N_{\theta}\\n=1,\ldots,N_{s}^m-1}}\right\}.
\end{equation}
 
From \eqref{eq:filterbank}, the empirical curvelet transform of the texture component $v$ is given by the detail coefficients as
 \begin{equation} \label{eq:waveletdetail2}
	 \mathcal{W}_{v}^{\mathcal{EC}}(m,n,x) = 
	 \mathcal{F}_2^* \left(\mathcal{F}_2(v)
	 	\overline{\mathcal{F}_2(\psi_{m,n})} \right)(x),
\end{equation}
and the approximation coefficients as
\begin{equation} \label{eq:waveletapproximation2}
 	\mathcal{W}_{v}^{\mathcal{EC}}(0,0,x) = 
	\mathcal{F}_2^* \left(\mathcal{F}_2(v)
		\overline{\mathcal{F}_2(\phi_1)}\right)(x),
\end{equation}
where $\mathcal F_2^*$ stands for the inverse 2D Fourier Transform.
We can reconstruct the image $v(x)$ by the inverse transform:
\begin{align}
 	\begin{split}
     	v(x) = \mathcal{F}^*_2 \bigg( 
		\mathcal{F}_2 \left(\mathcal{W}_{v}^{\mathcal{EC}}(0,0,\cdot)\right)
		\mathcal{F}_2 (\phi_1) \\			
		+ \sum_{m=1}^{N_\theta} \sum_{n=1}^{N_s^m-1}  
		\mathcal{F}_2 \left(\mathcal{W}_{v}^{\mathcal{EC}}(m,n,\cdot)\right) 
		\mathcal{F}_2 (\psi_{m,n})\bigg)(x).
	\end{split}
\end{align}
 
Each empirical curve\-let subband $\mathcal{W}_{v}^{\mathcal{EC}}(m,n,\cdot)$ contains either some textural pattern or noise. In many cases, some of these curvelets may  have extremely weak oscillatory patterns or pure noise, deeming them useless in our texture analysis. Therefore, we propose a pre-processing step and a post-processing step in order to obtain only empirical curve\-lets with meaningful information. 

\subsection{Improved Boundaries Detection}

In this section, we propose  modifications in the boundary detection algorithm in order to reduce the number of irrelevant polar wedges. Since the detection is performed in the pseudo-polar Fourier domain, the following processing will also be performed in the pseudo-polar domain. Hereafter, we will denote $\mathcal{F}_P(v)$ as the pseudo-polar Fourier transform of the input texture image $v$.

\begin{figure}[!t]
 	\centering
  	\begin{subfigure}[t]{0.5\columnwidth}
  		\centering
  		\includegraphics[width=0.9\columnwidth]{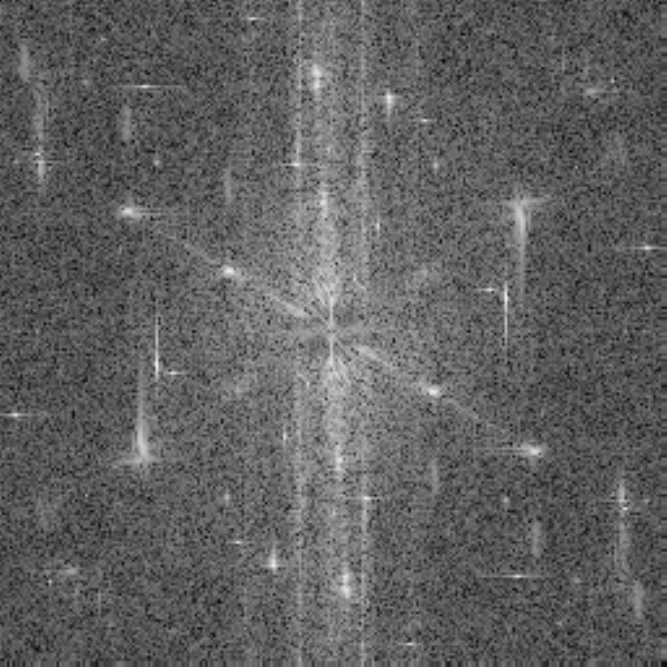}
 		\caption{Original spectrum}\label{fig:spec.a}
  	\end{subfigure}%
  	\begin{subfigure}[t]{0.5\columnwidth}
  		\centering
  		\includegraphics[width=0.9\columnwidth]{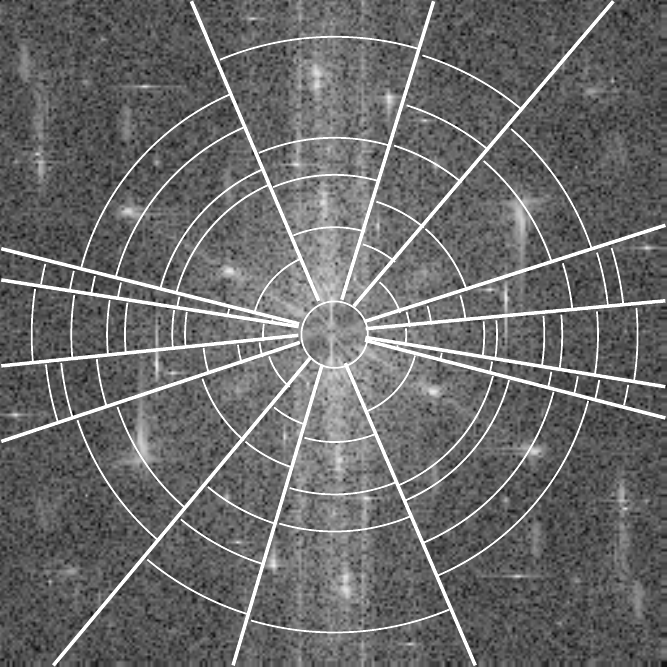}
  		\caption{Detected boundaries on the original spectrum}\label{fig:spec.b}
  		\end{subfigure}%
 	\vskip 1mm
  	 \begin{subfigure}[t]{0.5\columnwidth}
  		\centering
  		\includegraphics[width=0.9\columnwidth]{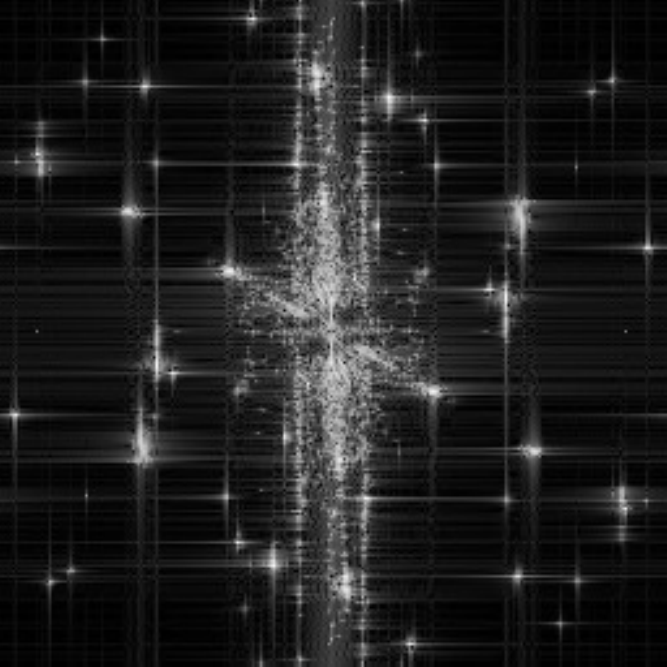}
  		\caption{Thresholded spectrum}\label{fig:spec.c}
  	\end{subfigure}%
  	\begin{subfigure}[t]{0.5\columnwidth}
 		\centering
 		\includegraphics[width=0.9\columnwidth]{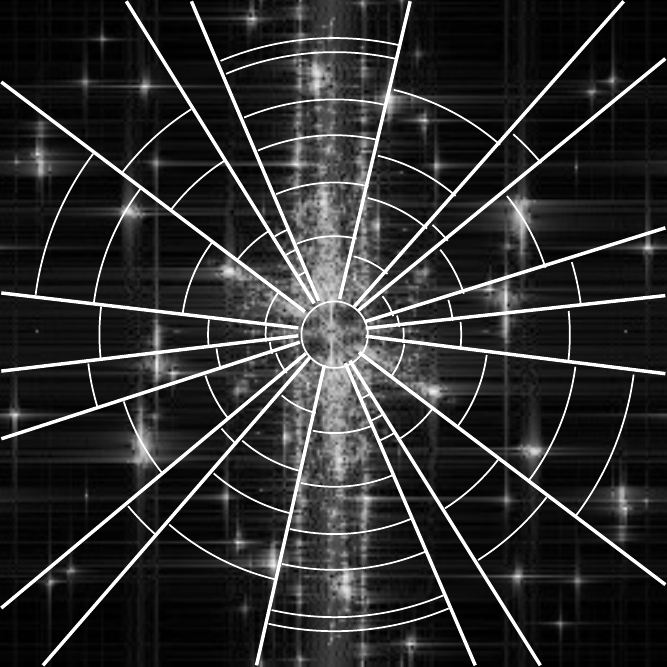}
  		\caption{Detected boundaries on the thresholded spectrum} \label{fig:spec.d}
  	\end{subfigure}%
  	\caption{Comparison of partitions obtained from the original spectrum and its thresholded version. The spectrum shown is reproduced and modified from \cite{guttentag2016hexagons} with permission. STM image spectrum is copyright from American Chemical Society}
  	\label{fig:specthr}
\end{figure}

Our first improvement is a preprocessing step before the actual detection. Since the goal is to isolate clusters of high magnitude frequencies in $\mathcal{F}_P(v)$ we suggest thresholding the Fourier coefficients in order to remove all frequencies that are not relevant. Let $T$ be the hard-\\thresholding operator defined as
\begin{equation}
 	T(a, \tau) = 
	\begin{cases}
 		0 & \text{if}\quad |a| \leq \tau \\
 		a & \text{if}\quad |a| > \tau
	\end{cases}.
\end{equation}
The preprocessing step consists of performing the detection on $T(\mathcal{F}_P(v),\tau)$ instead of $\mathcal{F}_P(v)$. The threshold $\tau$ is chosen as a  certain  percentile of the magnitude of the Fourier coefficients: we create a vector whose entries are the magnitudes of the Fourier coefficients listed in increasing  order and we set $\tau$ to be the value of the $n\times p-th$ entry of the vector, where $n$ is the length of the vector and $p$ the specified percentile (i.e. $p\in(0,1)$).  Figure~\ref{fig:specthr}  illustrates the effects of performing such thresholding. Figures~\ref{fig:spec.a} and \ref{fig:spec.c} show the original spectrum as well as its thresholded version (using a  92 percentile). Figures~\ref{fig:spec.b} and \ref{fig:spec.d} provide the corresponding detected partitions. We note that the partition obtained from the thresholded  spectrum  provides a separate angular sector for the high magnitude clusters near the top right corner while these clusters are associated with another angular sector in the partition  obtained from the original spectrum.

Unfortunately, sometimes some meaningless polar wed\-ges are detected in each angular sector. Our second improvement aims at removing these useless polar wedges by merging them  with their neighbors. Suppose that $\{\theta_m\}_{m=1}^{N_\theta}$ and $\left\{\{\omega_n^m\}_{n=1}^{N_{s}^m}\right\}_{m=1}^{N_\theta}$ are the set of angles and scales detected  on  $T(\mathcal{F}_P(v), \tau)$. In the pseudo-polar domain, a polar wedge corresponds to the rectangle $I_n^m$, whose area we denote by $A_n^m=area(I_n^m)=(\omega_{n+1}^m-\omega_n^m)(\theta_{m+1}-\theta_m)$.  We define the information density per polar wedge by
\begin{equation}\label{eq:density}
	M_n^m=\frac{\|T(\mathcal{F}_P(v), \tau) \mathbbm{1}_{I_n^m}\|_1}{A_n^m},
\end{equation}
where $\mathbbm{1}_{I_n^m}$ is the characteristic function over the domain $I_n^m$. As a reference, we use the largest density $\tilde{M} = \max_{m,n}{M_n^m}$ and we threshold at a certain fraction $\eta = 0.1$ as follows. For a given angular sector, we start from the polar wedge that is the farthest from the origin until we reach the one closer to the origin. If $M_n^m\leq \eta\tilde{M}$, then the $I_n^m$ polar wedge is irrelevant and we merge it with $I_{n-1}^m$, i.e we remove $\omega_{n-1}^m$ from the list  $\left\{\{\omega_n^m\}_{n=1}^{N_{s}^m}\right\}_{m=1}^{N_\theta}$ and update $N_s^m:=N_s^m-1$. If $M_n^m>\eta\tilde{M}$, we move to the next polar wedge $I_{n-1}^m$ and repeat the  procedure. 
The corresponding merging algorithm is summarized in Algorithm~\ref{alg:merge}.

\begin{algorithm}[th]
\caption{Merging Curvelet partition}\smallskip
\KwIn{Thresholded Fourier domain $T(\mathcal{F}_P(v),\tau)$, original boundaries $\{\theta_m\}_{m=1}^{N_\theta}$ and  
$\left\{\{\omega_n^m\}_{n=1}^{N_{s}^m}\right\}_{m=1}^{N_\theta}$}
  \begin{algorithmic}[1]
    \STATE Compute $M_n^m$ for $m=1\ldots N_\theta$ and $n=1\ldots N_s^m$ according to \eqref{eq:density}.
    \STATE Compute $\tilde{M} = \max_{m,n}M_n^m$.
    \FOR{$i = 1$ to $N_\theta$}
       \STATE Set $j := N_s^i$.
       \WHILE{$j \geq 3$}
           \IF{$M_{j-1}^i< 0.10\tilde{M}$}
	  \STATE Remove $\omega_{j-1}^i$ from $\left\{\{\omega_n^m\}_{n=1}^{N_{s}^m}\right\}_{m=1}^{N_\theta}$.
	  \STATE Set $N_s^i := N_s^i -1$.
	  \ENDIF
	  \STATE $j := j-1$.
      \ENDWHILE
    \ENDFOR
\end{algorithmic}
\KwOut{Updated boundaries $\left \{\{\theta_m\}_{m=1}^{N_\theta}, \left\{\{\omega_n^m\}_{n=1}^{N_{s}^m}\right\}_{m=1}^{N_\theta} \right \}$.}
\label{alg:merge}
\end{algorithm}

\begin{figure*}[!t]
	\centering
  	\begin{subfigure}[t]{0.33\textwidth}
  		\centering
  		\includegraphics[width=0.9\columnwidth]{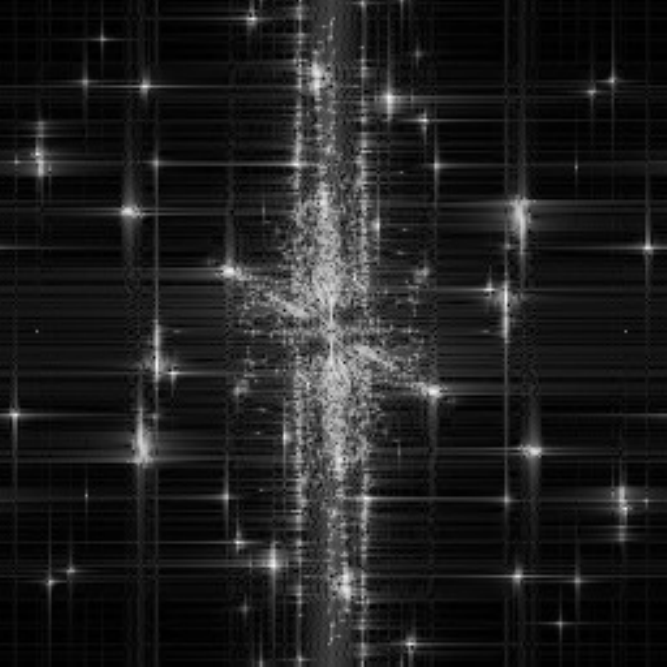}
 		\caption{Thresholded spectrum}\label{fig:part.a}
  	\end{subfigure}%
  	\begin{subfigure}[t]{0.33\textwidth}
  		\centering
  		\includegraphics[width=0.9\columnwidth]{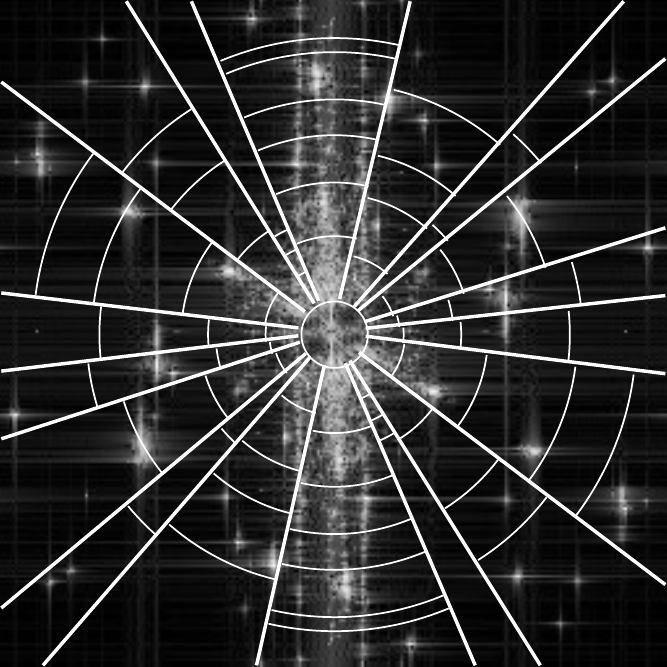}
  		\caption{Partition before merging}\label{fig:part.b}
  	\end{subfigure}%
  	\begin{subfigure}[t]{0.33\textwidth}
  		\centering
  		\includegraphics[width=0.9\columnwidth]{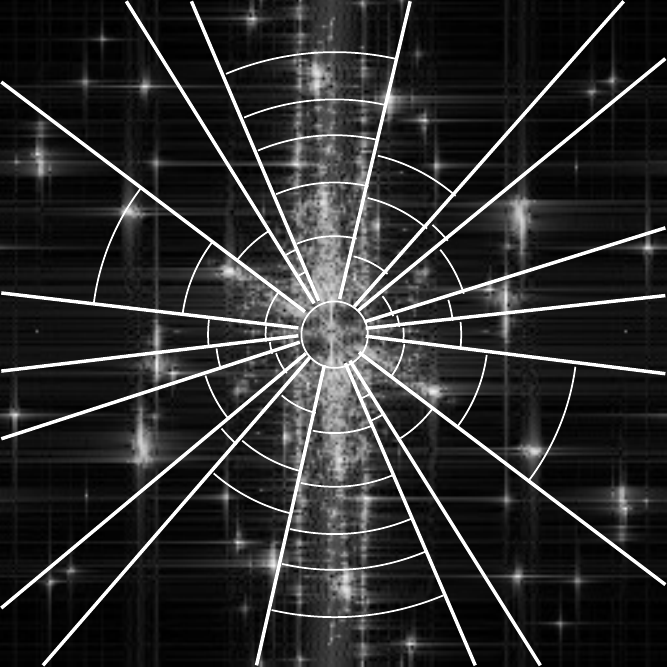}
  		\caption{Partition after merging}
  		\label{fig:part.c}
  	\end{subfigure}%
	\caption{Influence of the merging algorithm on the obtained partitions. The spectrum shown is reproduced and modified from \cite{guttentag2016hexagons} with permission. STM image spectrum is copyright American Chemical Society}
	\label{fig:merging}
\end{figure*}

Partitions obtained before and after merging are illustrated in Figure~\ref{fig:merging}. It is easy to see that some polar wedges having less information were merged to form a  new  set of polar wedges having a minimum amount of useful information (see for instance the most vertical angular sector, the thin outer wedge no longer exist in the updated 
partition).

Based on the updated partition, the empirical curvelet filter bank is then constructed accordingly to \eqref{eq:lowpasscurvelet} and \eqref{eq:polarwedge}. The full modified empirical curvelets transform is summarized in Algorithm~\ref{alg:MECT}.

\begin{algorithm}[th]
\caption{Modified Empirical Curvelet Transform}\smallskip
\KwIn{Image $v(x)$, Threshold Value $\tau$}
  \begin{algorithmic}[1]
    \STATE Compute the Pseudo-Polar FFT $\mathcal{F}_P(v)$.
    \STATE Threshold the Fourier coefficients to obtain $T(\mathcal{F}_P(v), \tau)$.
    \STATE Detect the original partition 
    	$ \left \{\{\theta_m\}_{m=1}^{N_\theta}, \left\{\{\omega_n^m\}_{n=1}^{N_{s}^m}\right\}_{m=1}^{N_\theta} \right \}$ 
	using Otsu's Method as described in \cite{GTO,SSHS}.
    \STATE Compute the updated set 
    	$ \left \{\{\theta_m\}_{m=1}^{N_\theta}, \left\{\{\omega_n^m\}_{n=1}^{N_{s}^m}\right\}_{m=1}^{N_\theta} \right \}$ 
	by applying the merging algorithm Algorithm~\ref{alg:merge}.
    \STATE Construct the corresponding curvelet filter bank 
    	$ \mathscr{B}^{\mathcal{MEC}} = \left\{\phi_1(x),
		 \{\psi_{m,n}(x)\}_{\substack{m=1,\ldots,N_{\theta}\\n=1,\ldots,N_{s}^m-1}}\right\}$ 
		 accordingly to \eqref{eq:lowpasscurvelet}-\eqref{eq:polarwedge}.
    \STATE Filter $v(x)$ using \eqref{eq:waveletdetail2}-\eqref{eq:waveletapproximation2} to obtain 
    	$ \mathcal{W}_{v}^{\mathcal{MEC}}= \left\{\mathcal{W}_{v}^{\mathcal{MEC}}(0,0,x), 
	\{ \mathcal{W}_{v}^{\mathcal{MEC}}(m,n,x)\}_{\substack{m=1,\ldots,N_{\theta}\\n=1,\ldots,N_s^m-1}} \right\}$
  \end{algorithmic}
\KwOut{Spectrum boundaries $\left \{\{\theta_m\}_{m=1}^{N_\theta}, 
	\left\{\{\omega_n^m\}_{n=1}^{N_{s}^m}\right\}_{m=1}^{N_\theta} \right \}$,
	empirical curvelet filter bank $\mathscr{B}^{\mathcal{MEC}}$,
	empirical curvelets coefficients $\mathcal{W}_{v}^{\mathcal{MEC}}$. }
\label{alg:MECT}
\end{algorithm}
As shown in \cite{GTO}, the tight frame property depends only on the construction process of the\\ curvelet filters themselves and does not depend on how the supports detection is done. Therefore, the original proof remains valid even within the framework of our modified support detection algorithm, implying that the tight frame property is preserved.

\subsection{Texture Features}
After applying our modified empirical curvelet transform to the texture component of the image, we finally construct the relevant information to characterize different textures. 
This information could be directly given by the empirical curvelets coefficients by reshaping each one into a vector corresponding to each pixel. However, this kind of feature vector  does  not have any inherent spatial information of the local neighborhood of its corresponding pixel. Instead, we consider computing the local \say{energy} of the curvelet coefficients. Then, we build a set of feature vectors associated to each pixel of the texture component. This set will form together a feature matrix whereby a clustering  algorithm can be applied to it. 

Define the local energy at $(m,n,x)$ as
\begin{align} \label{eq:L2energy}
    E_{m,n}^{\mathcal{MEC}}(x) = 
    \frac{\|\mathbbm{1}_{B(x,r_{m,n})}\mathcal{W}_{ v }^{\mathcal{MEC}}(m,n,\cdot)\|_2}{|B(x,r_{m,n})|}
\end{align}
where $B(x,r_{m,n})$ is a $(2r_{m,n} +1) \times (2r_{m,n}+1)$  neighborhood window centered at $x$. The radius $r_{m,n}$ is determined by the frequency of the texture pattern so that $E_{m,n}^{\mathcal{MEC}}(x)$ captures enough information around $x$. Hence, with $\omega_{n}^{m}$ related to frequency, we choose $r_{m,n} $ proportional to $\lceil \frac{1}{\omega_{n}^{m}}\rceil$ so that the radius relates to the period of the texture pattern. In practice, we set $r_{m,n} = \lceil \frac{\pi}{\omega_{n}^{m}}\rceil$. Note that for pixels near or at the border of the image, we apply symmetric padding to align with our assumption that texture patterns are periodic. 

After calculating \eqref{eq:L2energy} for each pixel for each empirical curvelet subband, we construct the feature matrix $\mathcal{D}$ by  casting all energy matrices $E_{m,n}^{\mathcal{MEC}}$ as vectors so that they form the columns of $\mathcal{D}$. Thus, $\mathcal{D}$ has the form
\begin{align}\label{eq:energymatrix}
	\mathcal{D}= \left(\begin{matrix}
    		\rvert & \ldots & \rvert &\rvert & \ldots & \rvert \\
    		E_{1,1}^{\mathcal{MEC}} & \ldots & E_{1, N_s^1-1}^{\mathcal{MEC}} 
		&  E_{2, 1}^{\mathcal{MEC}}& \ldots & E_{N_{\theta}, N_s^{N_\theta} -1}^{\mathcal{MEC}} \\
    		\rvert  & \ldots & \rvert &\rvert & \ldots & \rvert     
    	\end{matrix}\right).
\end{align}
Because texture patterns are locally periodic in an image, pixels  belonging to the same texture pattern should both have similar energies as defined by \eqref{eq:L2energy}. Hence, we could group the column vectors as belonging to the same class. We can then apply any clustering algorithm, such as $k$-means or multiclass MBO clustering \citep{garcia2014multiclass,Merkurjev}, on $\mathcal{D}$ to identify the texture patterns  of the image.

\section{Experimental Results}\label{sec:experiments}
In this section, we present results of our framework applied to the images in Figure~\ref{fig:STMimages}. All algorithms in this paper were implemented in MATLAB R2016a. The  codes and the results are available at\\ \url{https://github.com/kbui1993/Microscopy-Codes}.

In order to segment the images according to intensities or texture patterns, we first perform the cartoon+tex\-ture decomposition on each of them to obtain their cartoon  and texture components. In our experiments, we select $\sigma = 3$ in Algorithm~\ref{alg:NonLinear_CTD}. This parameter leads to more appealing segmentation results compared to  other values of $\sigma$. Moreover, $\sigma = 3$ is the minimum value for which humans could perceive region as textures \citep{buades2011cartoon}. 

After obtaining the cartoon and texture components for each image, we apply our proposed methods to each component. Our results are divided into two subsections. The cartoon  segmentation results are given in Section~\ref{subsec:cartoon} while the texture segmentation results are given in Section~\ref{subsec:texture}. 

\subsection{Cartoon Segmentation Results}\label{subsec:cartoon}

\begin{figure*}[!t]
\begin{center}
\begin{tabular}{m{1mm}>{\centering\arraybackslash}>{\centering\arraybackslash}m{0.2\textwidth}>{\centering\arraybackslash}m{0.2\textwidth}>{\centering\arraybackslash}m{
0.2\textwidth}>{\centering\arraybackslash}m{0.2\textwidth}}
\centering
& \textbf{original} & \textbf{cartoon} & \textbf{multiphase} & \textbf{local multiphase} \\
\textbf{(a)}& 
  \includegraphics[width=0.2\textwidth]{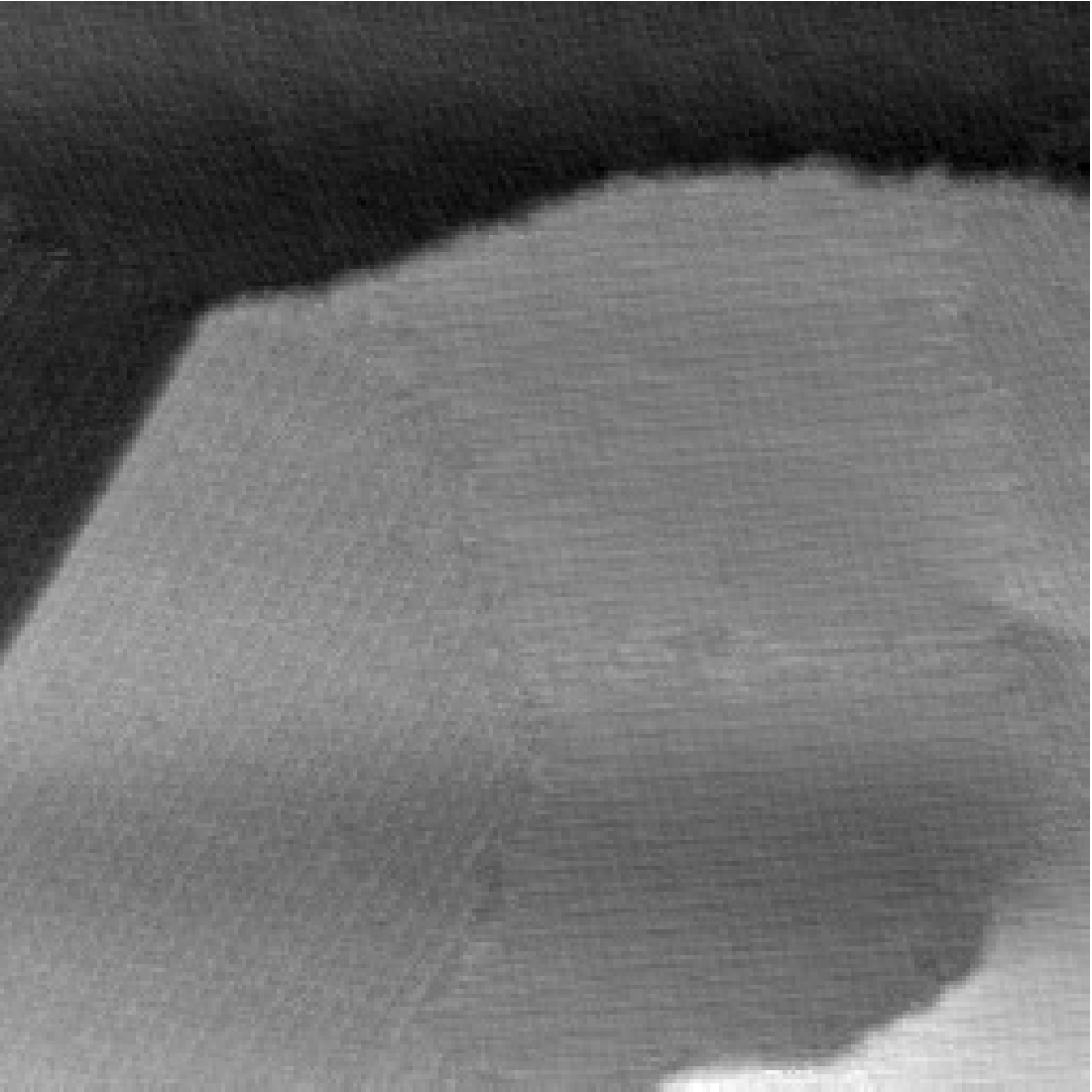} & 
  \includegraphics[width=0.2\textwidth]{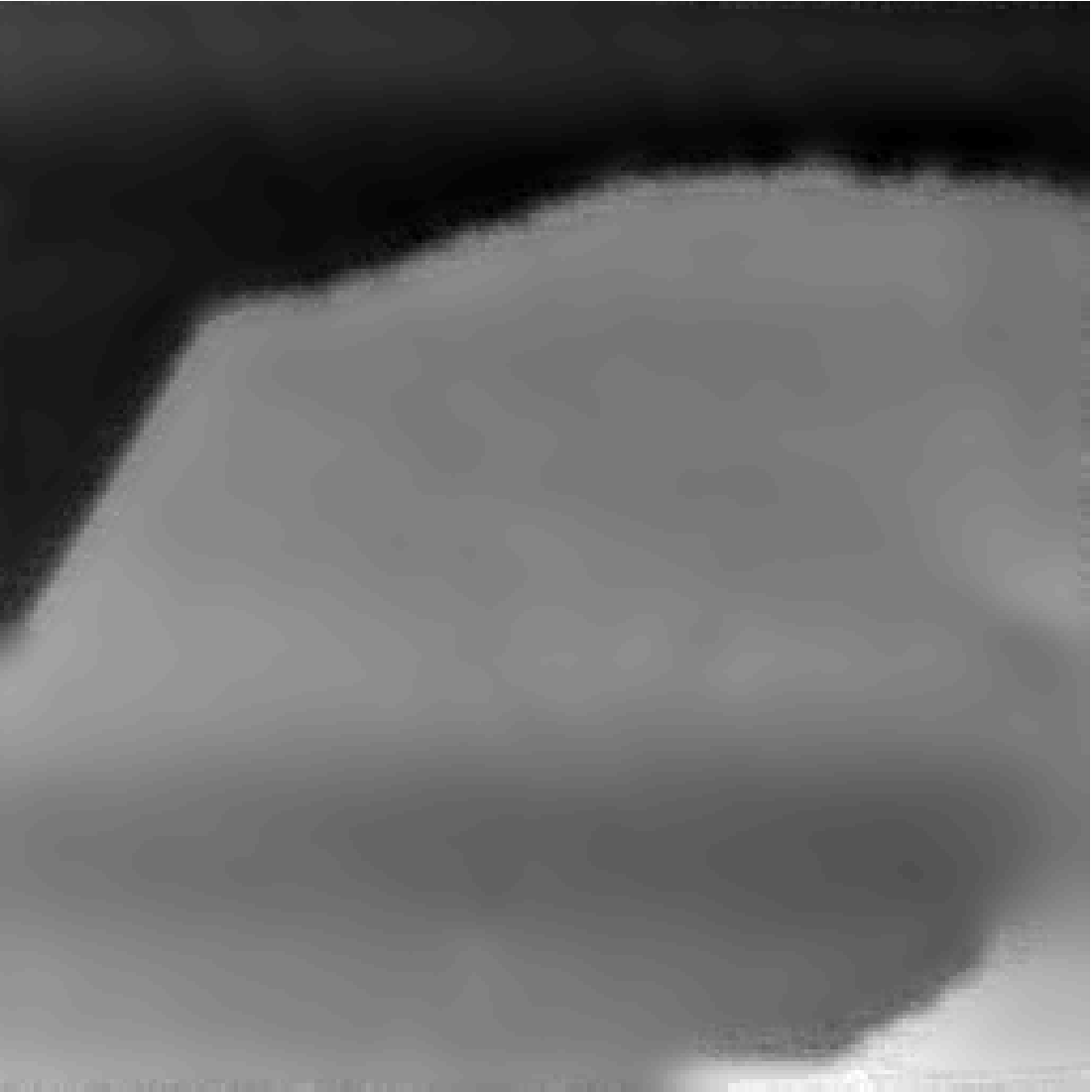}  &  
  \includegraphics[width=0.2\textwidth]{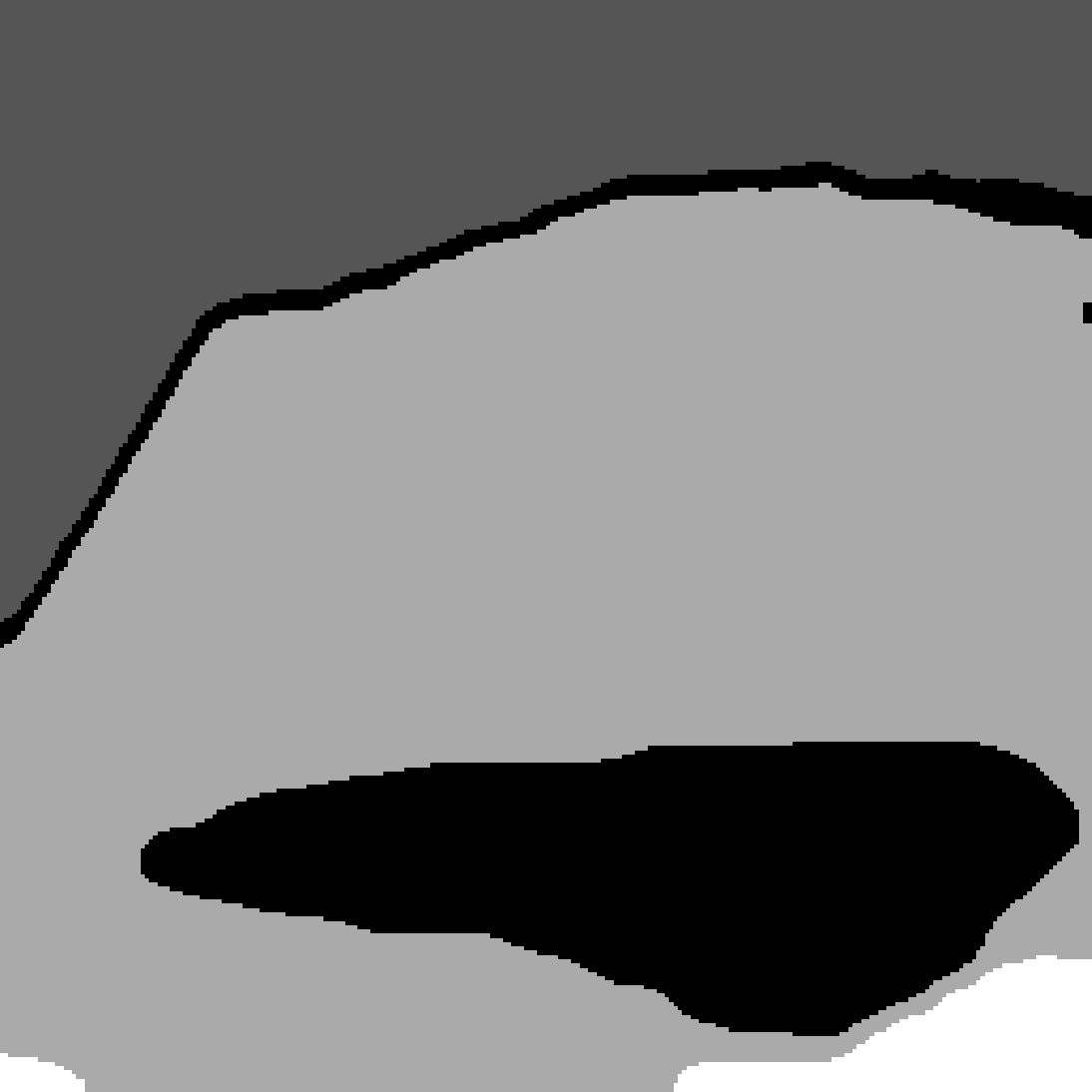} &
  \includegraphics[width=0.2\textwidth]{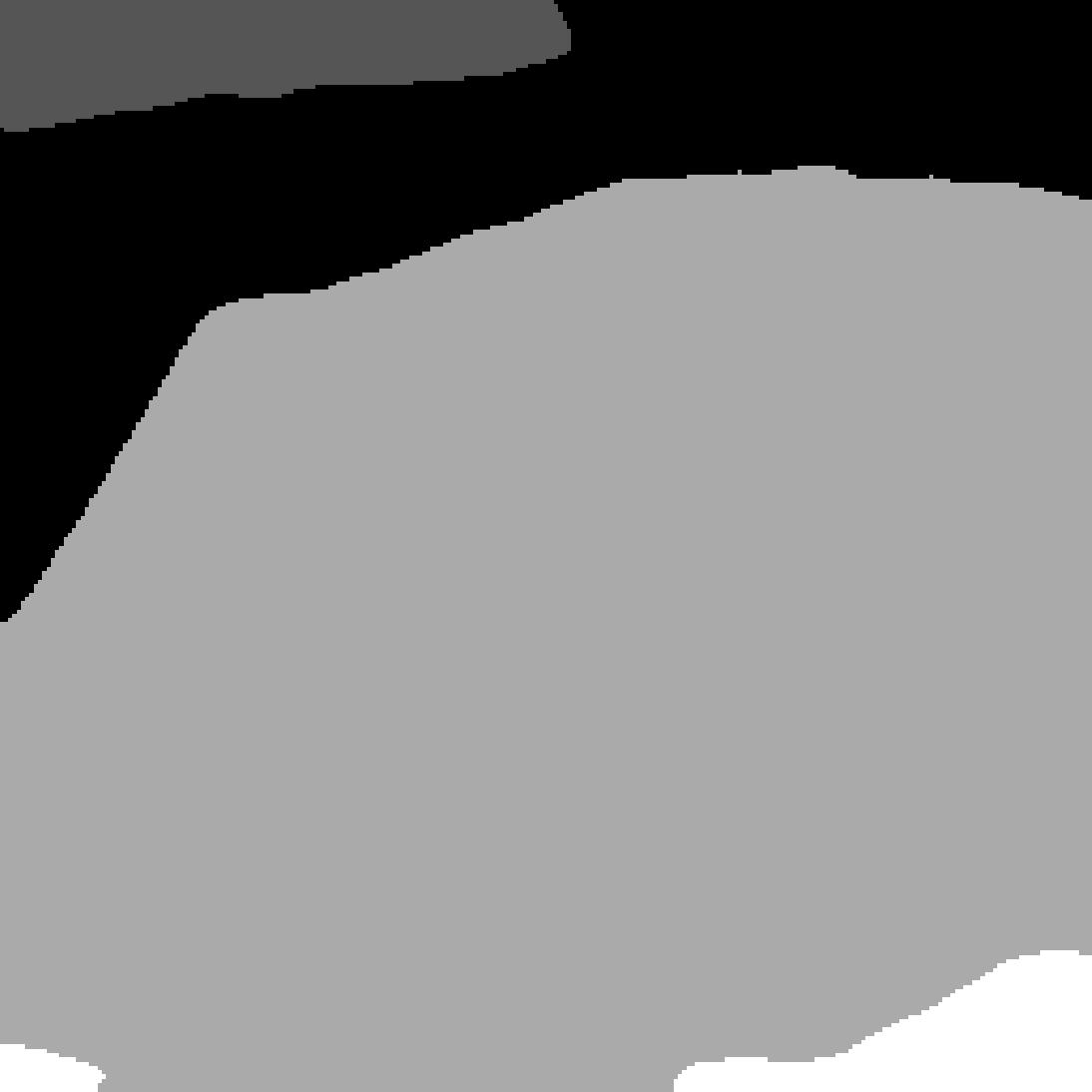} \\
\textbf{(b)}& 
  \includegraphics[width=0.2\textwidth]{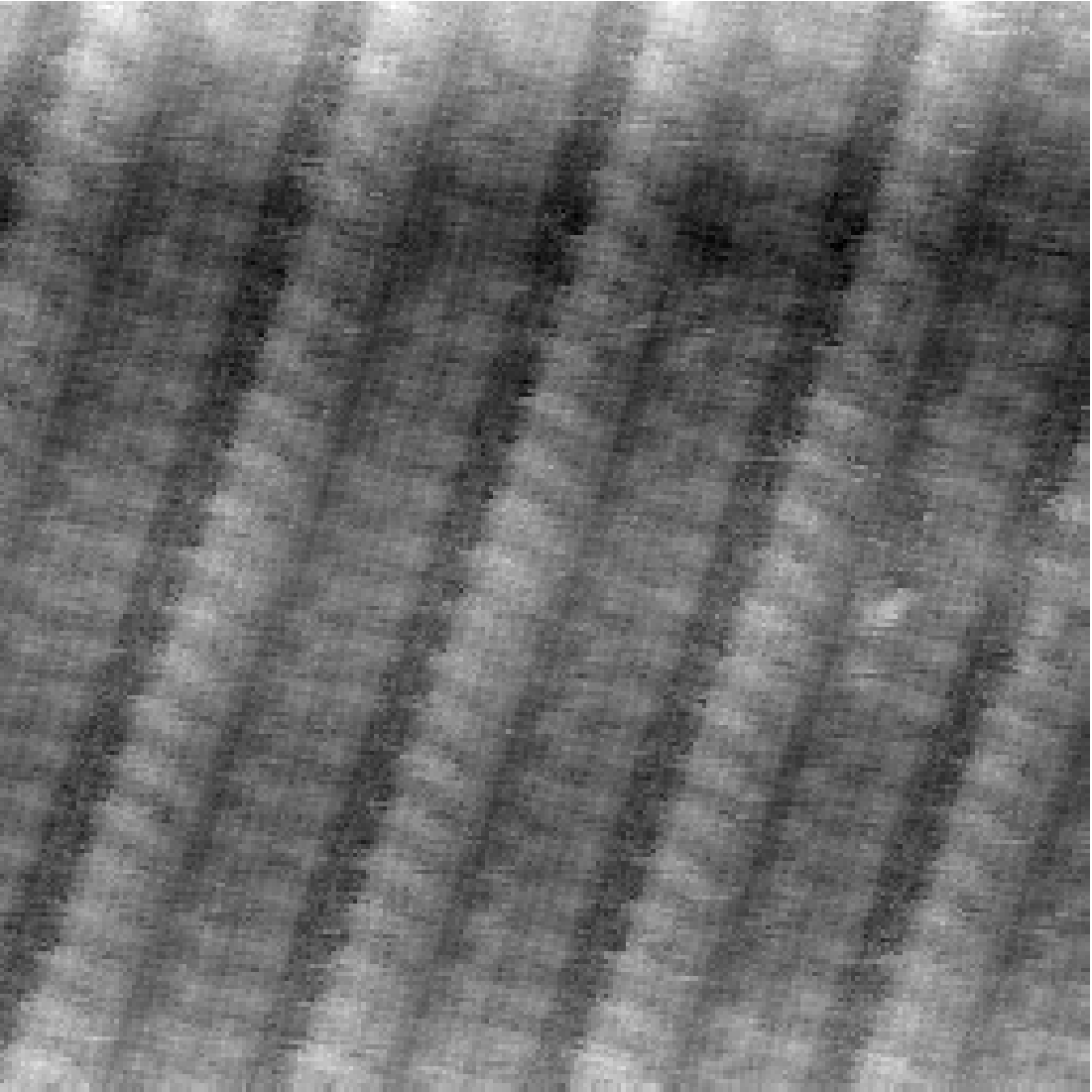} & 
  \includegraphics[width=0.2\textwidth]{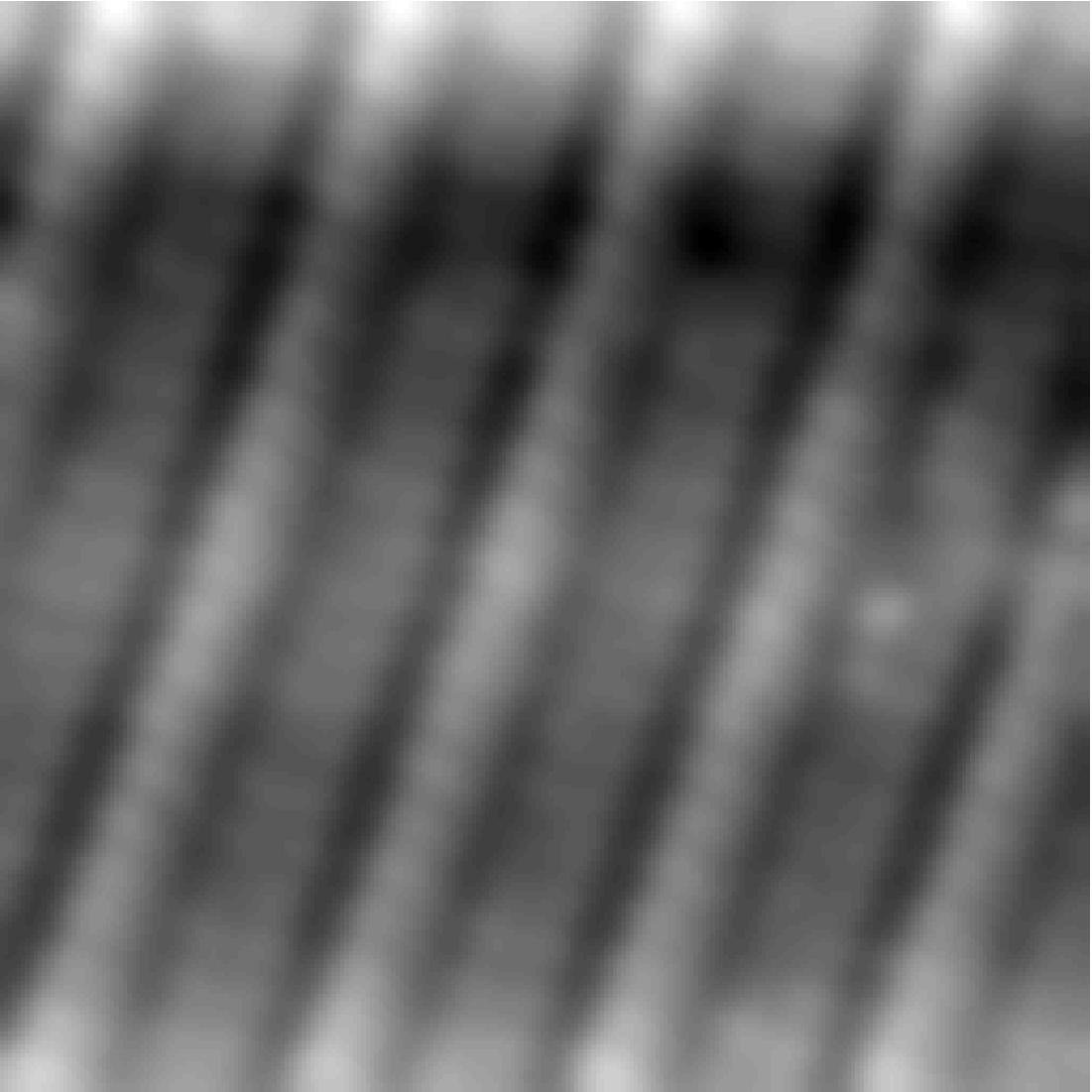}  &  
  \includegraphics[width=0.2\textwidth]{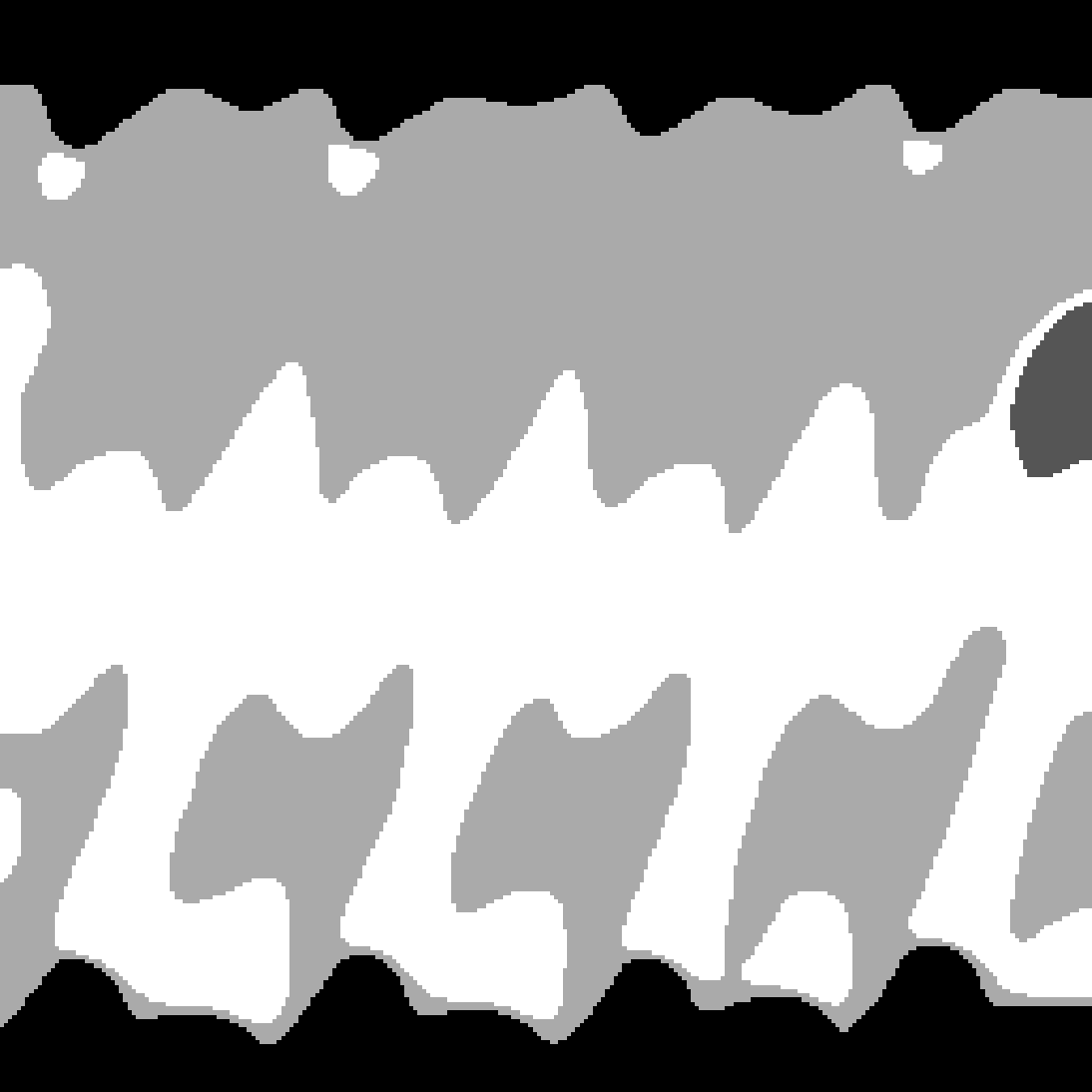} &
  \includegraphics[width=0.2\textwidth]{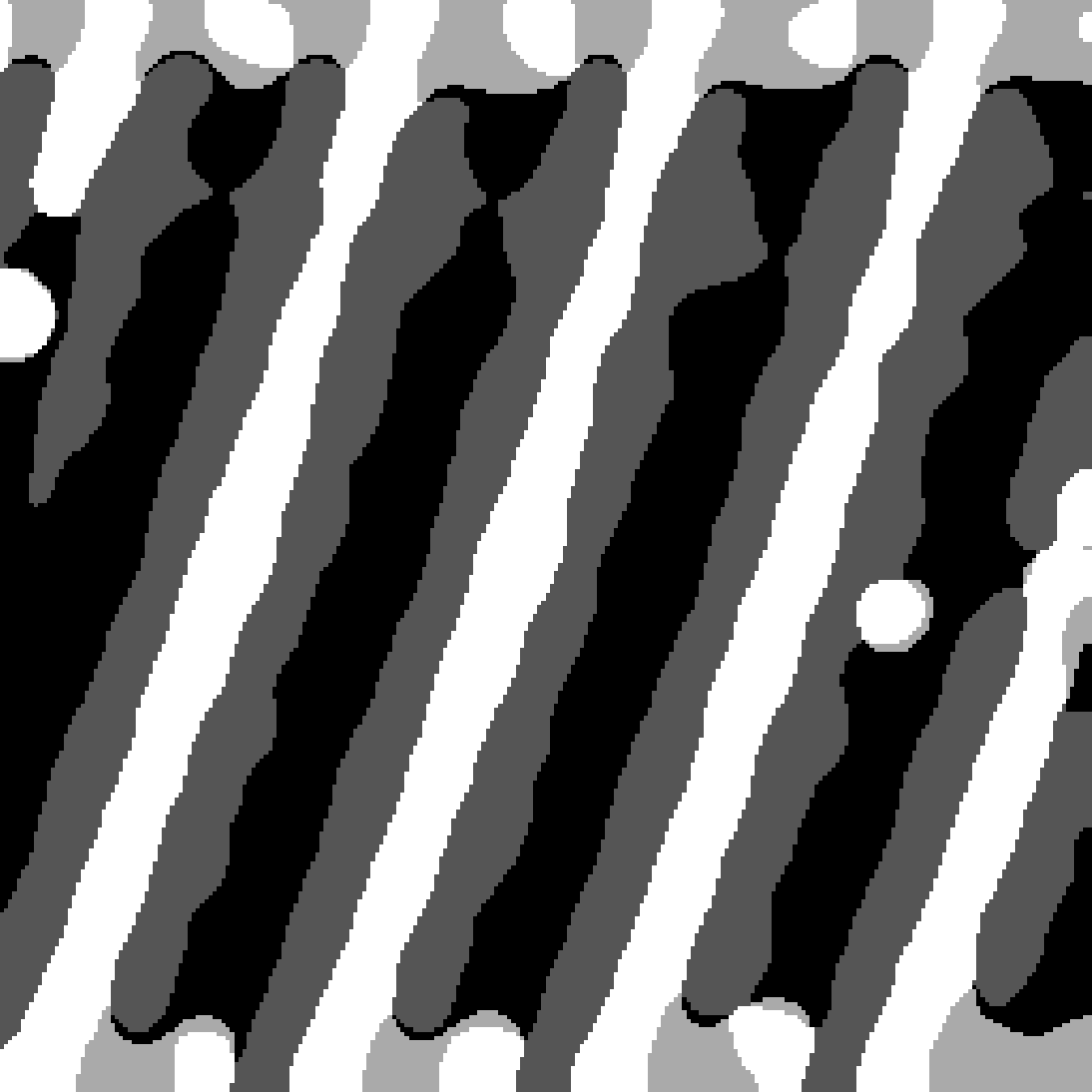} \\
\textbf{(c)}& 
  \includegraphics[width=0.2\textwidth]{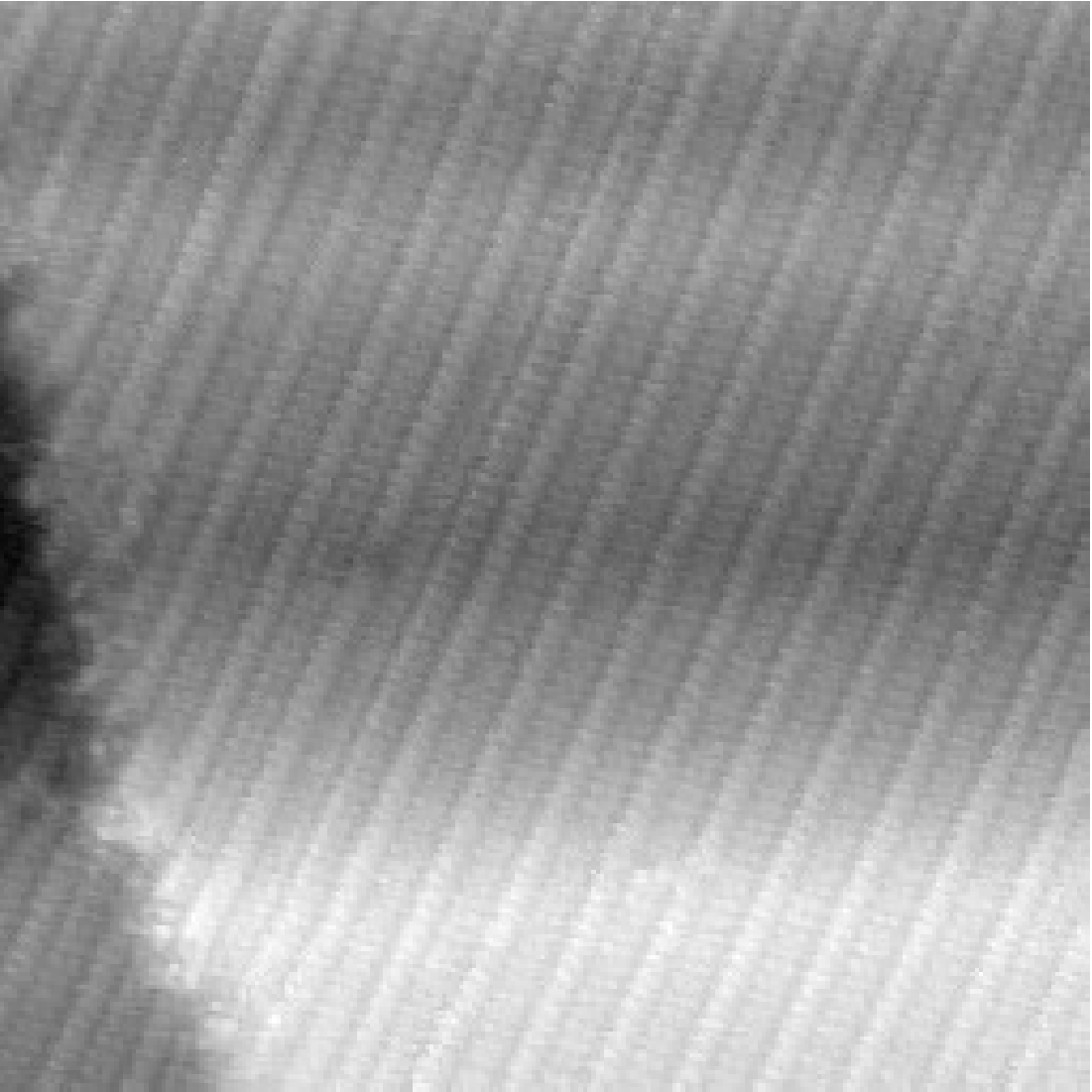} & 
  \includegraphics[width=0.2\textwidth]{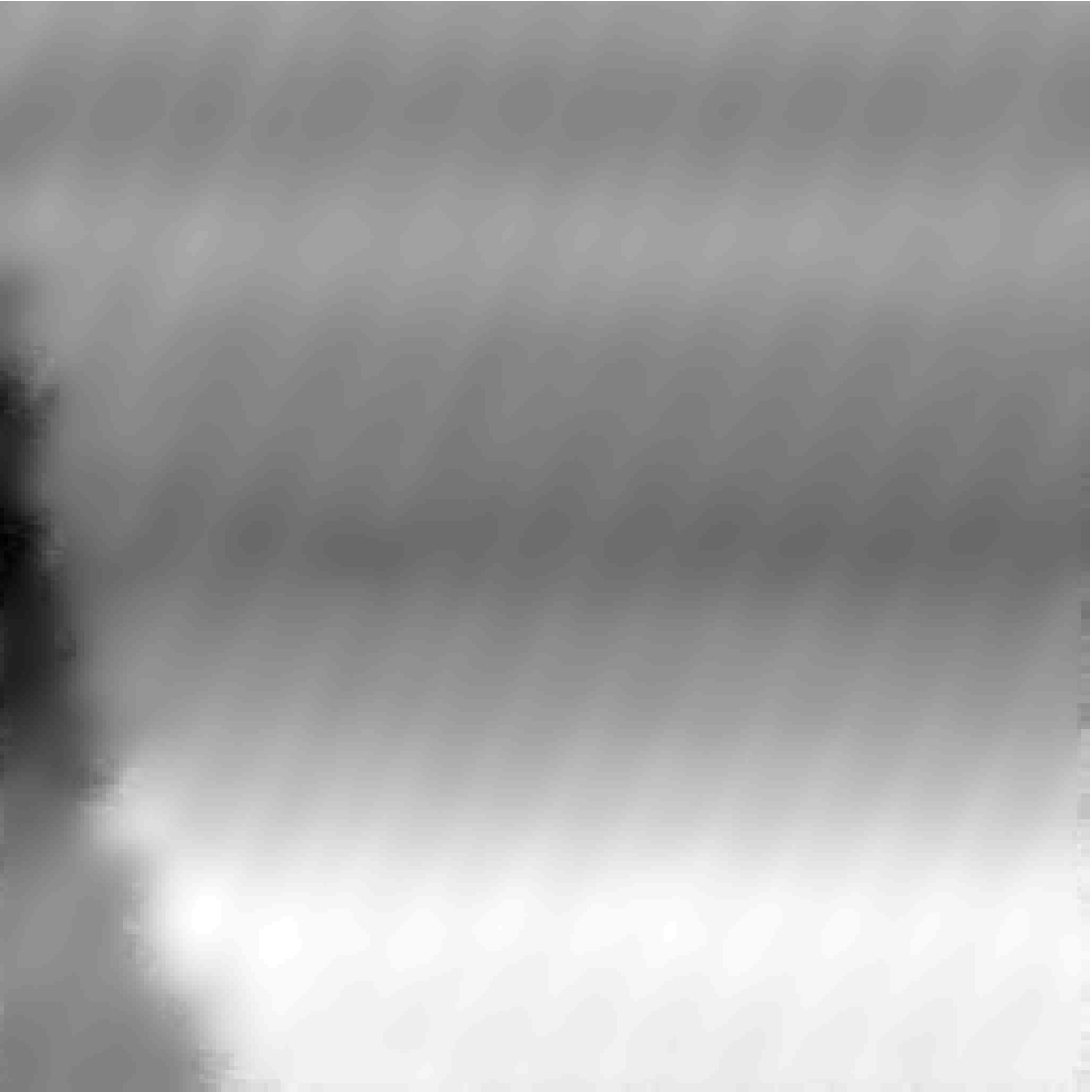}  &  
  \includegraphics[width=0.2\textwidth]{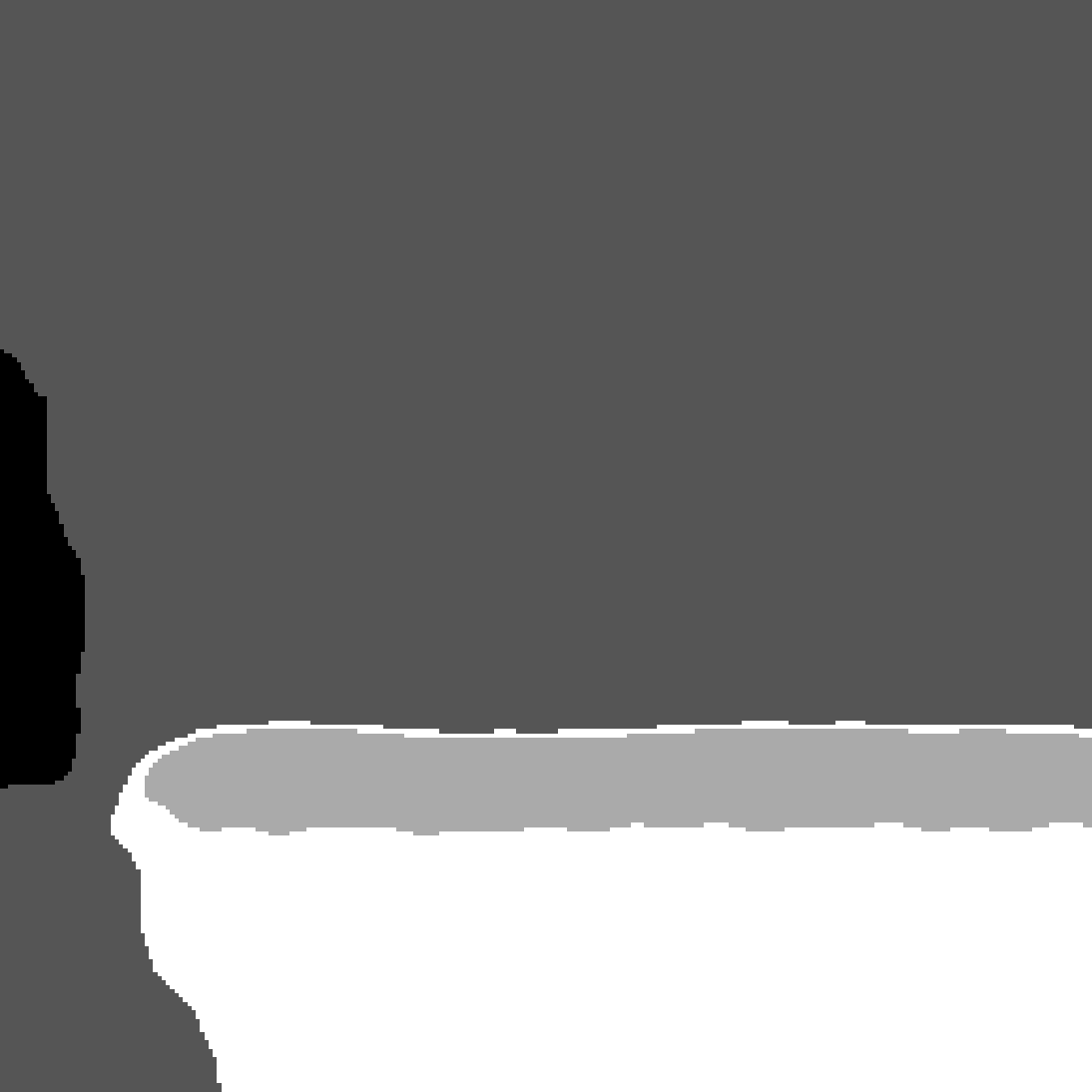} &
  \includegraphics[width=0.2\textwidth]{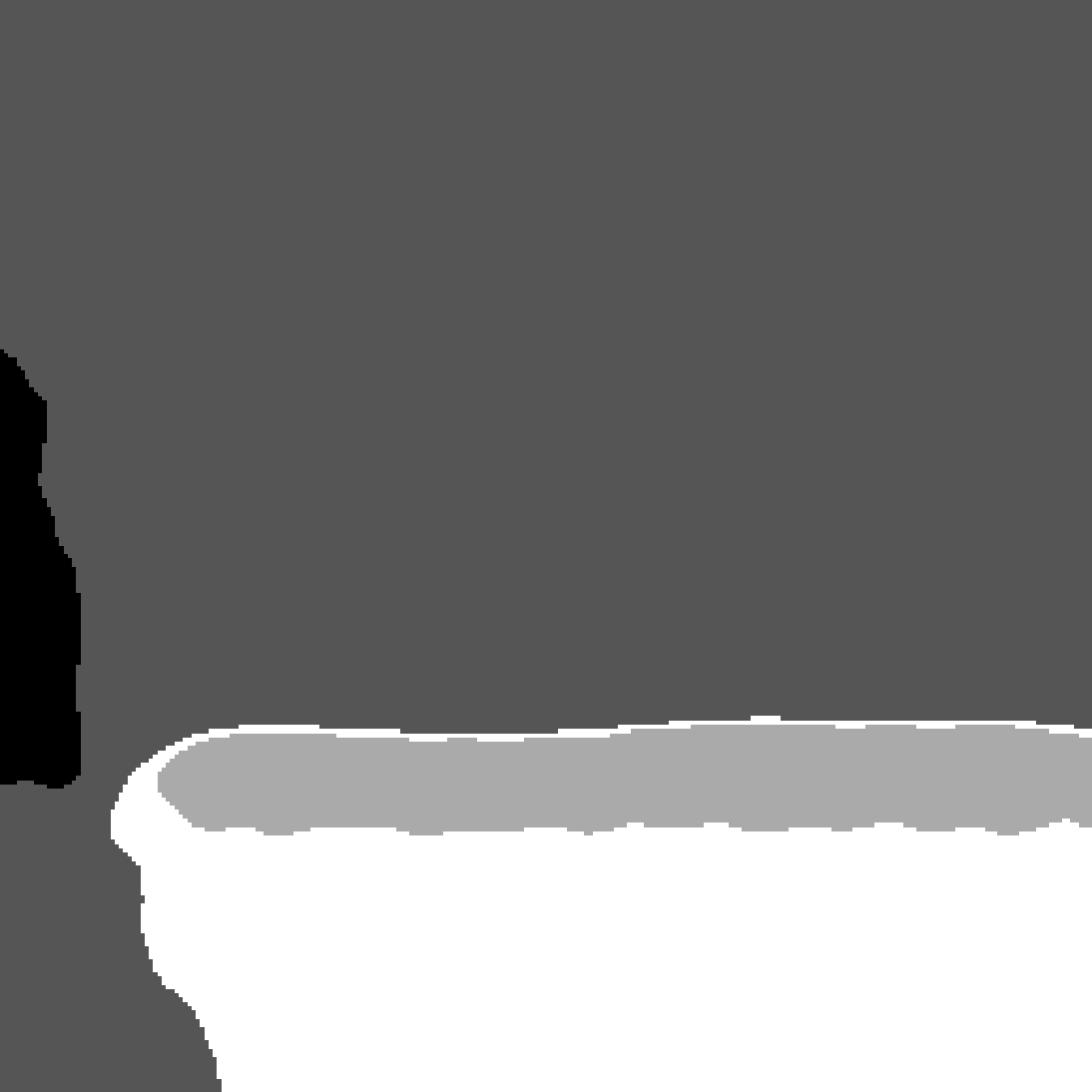} \\
\textbf{(d)}& 
  \includegraphics[width=0.2\textwidth]{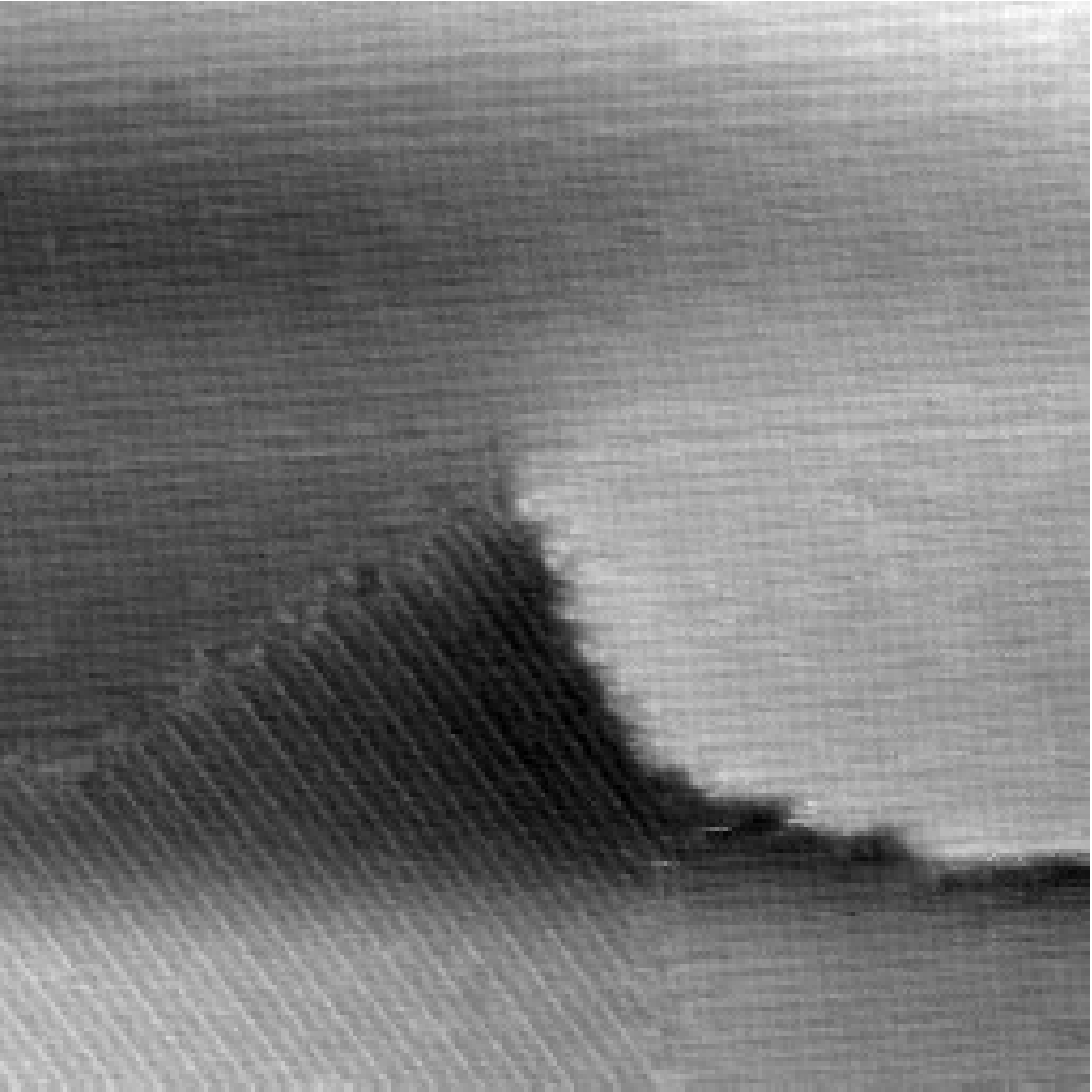} & 
  \includegraphics[width=0.2\textwidth]{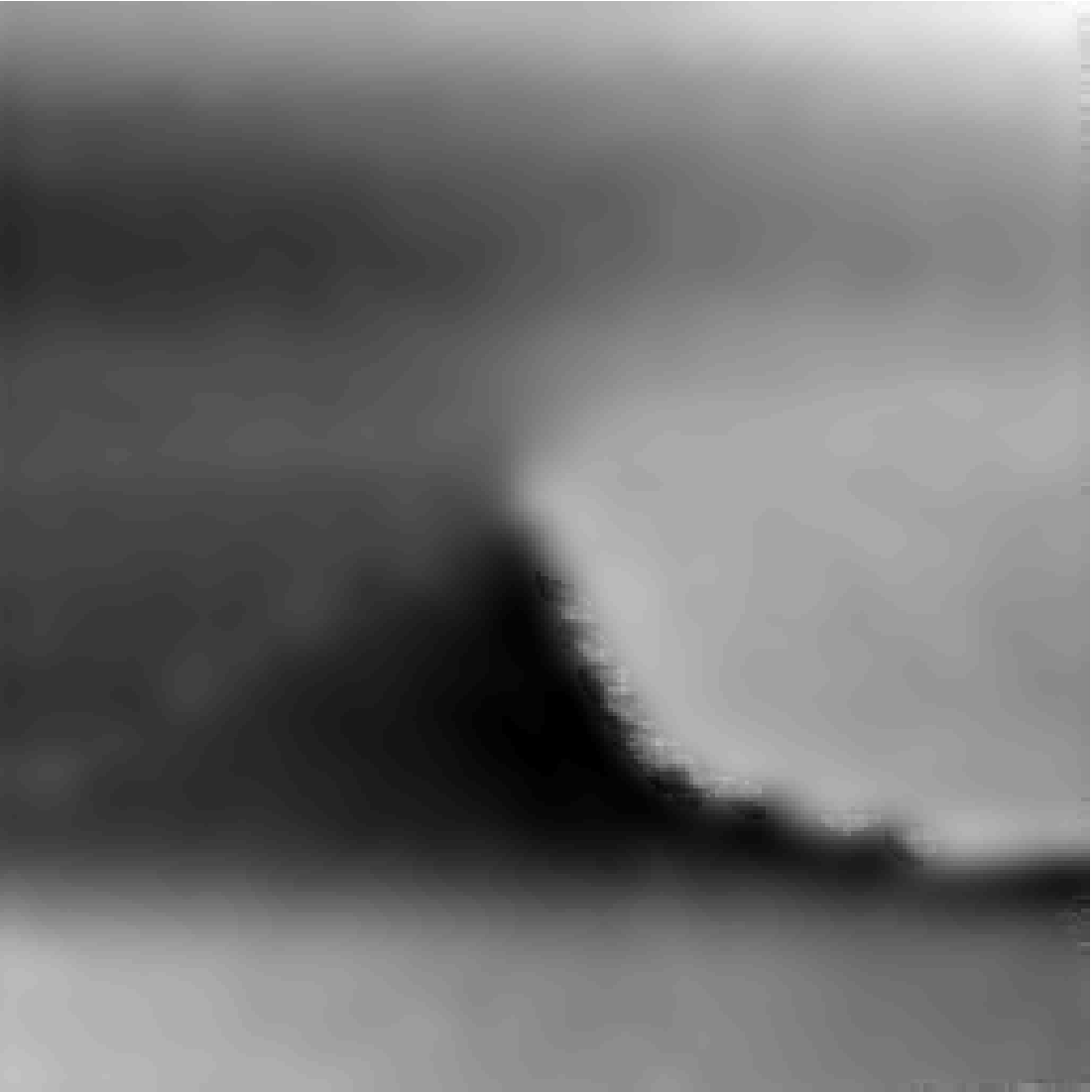}  &  
  \includegraphics[width=0.2\textwidth]{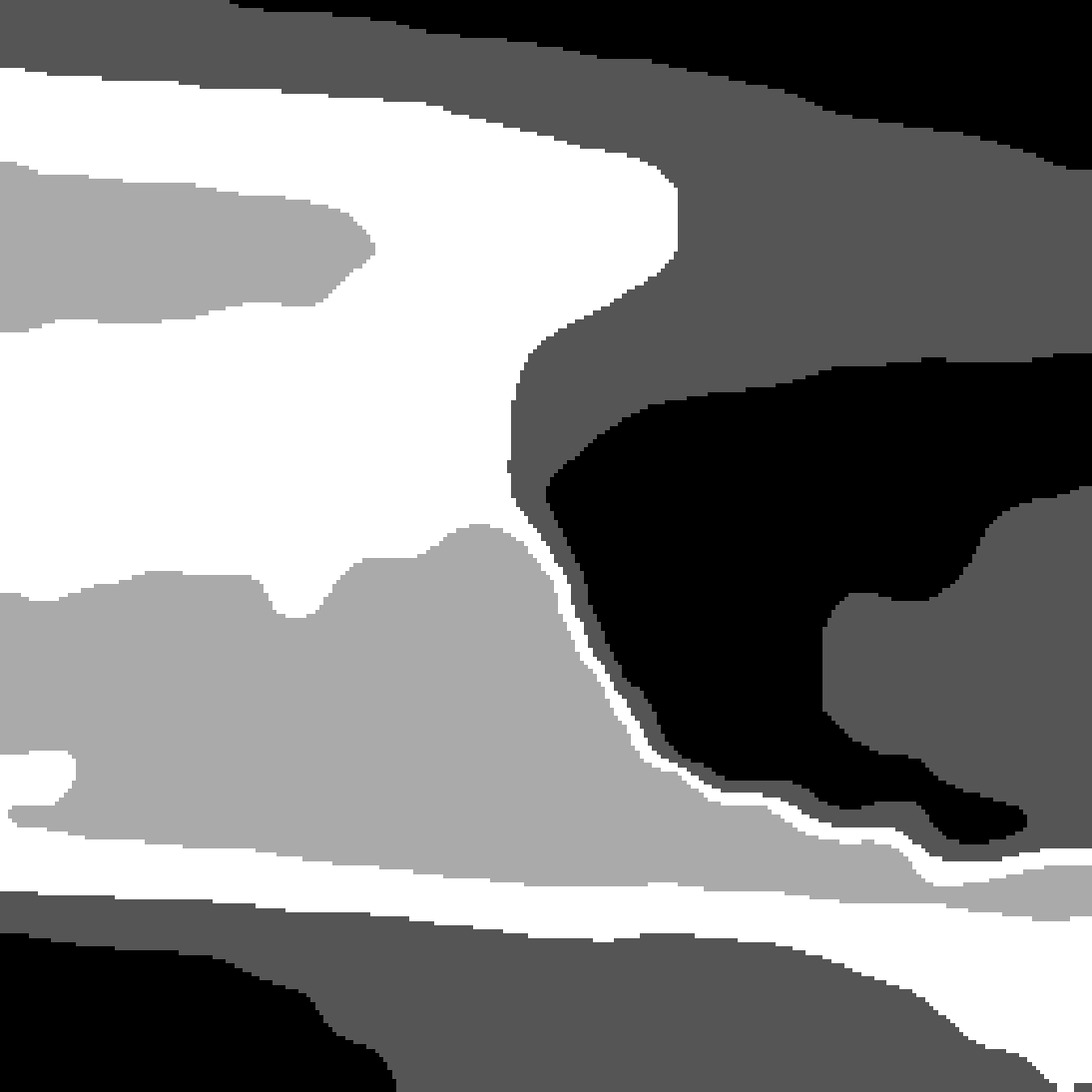} &
  \includegraphics[width=0.2\textwidth]{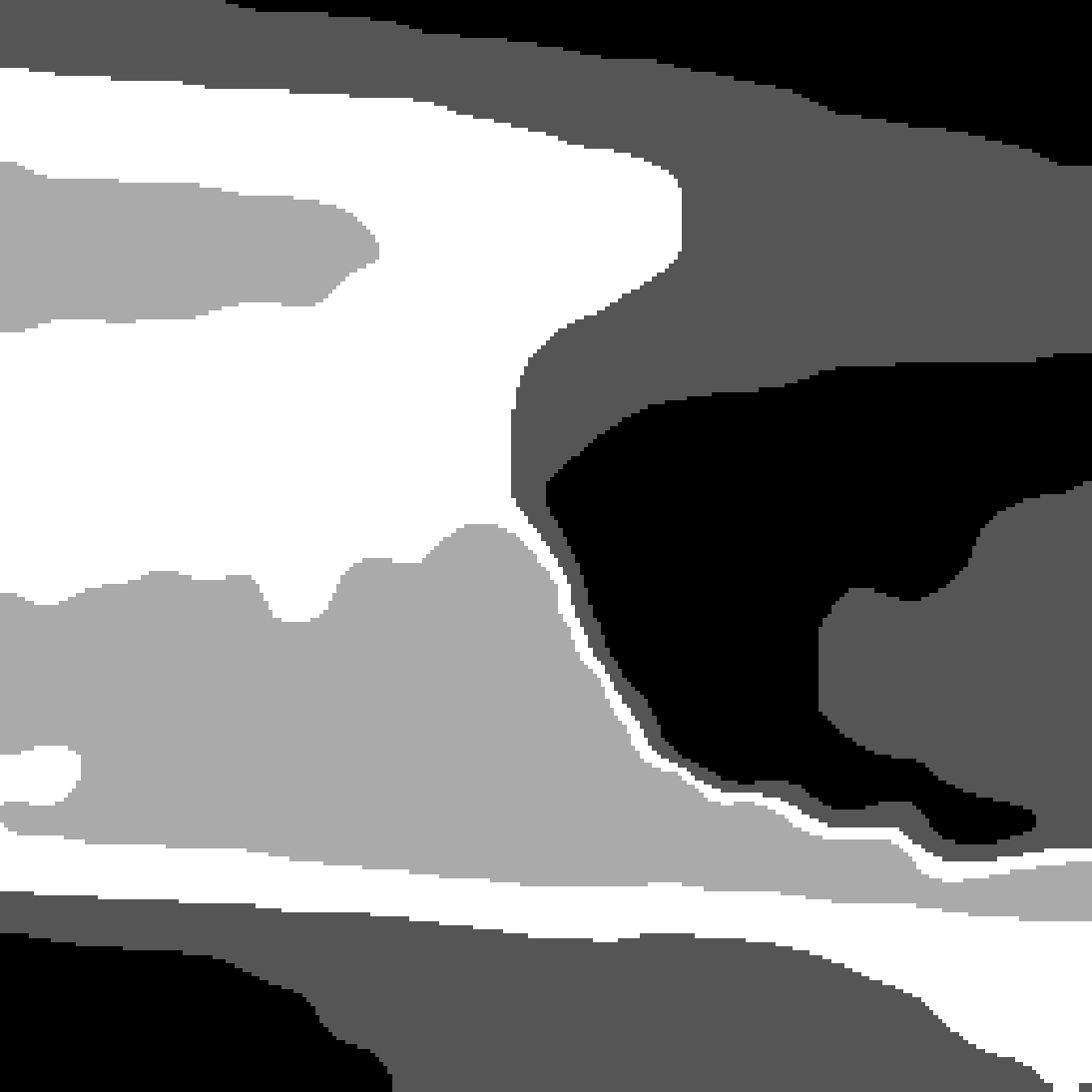} \\
\end{tabular}
\end{center}
\caption{Comparison between the multiphase CV and local MCV algorithms. 
The parameters are
\textbf{(a)} $\lambda= 10$, $\mu = 10^{-3} \times 255^2$, $\beta = 10$, $dt_{CV} = 0.75$, $dt_{CVloc} = 3.2$ 
\textbf{(b)} $\lambda = 10$, $\mu = 10^{-3} \times 255^2$, $\beta = 300$, $dt_{CV} = 4$, $dt_{CVloc} = 4$ 
\textbf{(c)} $\lambda= 10$, $\mu = 10^{-3}  \times 255^2$, $\beta = 60$, $dt_{CV} = 2$, $dt_{CVloc} = 2$ 
\textbf{(d)} $\lambda= 10$, $\mu = 10^{-3} \times 255^2$, $\beta = 10$, $dt_{CV} = 2.5$, $dt_{CVloc} = 2$. Raw scanning tunneling microscope images of cyanide on Au\{111\}, reproduced from \cite{guttentag2016hexagons} with permission. Images copyright American Chemical Society}
\label{fig:result1}
\end{figure*}

\begin{figure*}[!t]
\begin{center}
\begin{tabular}{m{1mm}>{\centering\arraybackslash}>{\centering\arraybackslash}m{0.2\textwidth}>{\centering\arraybackslash}m{0.2\textwidth}>{\centering\arraybackslash}m{
0.2\textwidth}>{\centering\arraybackslash}m{0.2\textwidth}}
\centering
& \textbf{original} & \textbf{cartoon} & \textbf{multiphase} & \textbf{local multiphase} \\
\textbf{(a)}& 
  \includegraphics[width=0.2\textwidth]{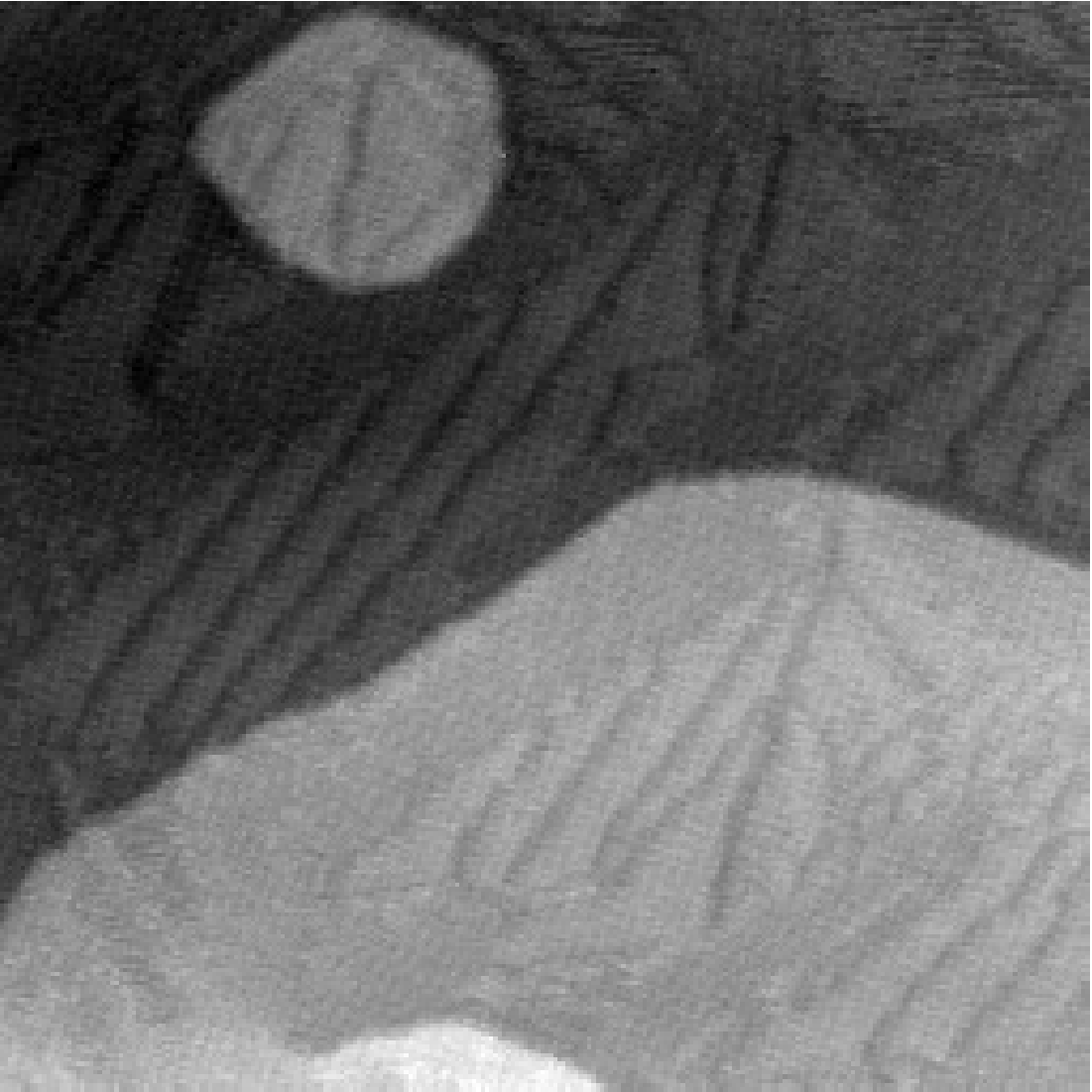} & 
  \includegraphics[width=0.2\textwidth]{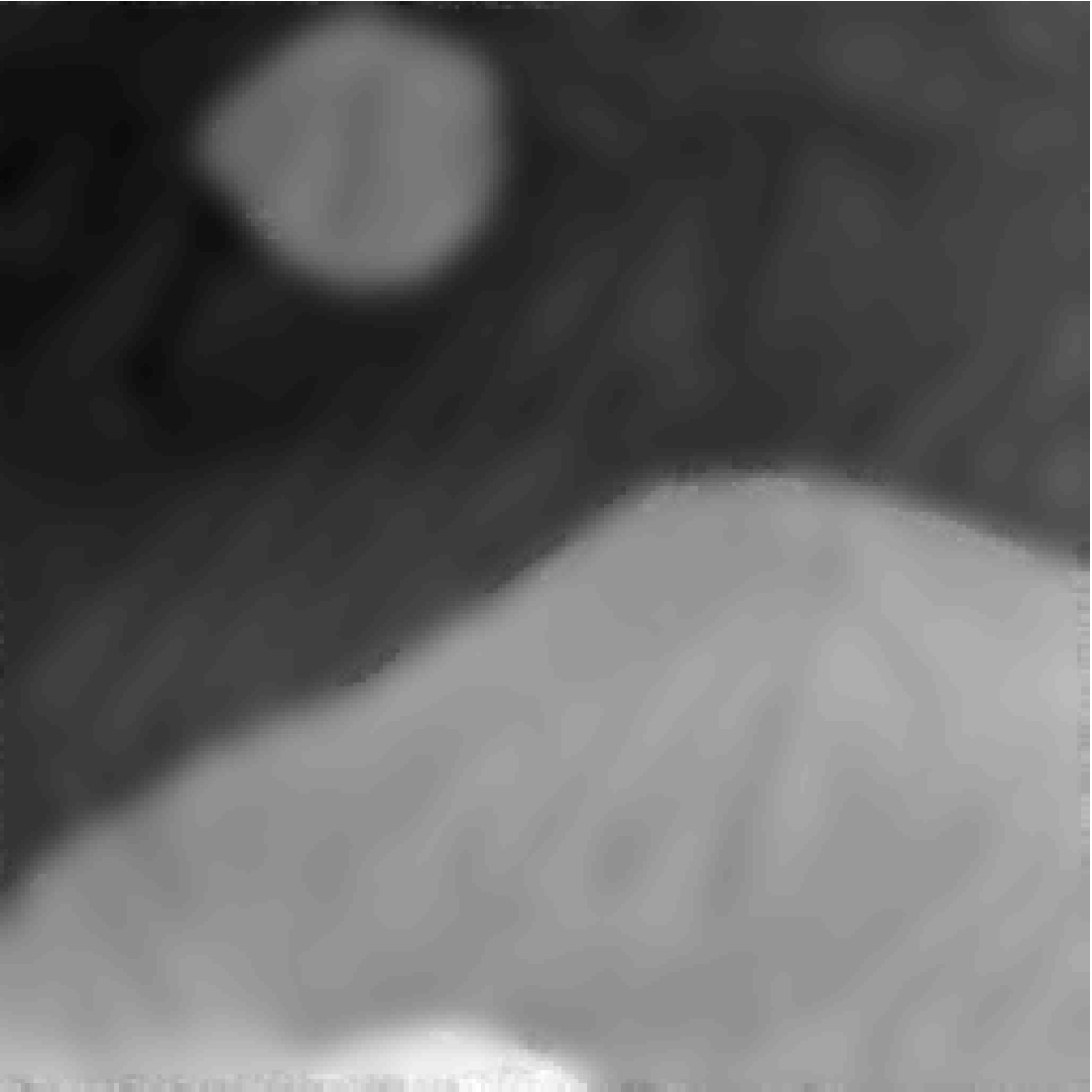}  &  
  \includegraphics[width=0.2\textwidth]{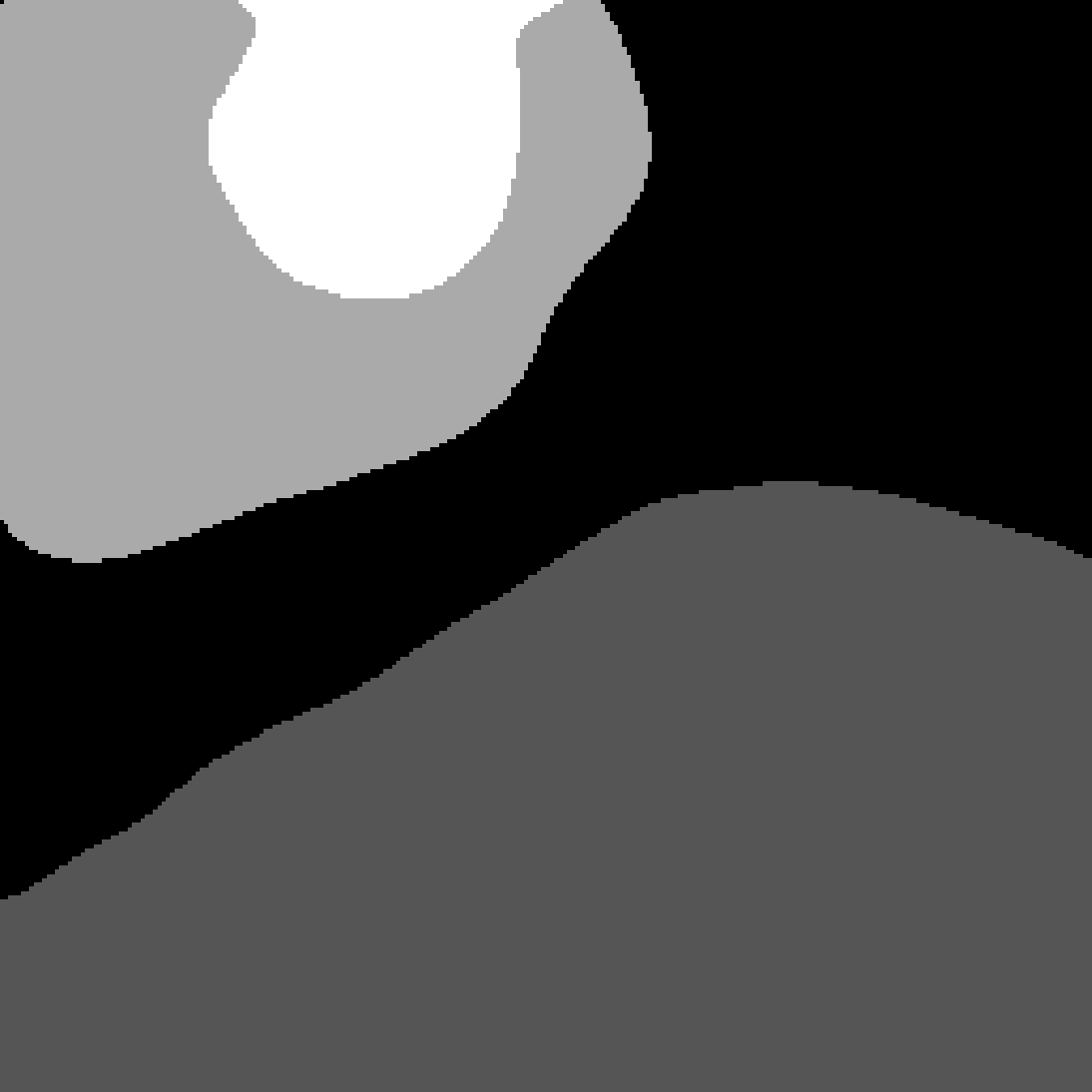} &
  \includegraphics[width=0.2\textwidth]{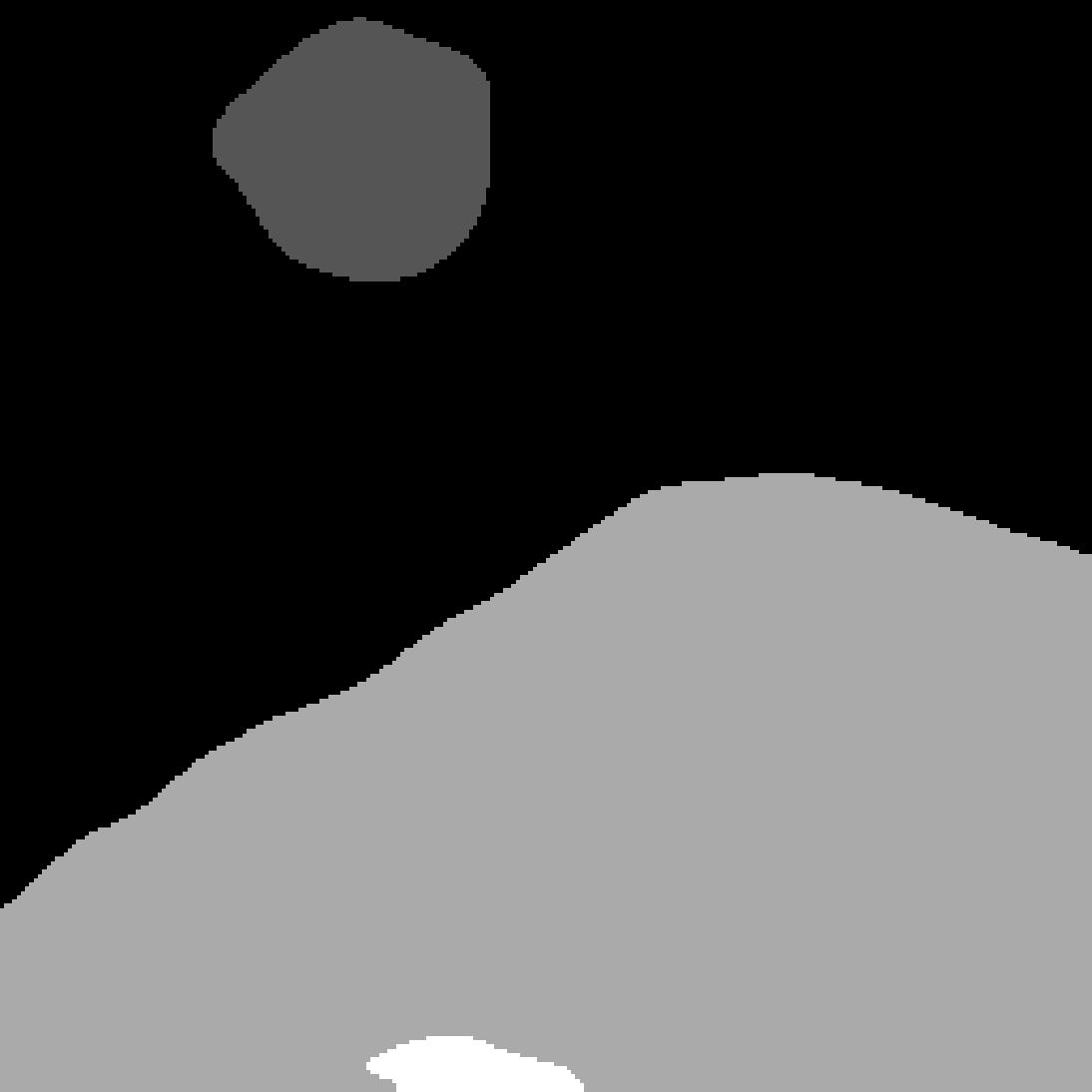} \\
\textbf{(b)}& 
  \includegraphics[width=0.2\textwidth]{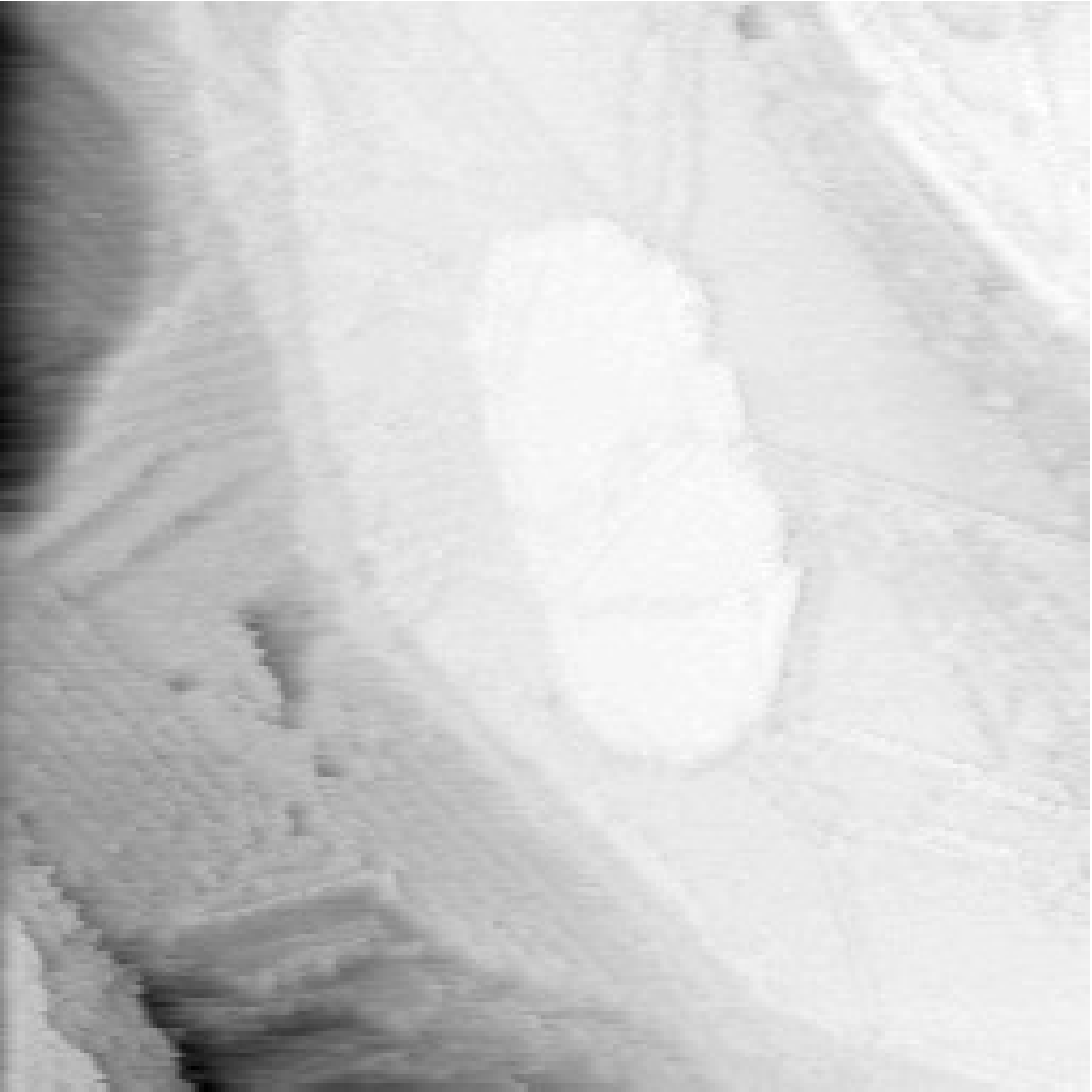} & 
  \includegraphics[width=0.2\textwidth]{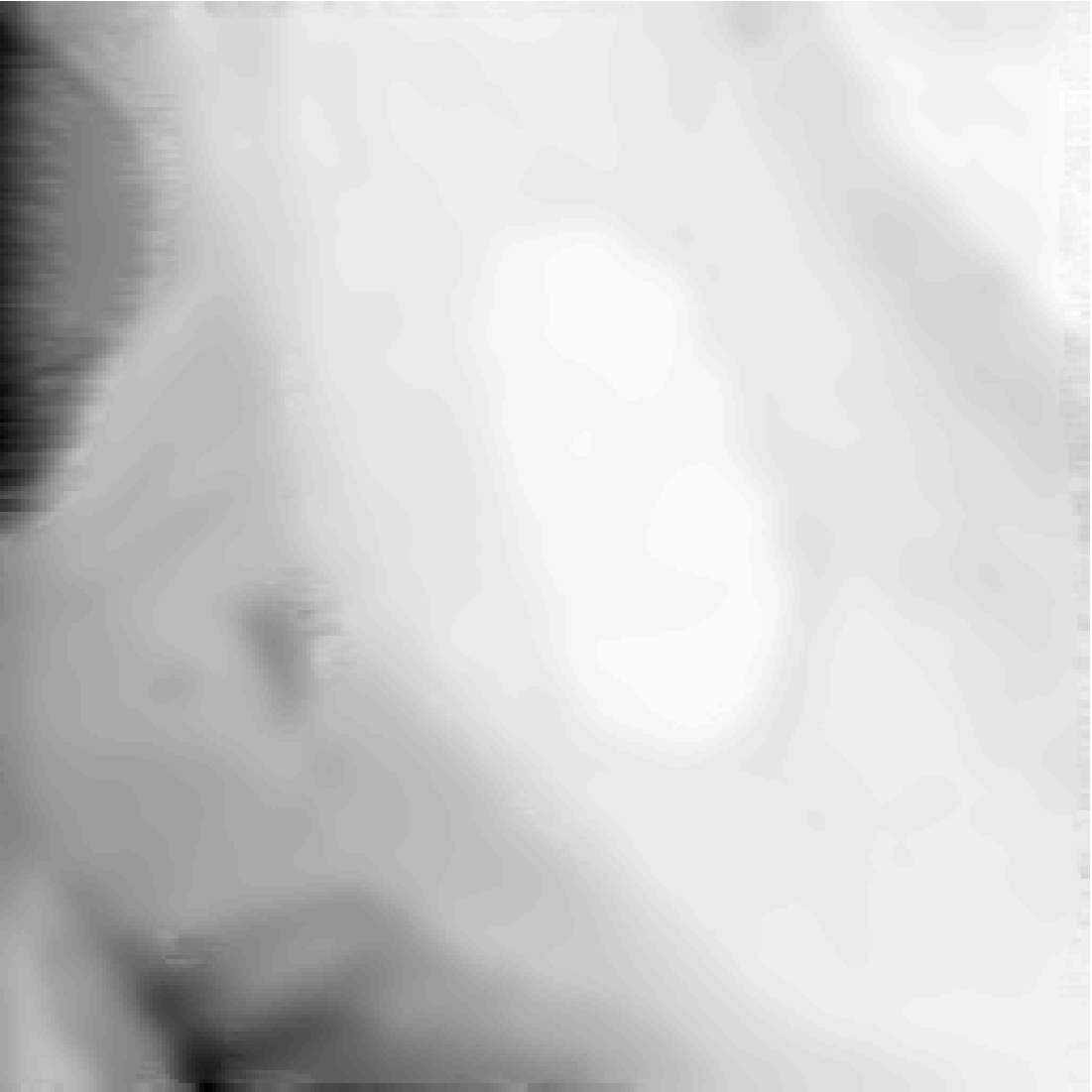}  &  
  \includegraphics[width=0.2\textwidth]{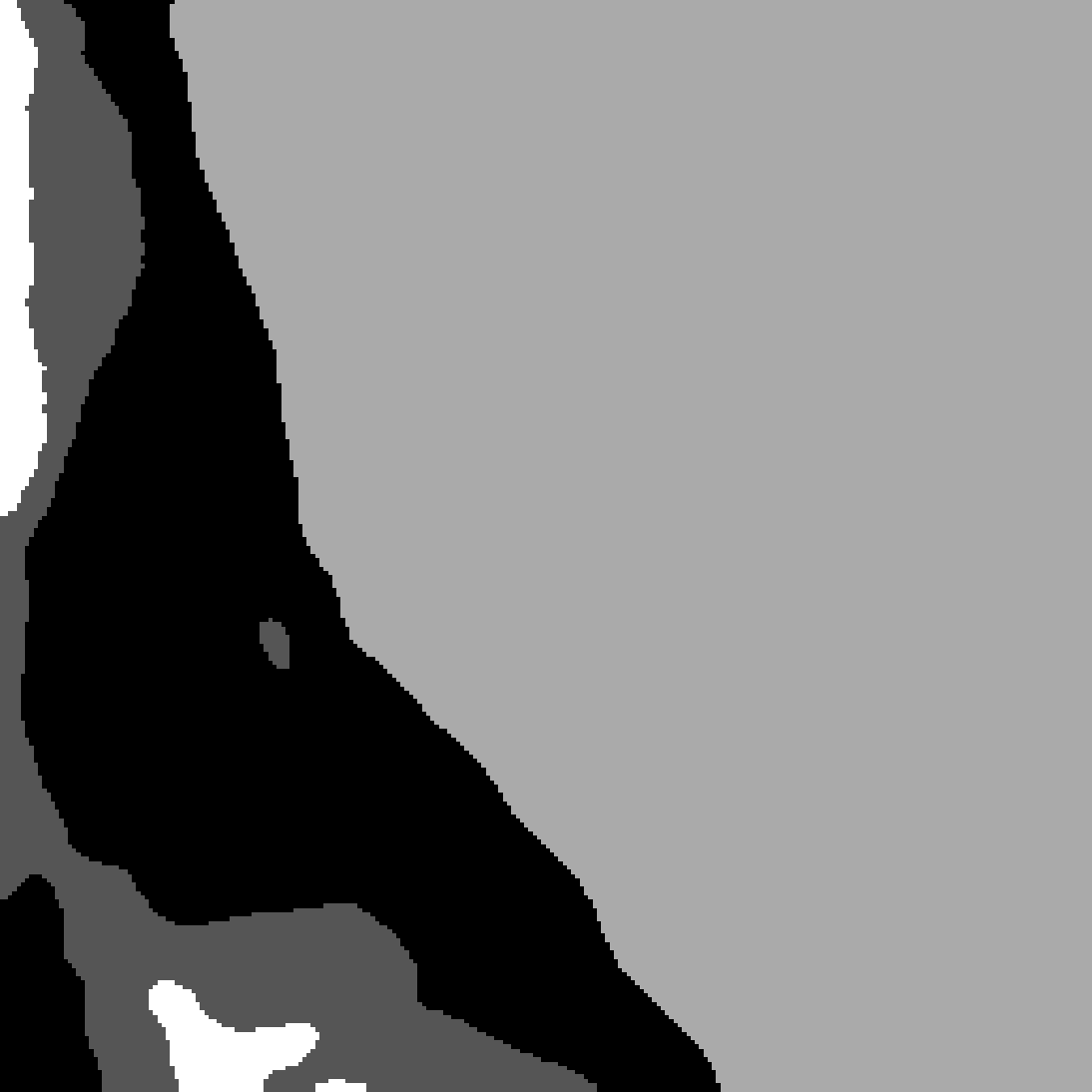} &
  \includegraphics[width=0.2\textwidth]{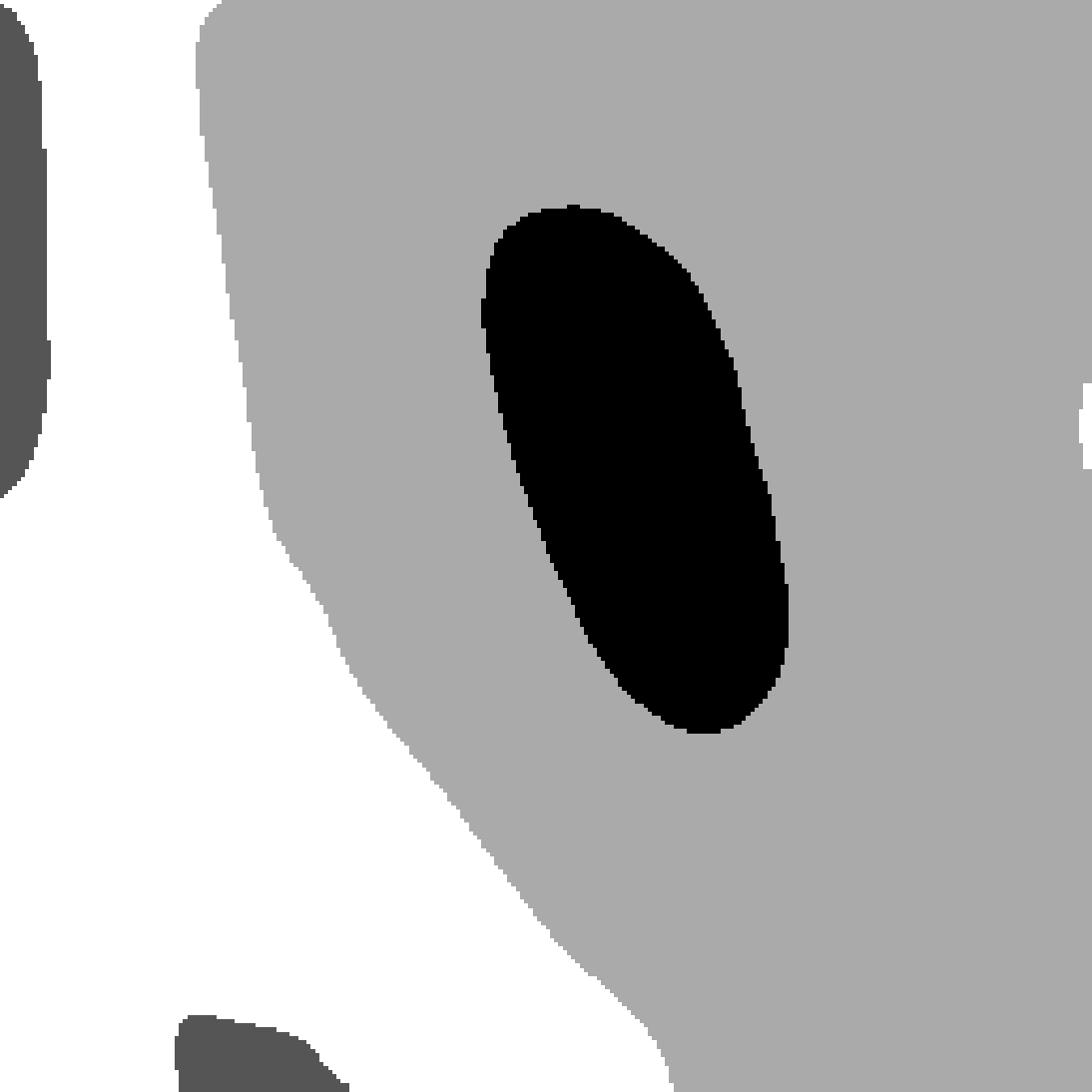} \\
\textbf{(c)}& 
  \includegraphics[width=0.2\textwidth]{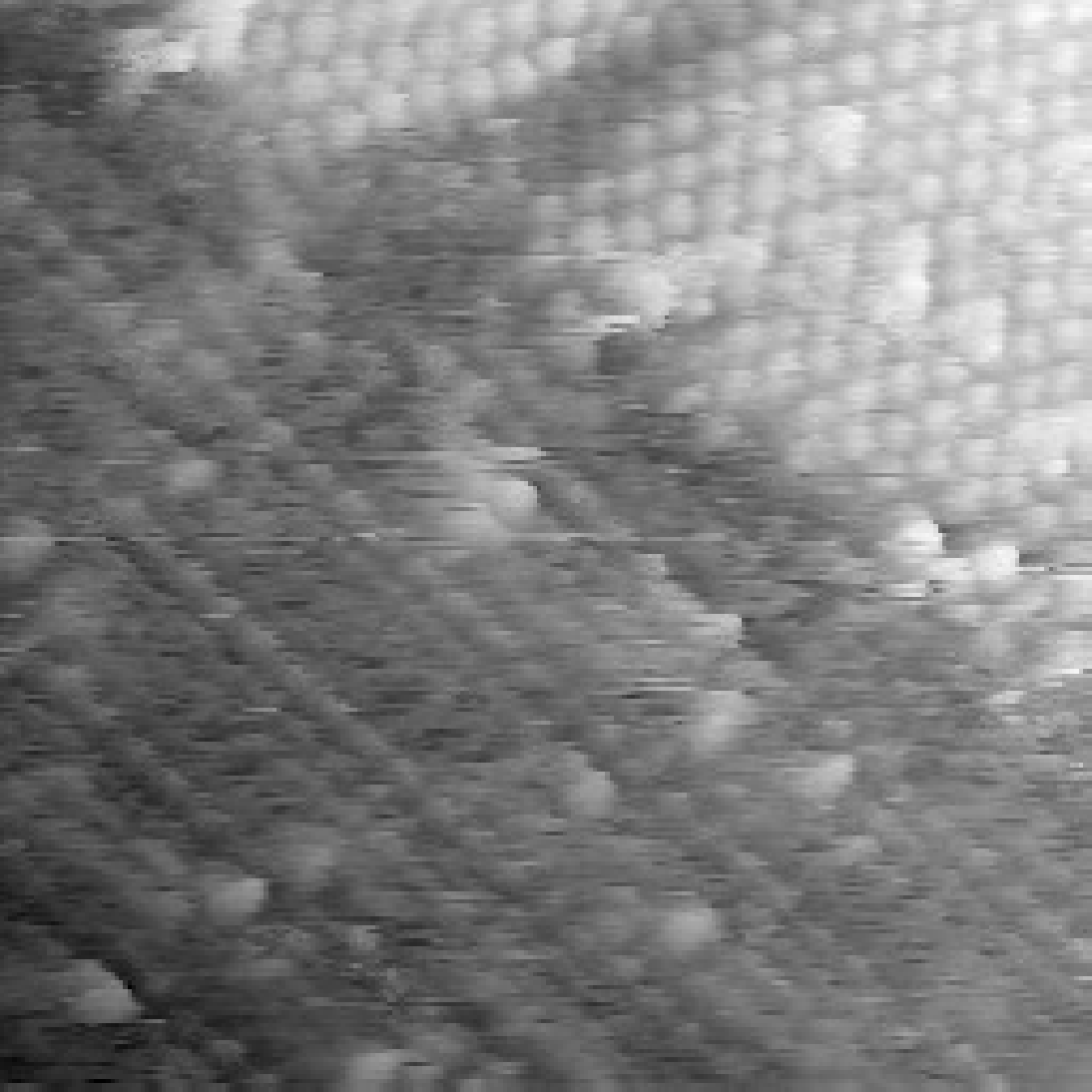} & 
  \includegraphics[width=0.2\textwidth]{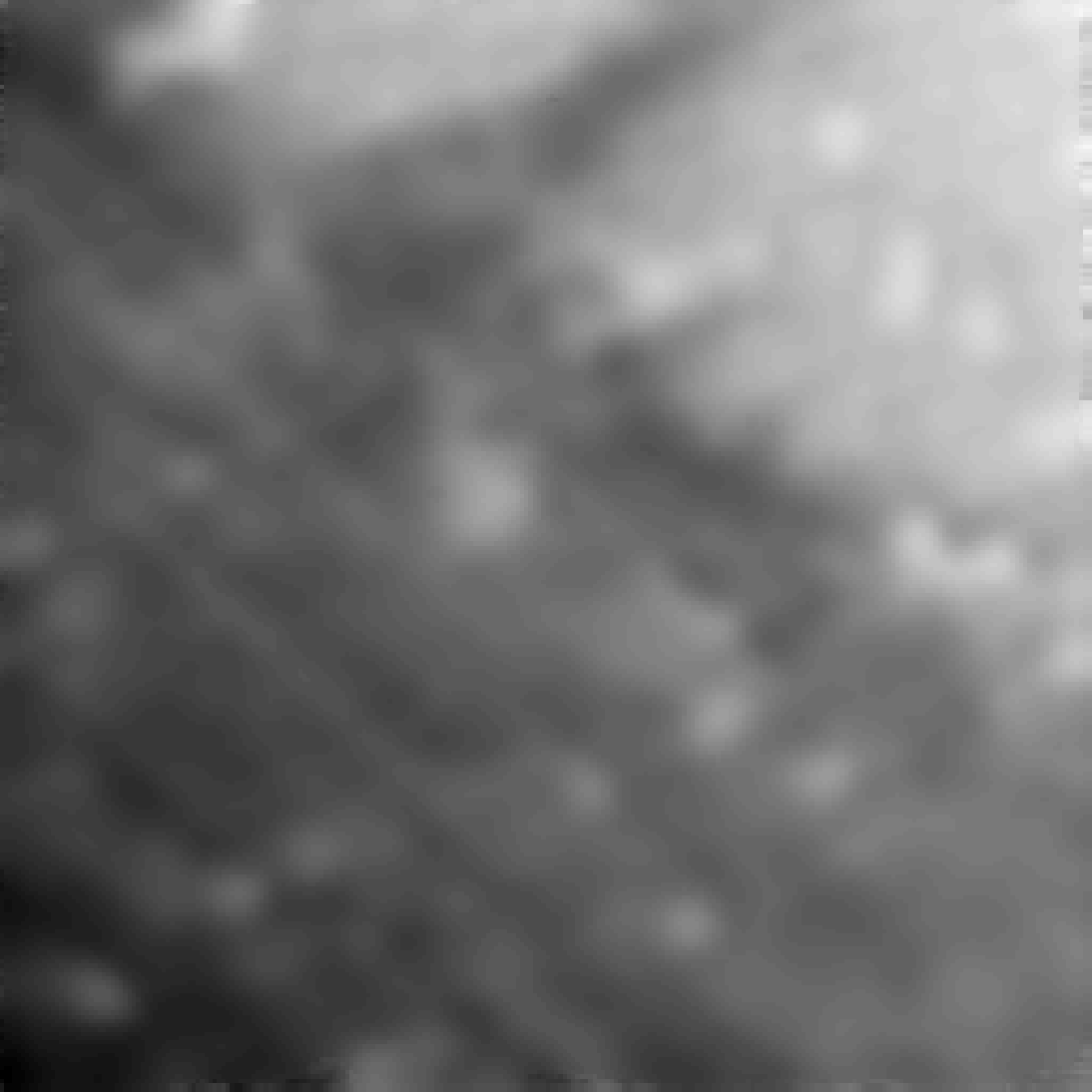}  &  
  \includegraphics[width=0.2\textwidth]{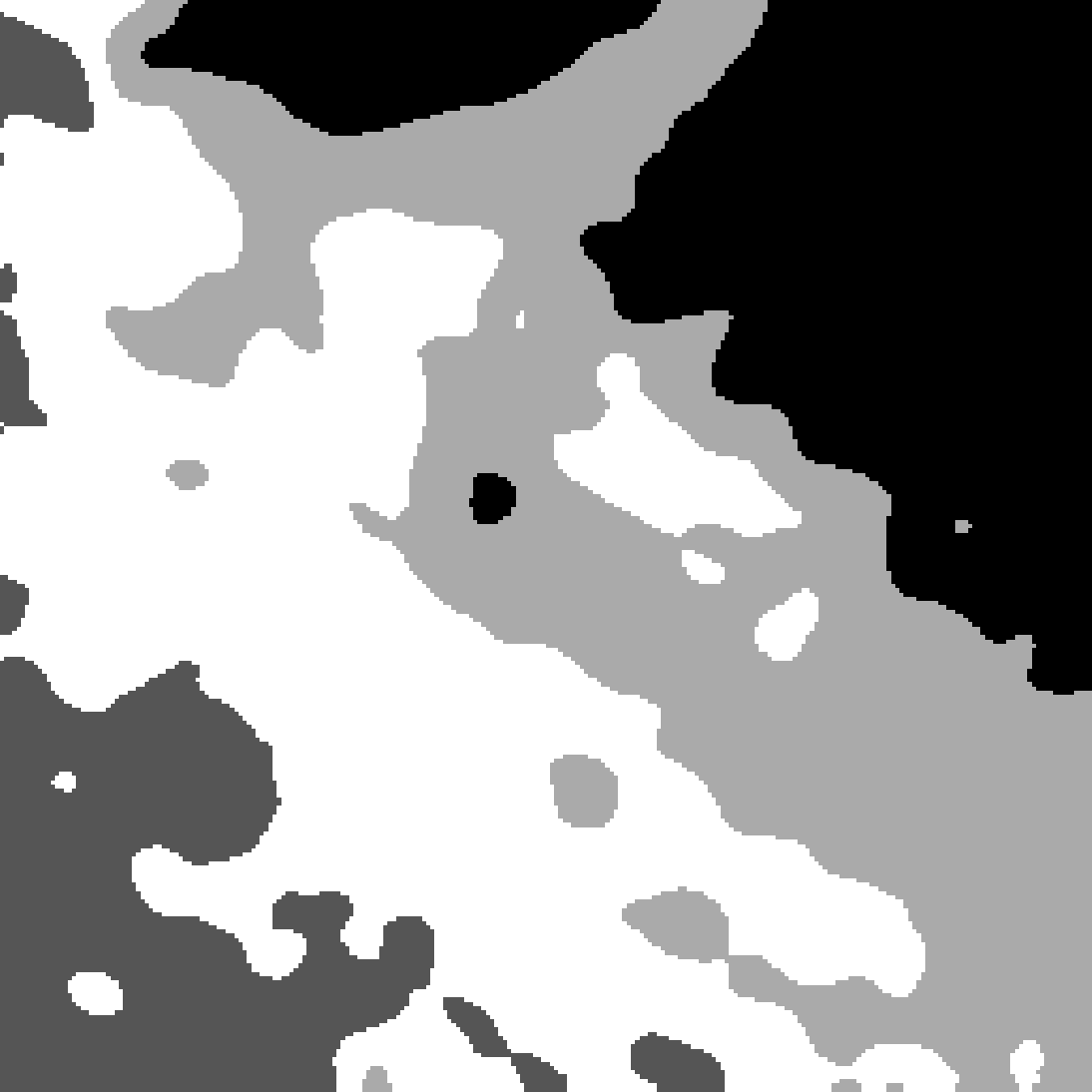} &
  \includegraphics[width=0.2\textwidth]{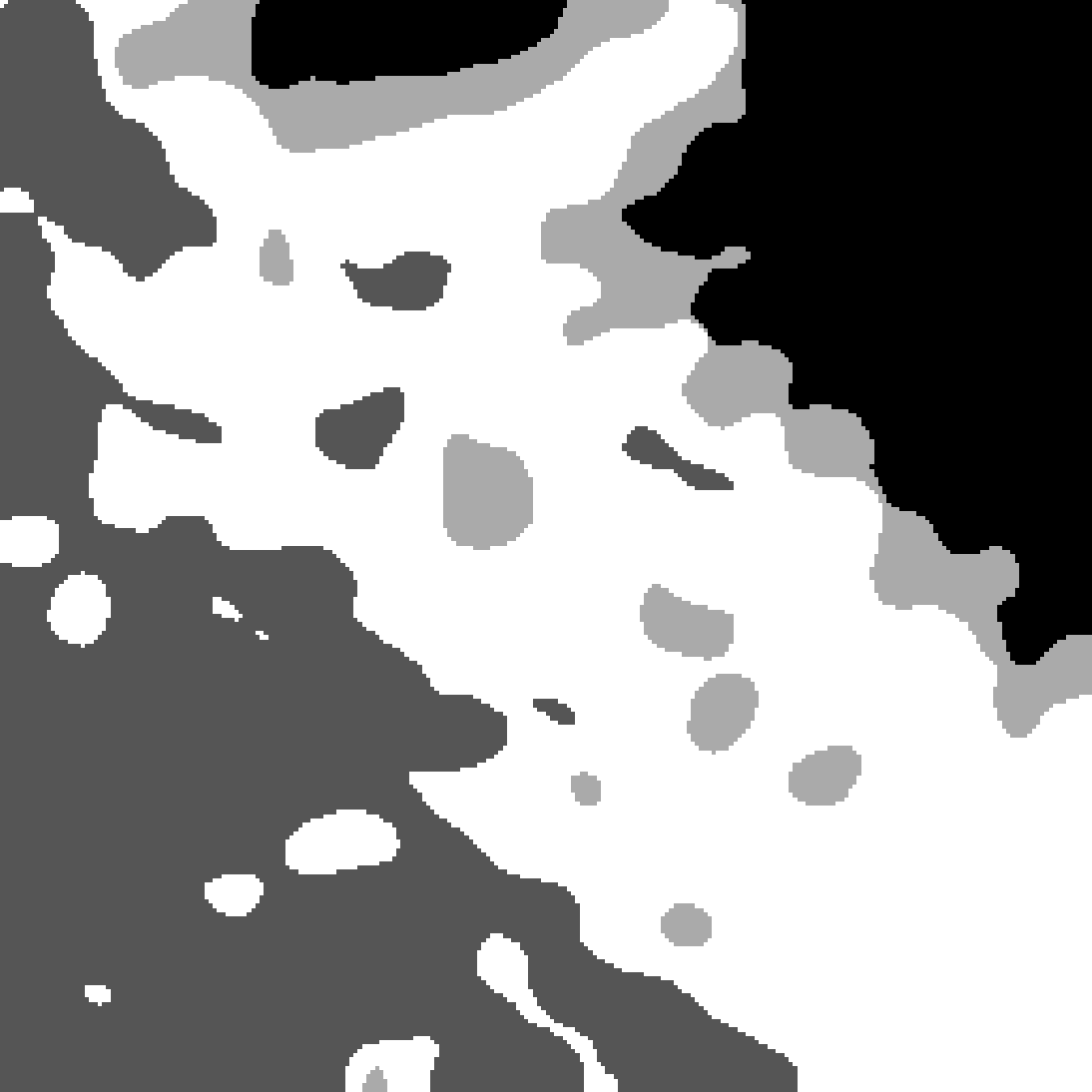} \\
\textbf{(d)}& 
  \includegraphics[width=0.2\textwidth]{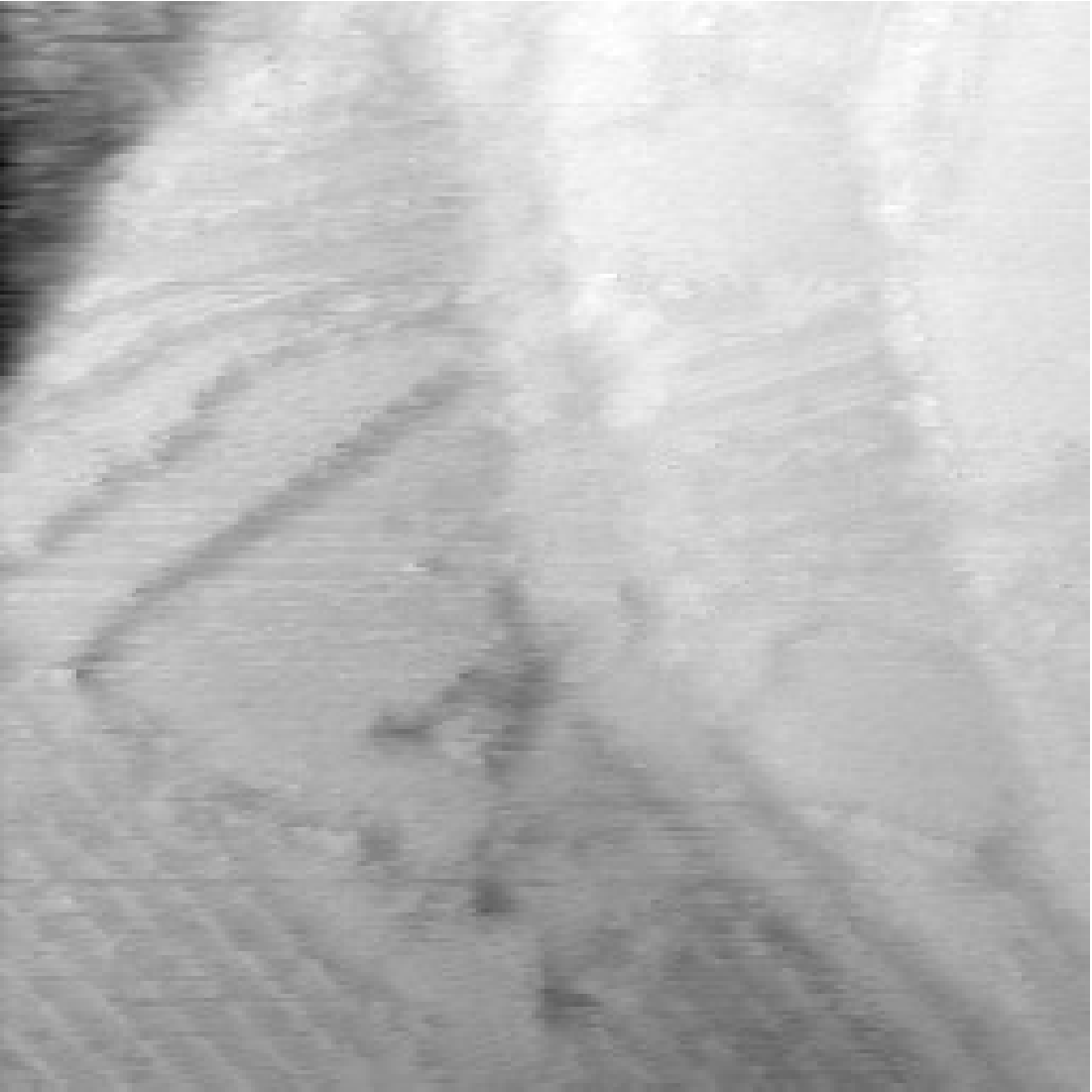} & 
  \includegraphics[width=0.2\textwidth]{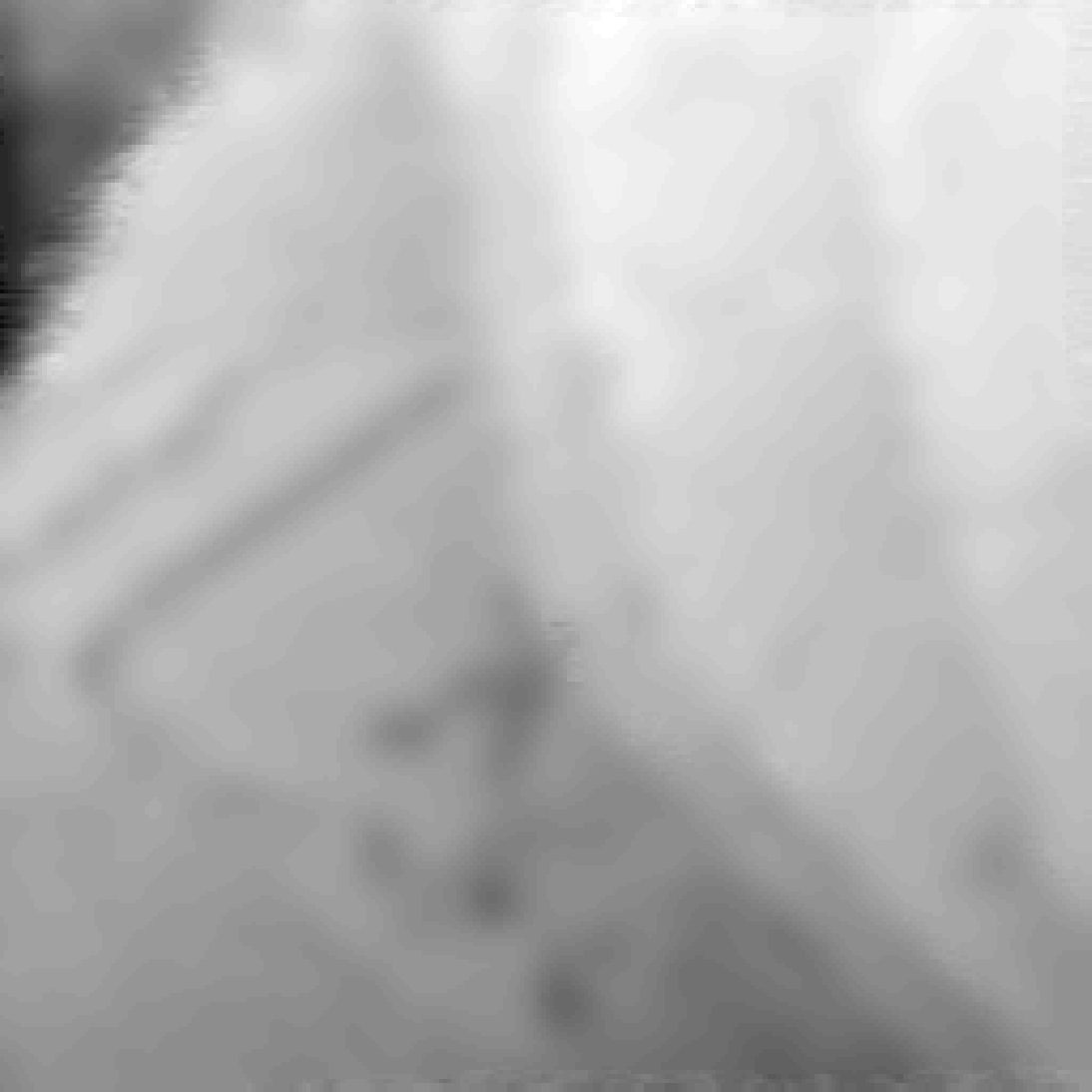}  &  
  \includegraphics[width=0.2\textwidth]{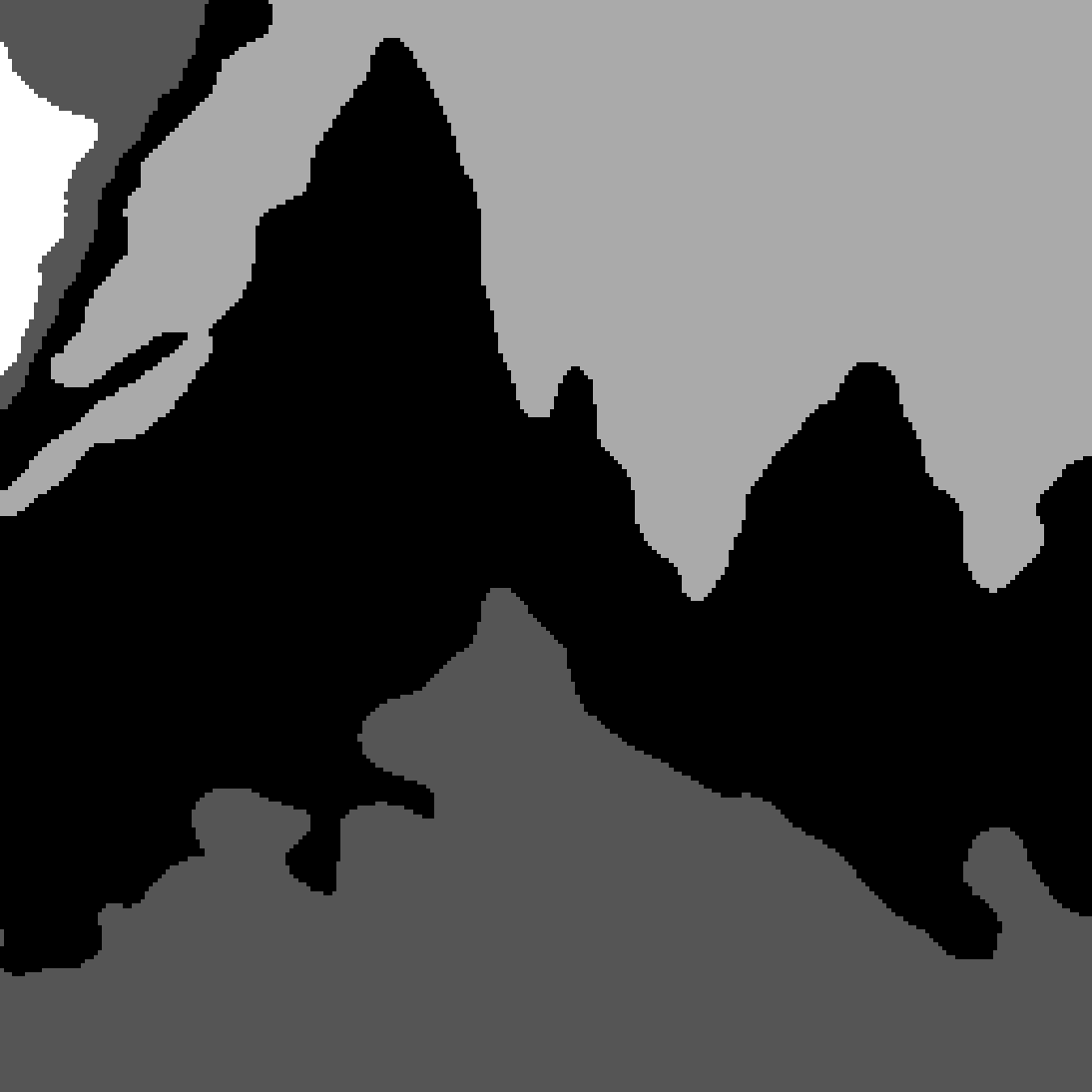} &
  \includegraphics[width=0.2\textwidth]{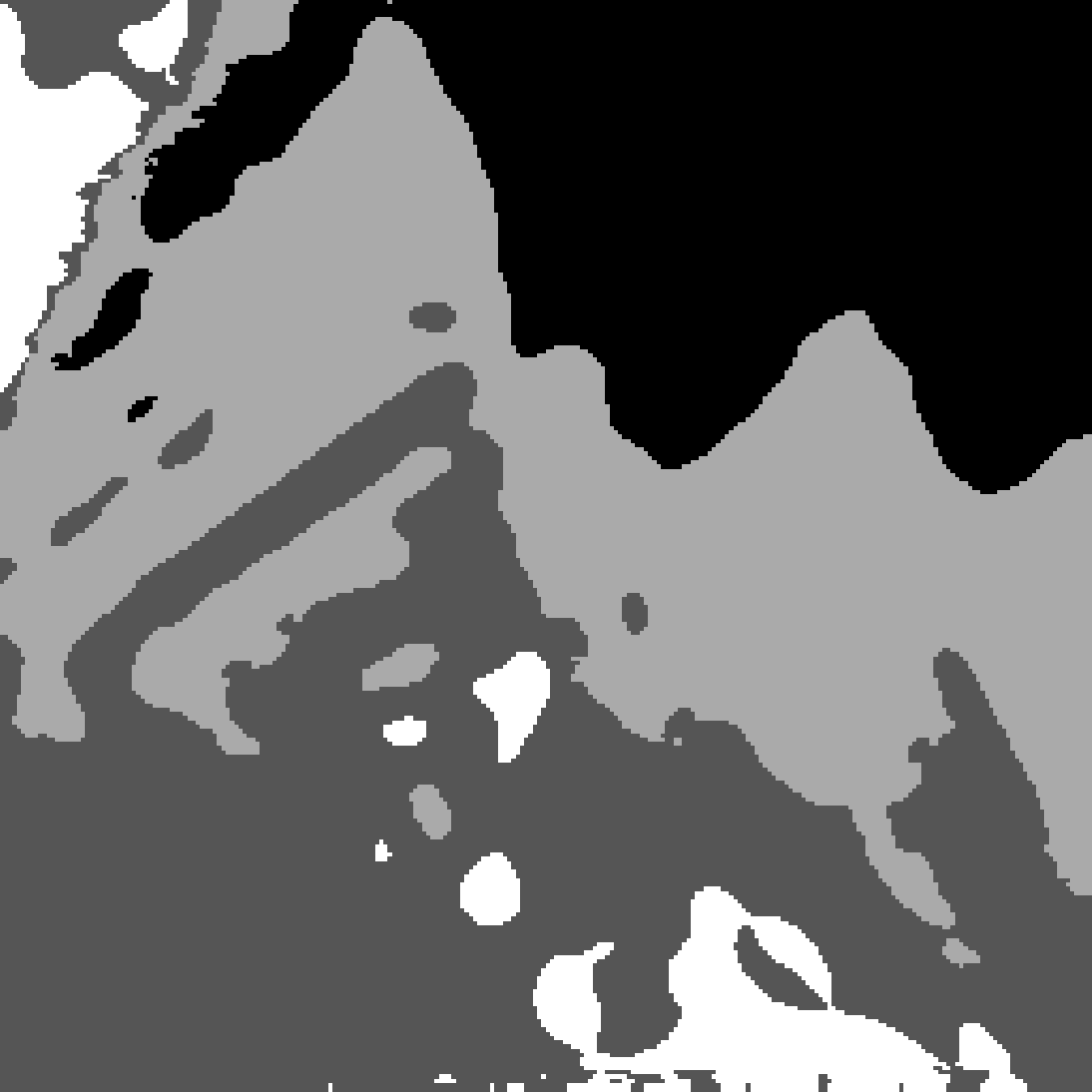} \\
\end{tabular}
\end{center}
\caption{Comparison between the multiphase CV and local MCV algorithms. 
The parameters are 
\textbf{(a)} $\lambda= 5$, $\mu = 10^{-1} \times 255^2$, $\beta= 70$, $dt_{CV} = 0.35$, $dt_{CVloc} = 0.10$ 
\textbf{(b)} $\lambda= 10$, $\mu = 10^{-2} \times 255^2$, $\beta = 30$, $dt_{CV} = 0.1$, $dt_{CVloc} = 6.5$ 
\textbf{(c)} $\lambda= 7$, $\mu = 10^{-4}  \times 255^2$, $\beta = 65$, $dt_{CV} = 5$, $dt_{CVloc} = 18$ 
\textbf{(d)} $\lambda= 5$, $\mu = 10^{-4} \times 255^2$, $\beta = 50$, $dt_{CV} = 12$, $dt_{CVloc} = 0.6$. Raw scanning tunneling microscope images of cyanide on Au\{111\}, reproduced from \cite{guttentag2016hexagons} with permission. Images copyright American Chemical Society}
\label{fig:result2}
\end{figure*}

For comparison, we apply the multiphase algorithm based on the MBO scheme \citep{ET} and the proposed local multiphase segmentation algorithm (Algorithm \ref{alg:LMCV}) to the cartoon component of each image in Figure \ref{fig:STMimages}. Convergence for both methods is reached when both phase fields $u_1$ and $u_2$ do not change within the same 
iteration. Otherwise, the maximum number of iterations is set at 200. For each image, the values for $\lambda$ and  $\mu$ are set the same between both methods, but the time step  $dt$ might be different because both methods are minimizing different energy functionals. Finally, the standard deviation for the Gaussian filter in Algorithm~\ref{alg:LMCV} is set to 10.

Here, we discuss the parameter selection for both the multiphase and local multiphase methods. In our experiments, the 
values of $\lambda$ in \eqref{eq:CV} are set to 5, 7, or 10. If there are multiple regions or regions of varying sizes in the image, the value $\mu$ should be small.  Otherwise, it should be large. For our images of size $255 \times 255$, we adopt four values for $\mu$: $10^{-4} \times 255^2$, $10^{-3} \times 255^2$, $10^{-2} \times 255^2$, and $10^{-1} \times 255^2$. The parameter $\beta$ from \eqref{eq:CVloc} is set as the same value as $\lambda$ if there is intensity homogeneity. Otherwise, if there is intensity inhomogeneity, $\beta$ is set to be larger than $\lambda$ to ensure that potential regions with weak edges are detected. To ensure $\Gamma$-convergence of \eqref{eq:CVloc-eps} when running Algorithm \ref{alg:LMCV}, the time step $dt$ needs to be small (we set  here $dt < 20$), and the number of iterations needs to be large, which is why the maximum is set to 200.

The results of each method are shown along with the original and the cartoon images in Figures~\ref{fig:result1} and \ref{fig:result2}. We observe that the results differ between the  two methods. In general, local multiphase is robust against illumination bias and intensity inhomogeneity, thus leading to segmentation results that look similar to the cartoon  images. On the other hand, multiphase produces results with artificial regions, \textit{e.g.}, Figure~\ref{fig:result1}a-c, or results that ignore weak-edged regions, \textit{e.g.}, Figure~\ref{fig:result1}d and Figure~\ref{fig:result2}a-d. 

In Figure~\ref{fig:result1}a, the multiphase result detects the streak of shadow in the middle of the image along with an edge as a region. However, these should not appear as distinct regions at all. On the other hand, the segmentation result of the local multiphase result works properly. Moreover, it detects smaller and more subtle regions of the cartoon image, which are located in the top left and bottom right. 

Figure~\ref{fig:result1}b is heavily affected by illumination bias. Hence, we have a poor segmentation result from the multiphase algorithm. On the other hand, the result from local multiphase is able to identify the natural regions of its corresponding cartoon image, thus appearing extremely similar to it.

In Figure~\ref{fig:result1}c-d, both multiphase and local multiphase produce highly similar results. The results capture regions corresponding to the different light intensities in their corresponding cartoon images. In Figure~\ref{fig:result1}c, the results for both multiphase and local multiphase look similar to their cartoon image. In Figure~\ref{fig:result1}d, the segmentation results by multiphase and local multiphase appear to be dissimilar to the cartoon image, but upon closer inspection, the results detect regions of similar intensities that might be difficult to discern from the cartoon image. 

In Figure~\ref{fig:result2}a, the local multiphase result resembles the cartoon image more than does the multiphase result. The local multiphase result is able to capture the \say{island} region on the top right and the small streak at the very bottom of the image. The multiphase result is unable to capture the small streak and it attempts to segment the \say{island} region. 

In Figure~\ref{fig:result2}b, unlike multiphase result, the local multiphase result is able to identify the oval region at the right of the image. The oval region has weak edges because of the apparent intensity inhomogeneity, which the multiphase algorithm is unable to detect it. 

As for Figure~\ref{fig:result2}c, both multiphase and local multiphase results appear similar to the cartoon image, but local multiphase is able to identify the small \say{islands} in the middle of the image. The multiphase result tends to identify larger regions. but for one or two regions of the result, some parts of one region do not have similar gray-level intensities according to the cartoon image.

In Figure~\ref{fig:result2}d, both results provide a segmentation based on the gray-level intensity. Local multiphase is able to detect shadows, such as those in the top left corner and at the bottom of the image, with better precision. The multiphase segmentation captures wider transition regions. However, the slow varying vertical edges are not localized by neither of the two methods, in which case, most likely, a nonlocal version of the segmentation algorithms above would help improve the results.

In summary, because of the local term, the local multiphase algorithm provides better segmentation results than does the multiphase algorithm, The local term enables the local multiphase algorithm to capture with better  precision regions with shadows and intensity inhomogeneities. The multiphase algorithm, on the other hand, fails to detect such regions, and its results are significantly worse under the presence of illumination bias and intensity inhomogeneity. Therefore, the local multiphase segmentation is preferred to the multiphase segmentation in the case of STM images.

\subsection{Texture Segmentation Results}\label{subsec:texture}

\begin{figure*}[!t]
\begin{center}
\begin{tabular}{m{1mm}>{\centering\arraybackslash}>{\centering\arraybackslash}m{0.2\textwidth}>{\centering\arraybackslash}m{0.2\textwidth}>{\centering\arraybackslash}m{
0.2\textwidth}>{\centering\arraybackslash}m{0.2\textwidth}}
\centering
& \textbf{original} & \textbf{texture} & \textbf{$k$-means} & \textbf{multiclass MBO} \\
\textbf{(a)}& 
  \includegraphics[width=0.2\textwidth]{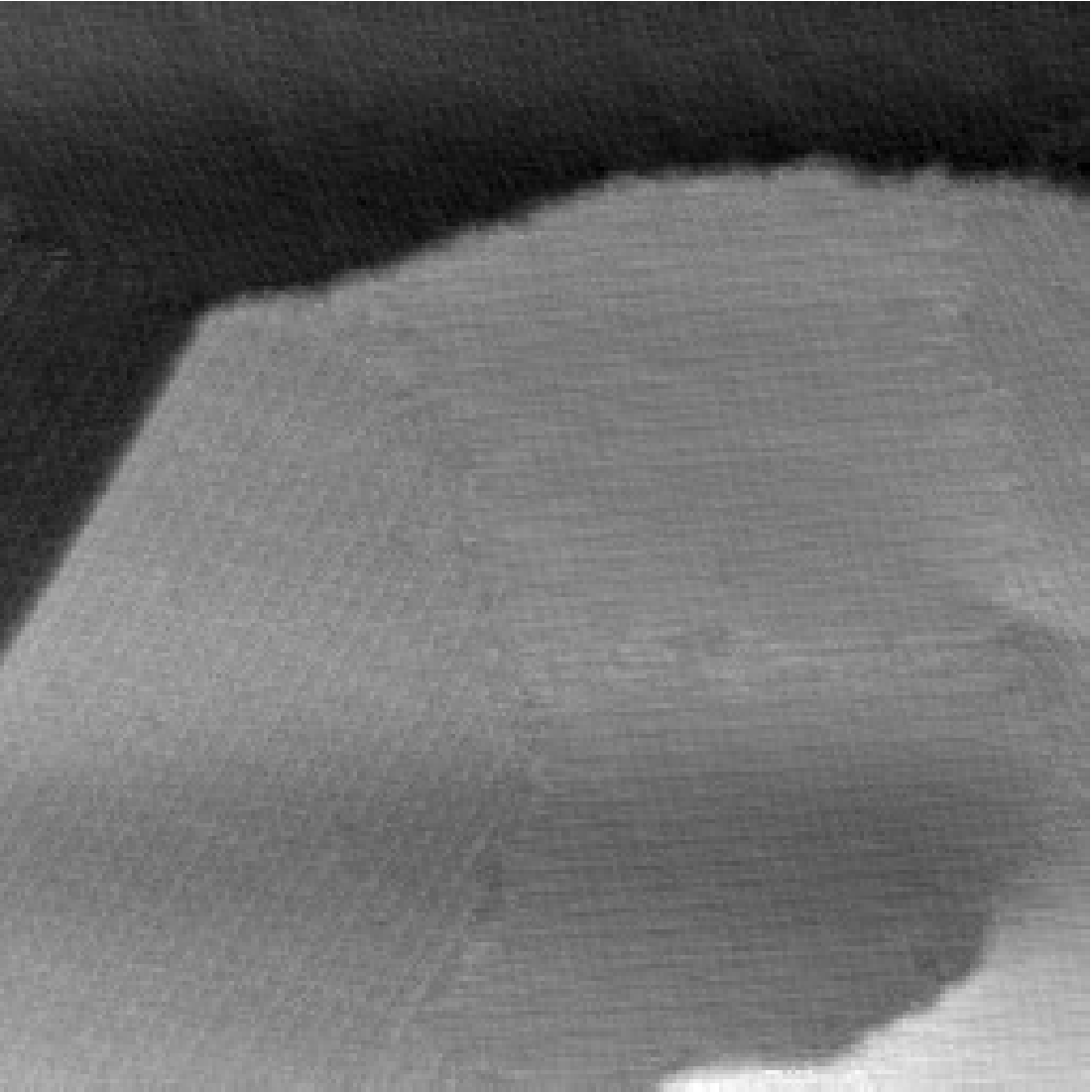} & 
  \includegraphics[width=0.2\textwidth]{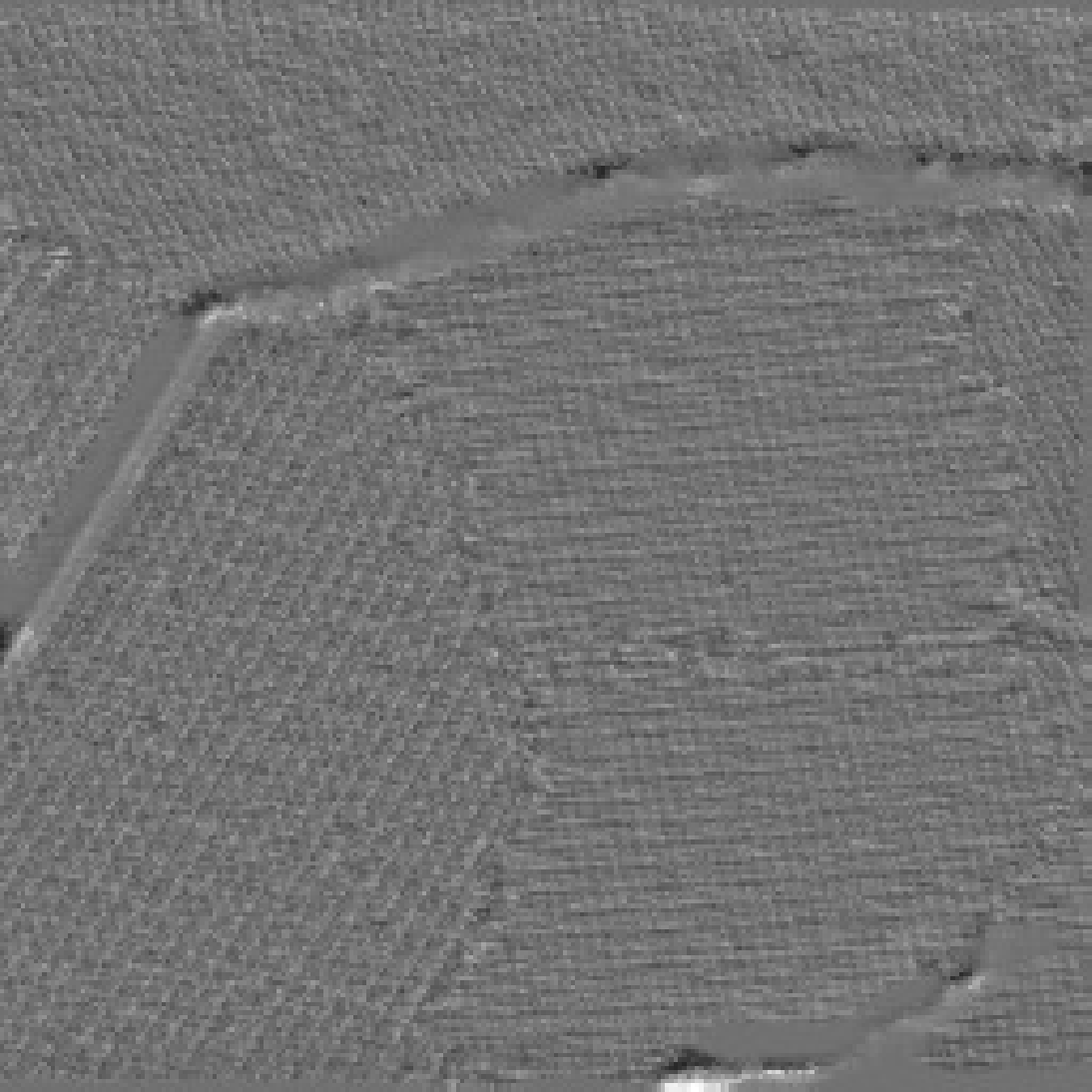}  &  
  \includegraphics[width=0.2\textwidth]{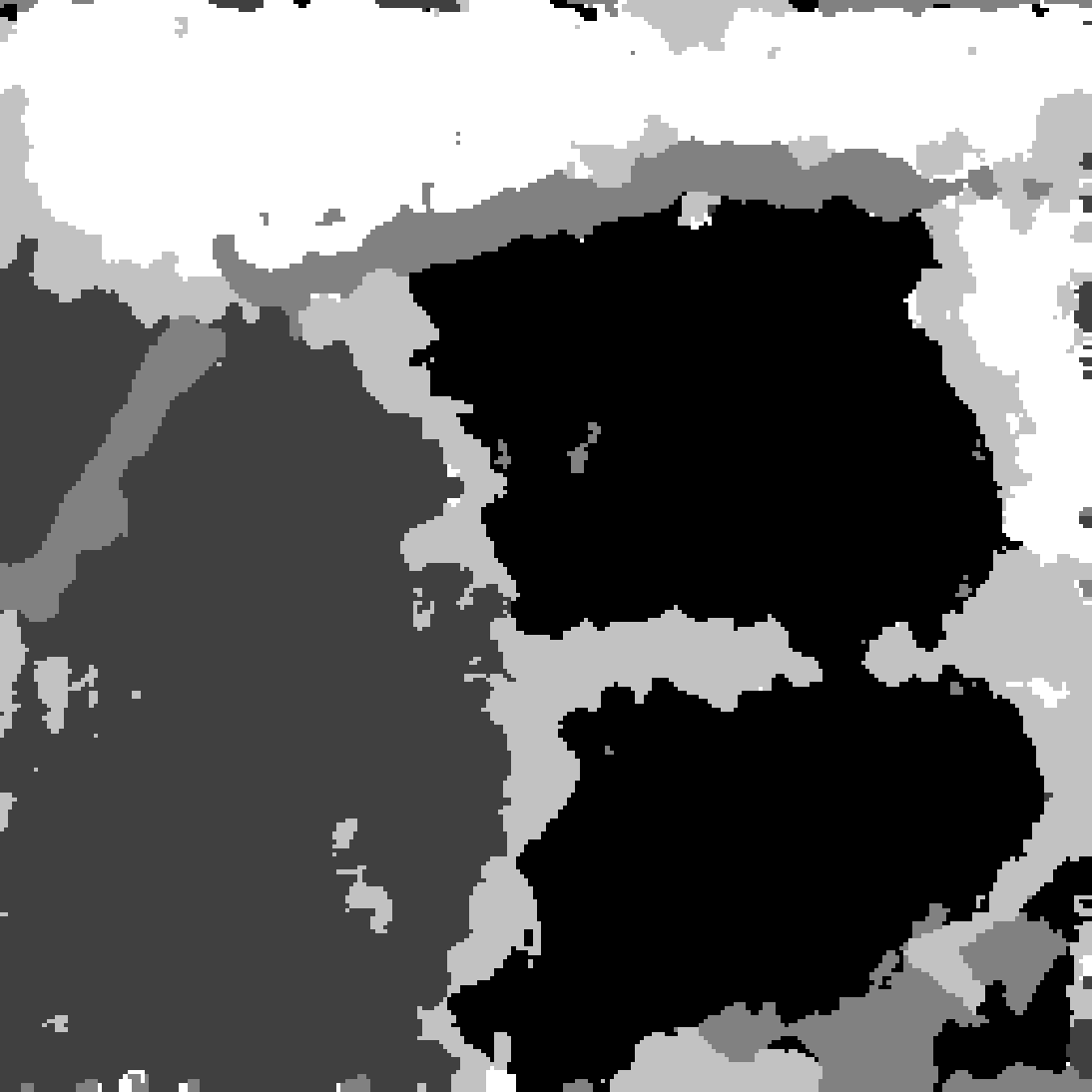} &
  \includegraphics[width=0.2\textwidth]{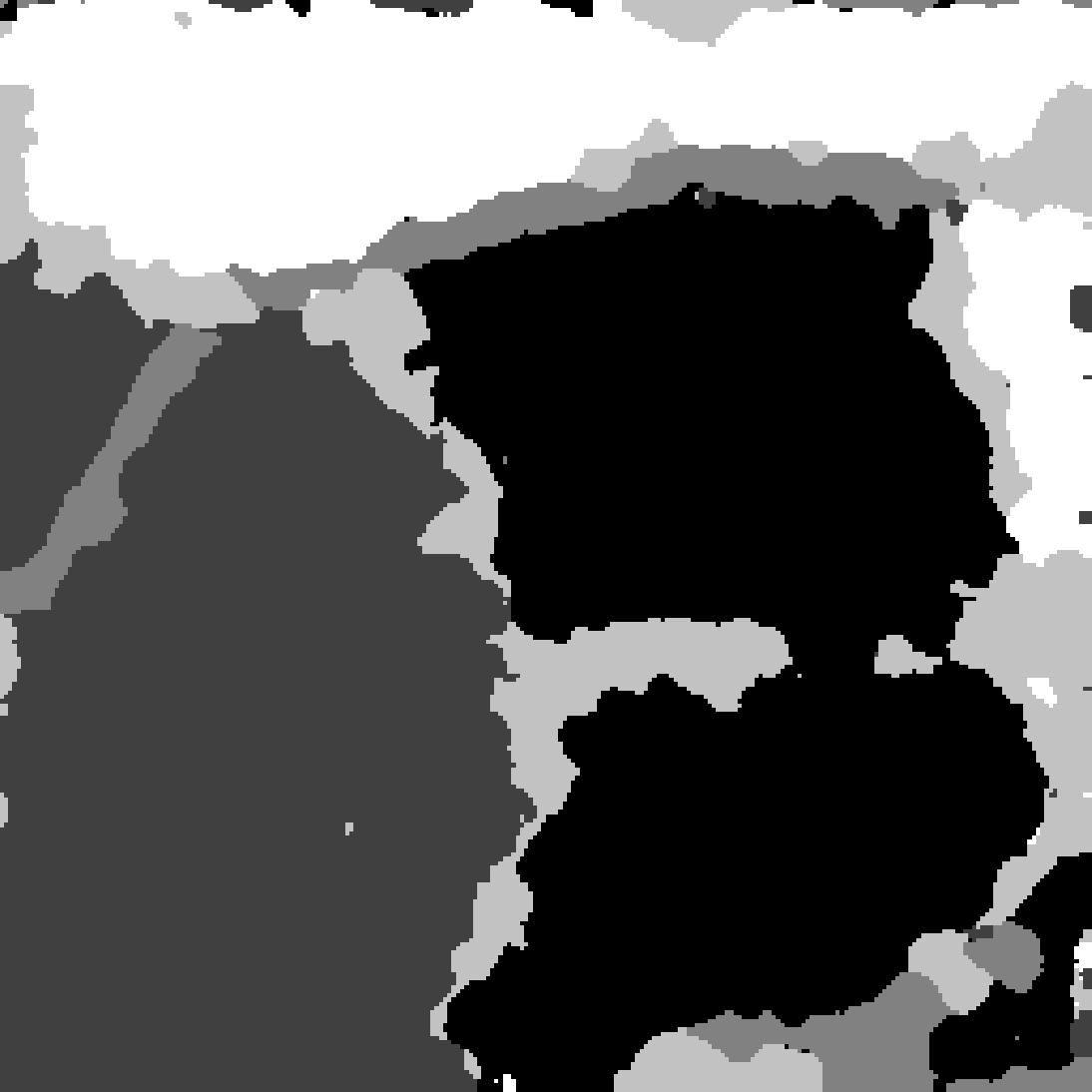} \\
\textbf{(b)}& 
  \includegraphics[width=0.2\textwidth]{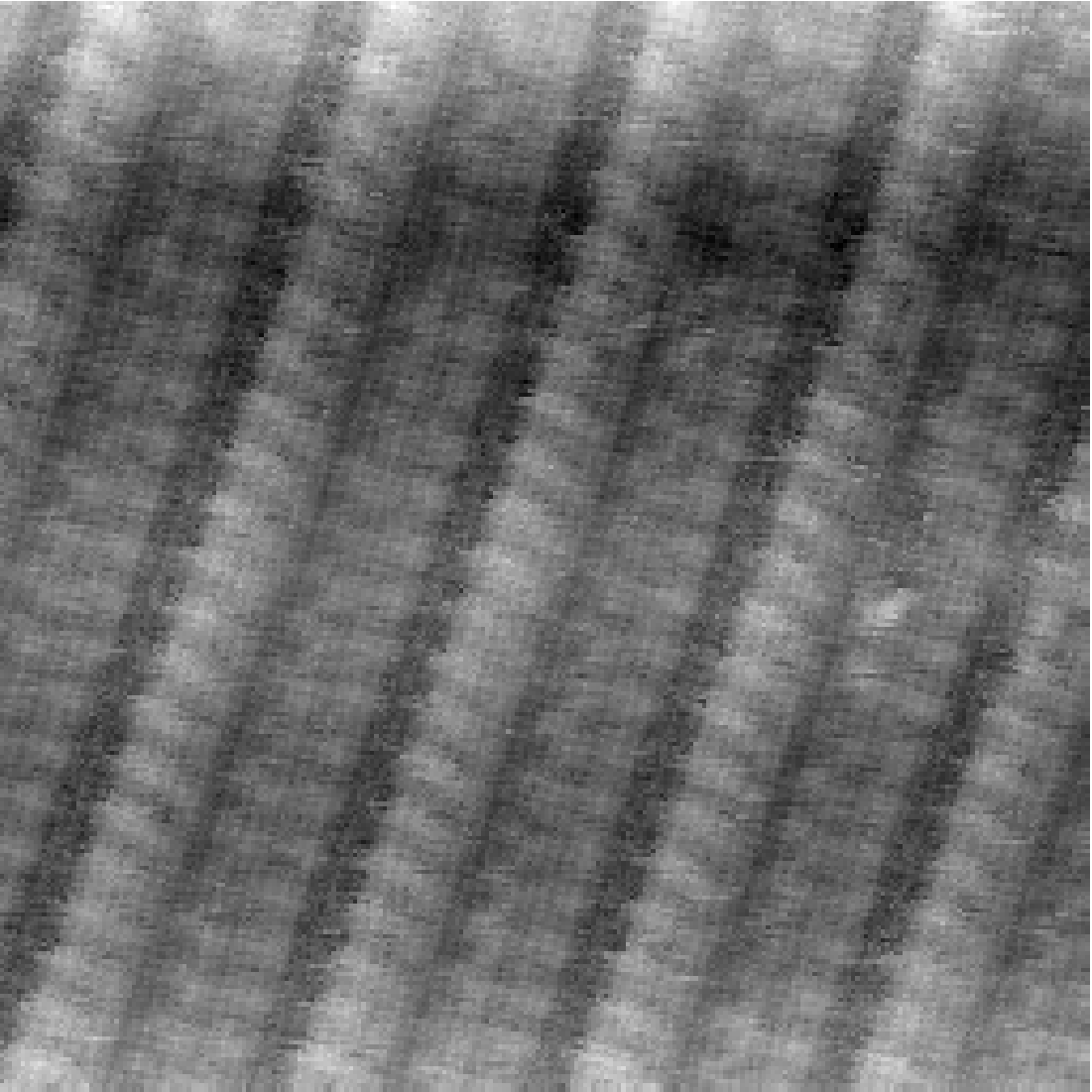} & 
  \includegraphics[width=0.2\textwidth]{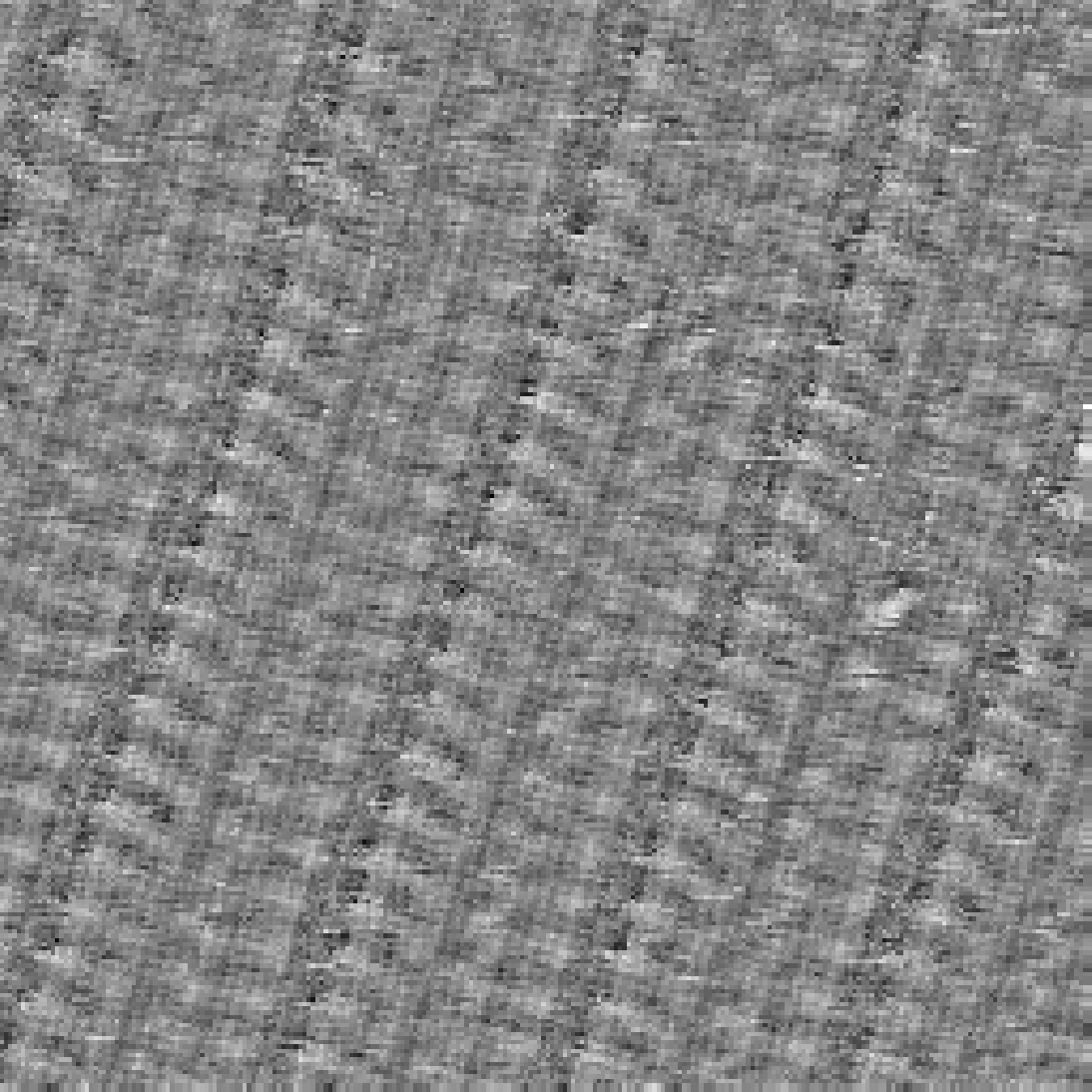}  &  
  \includegraphics[width=0.2\textwidth]{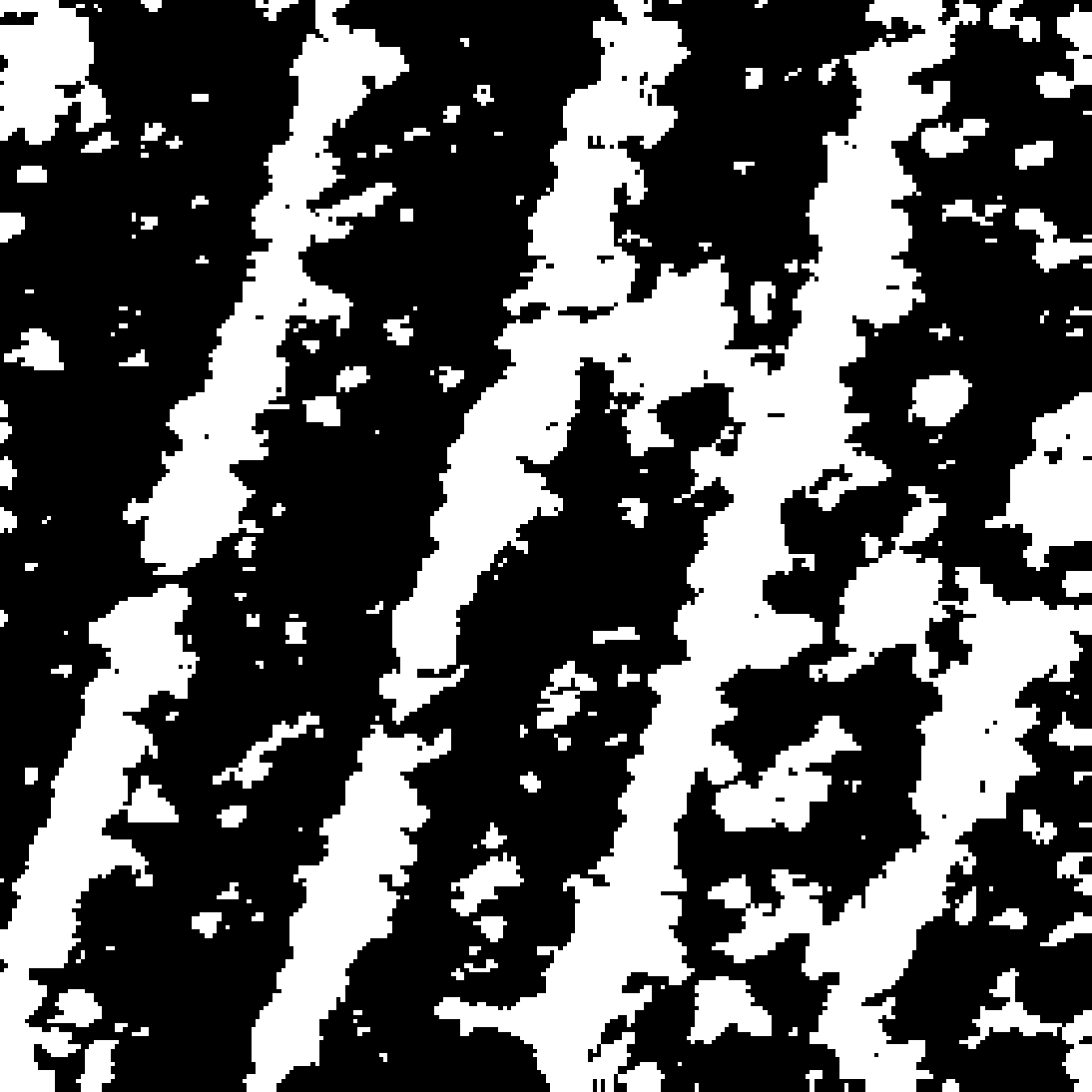} &
  \includegraphics[width=0.2\textwidth]{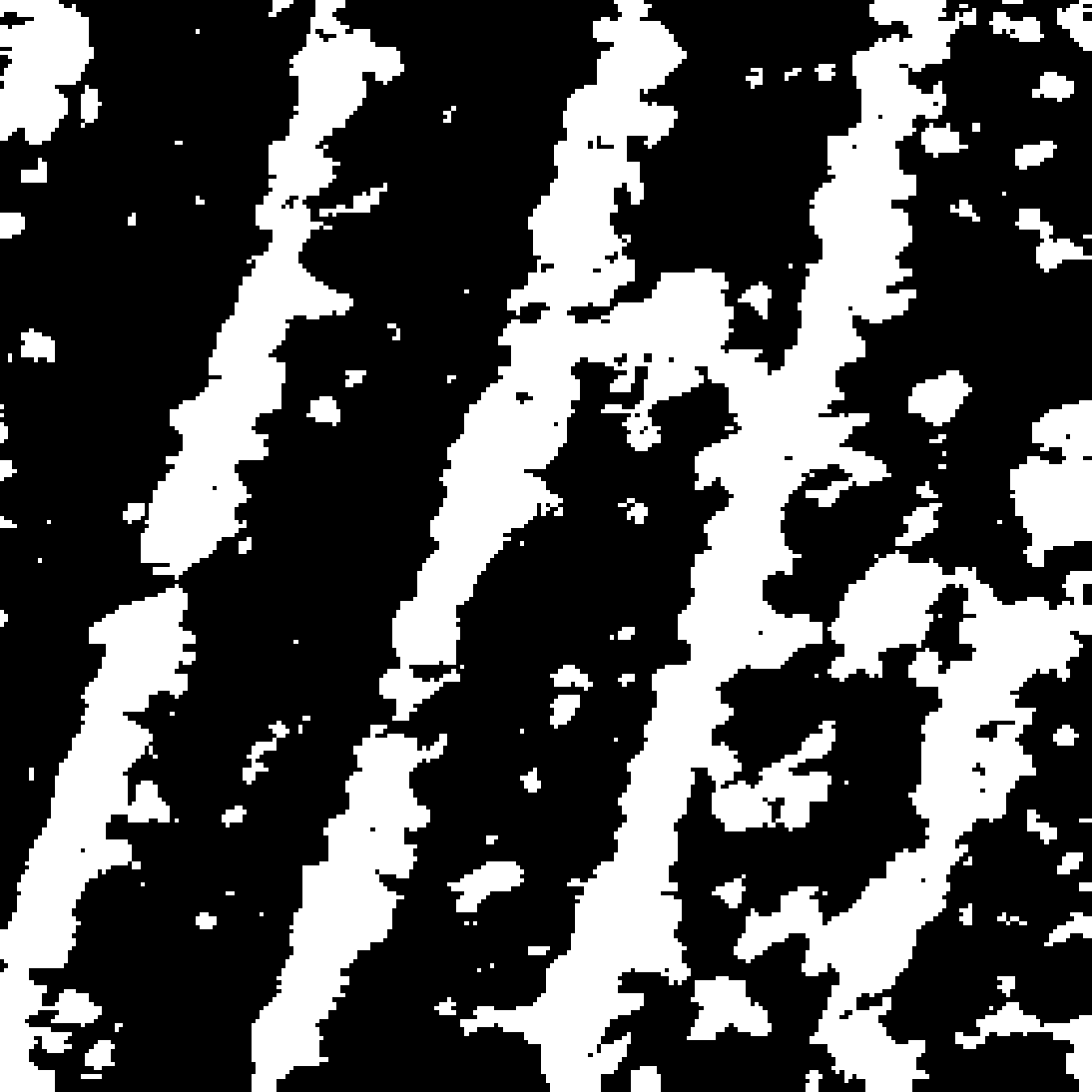} \\
\textbf{(c)}& 
  \includegraphics[width=0.2\textwidth]{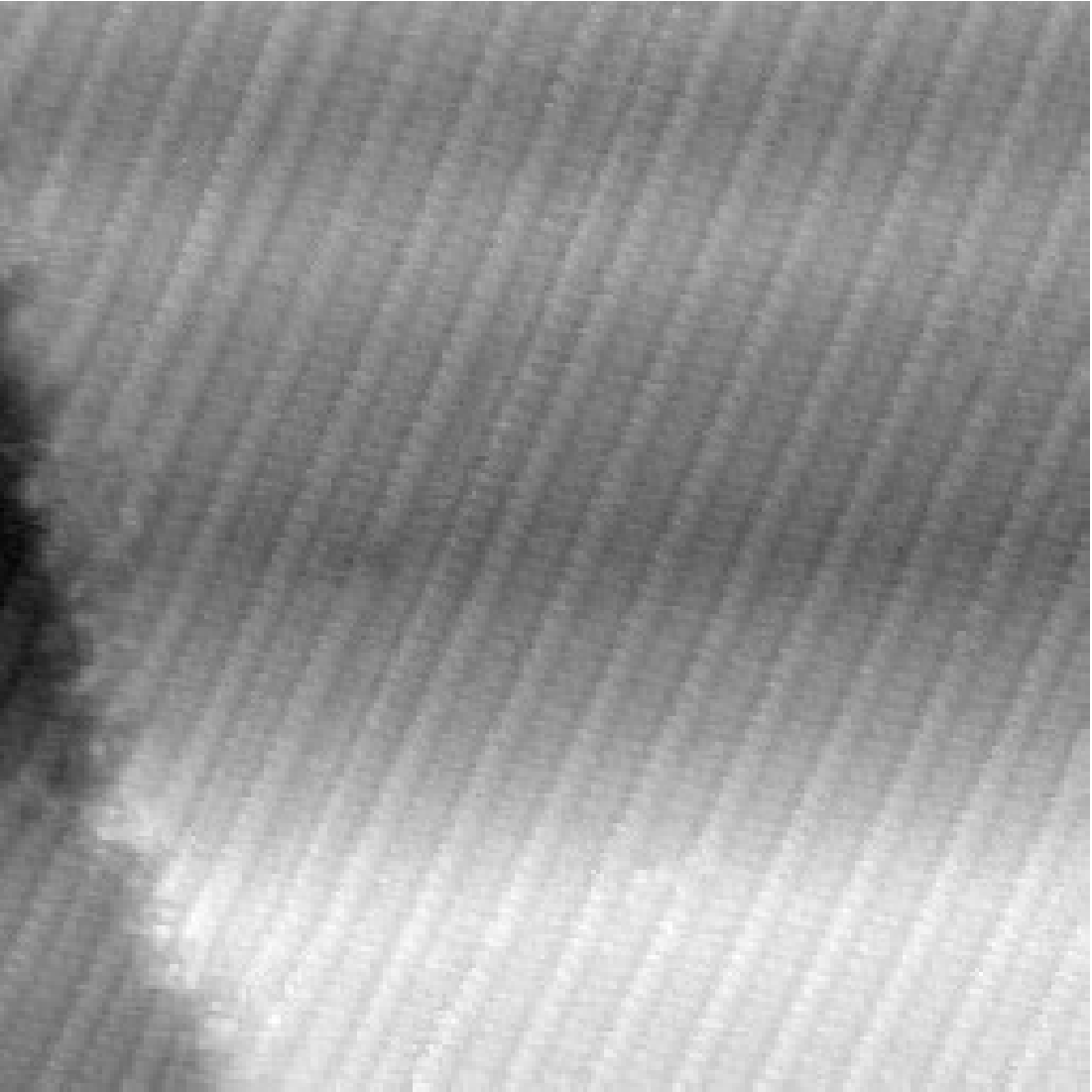} & 
  \includegraphics[width=0.2\textwidth]{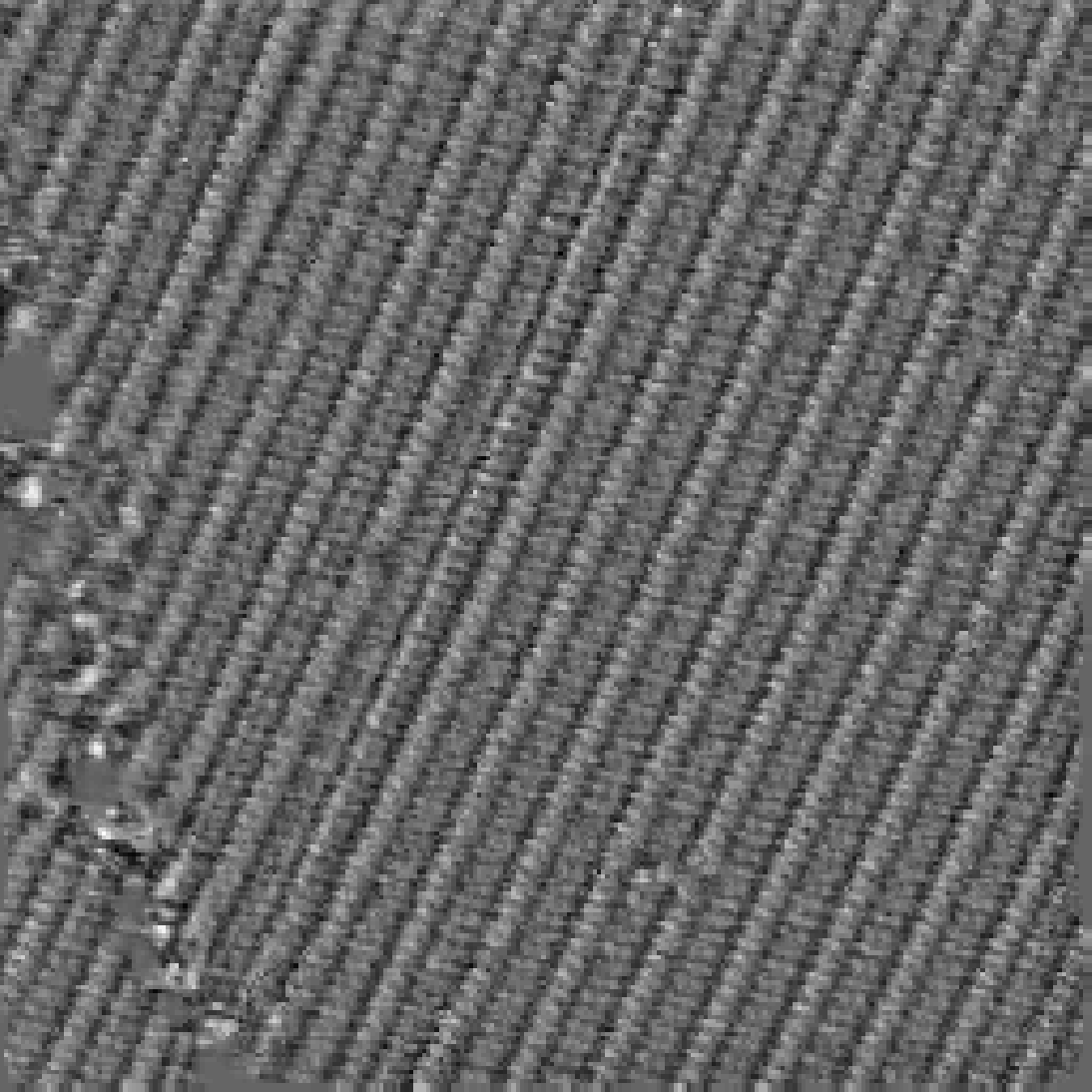}  &  
  \includegraphics[width=0.2\textwidth]{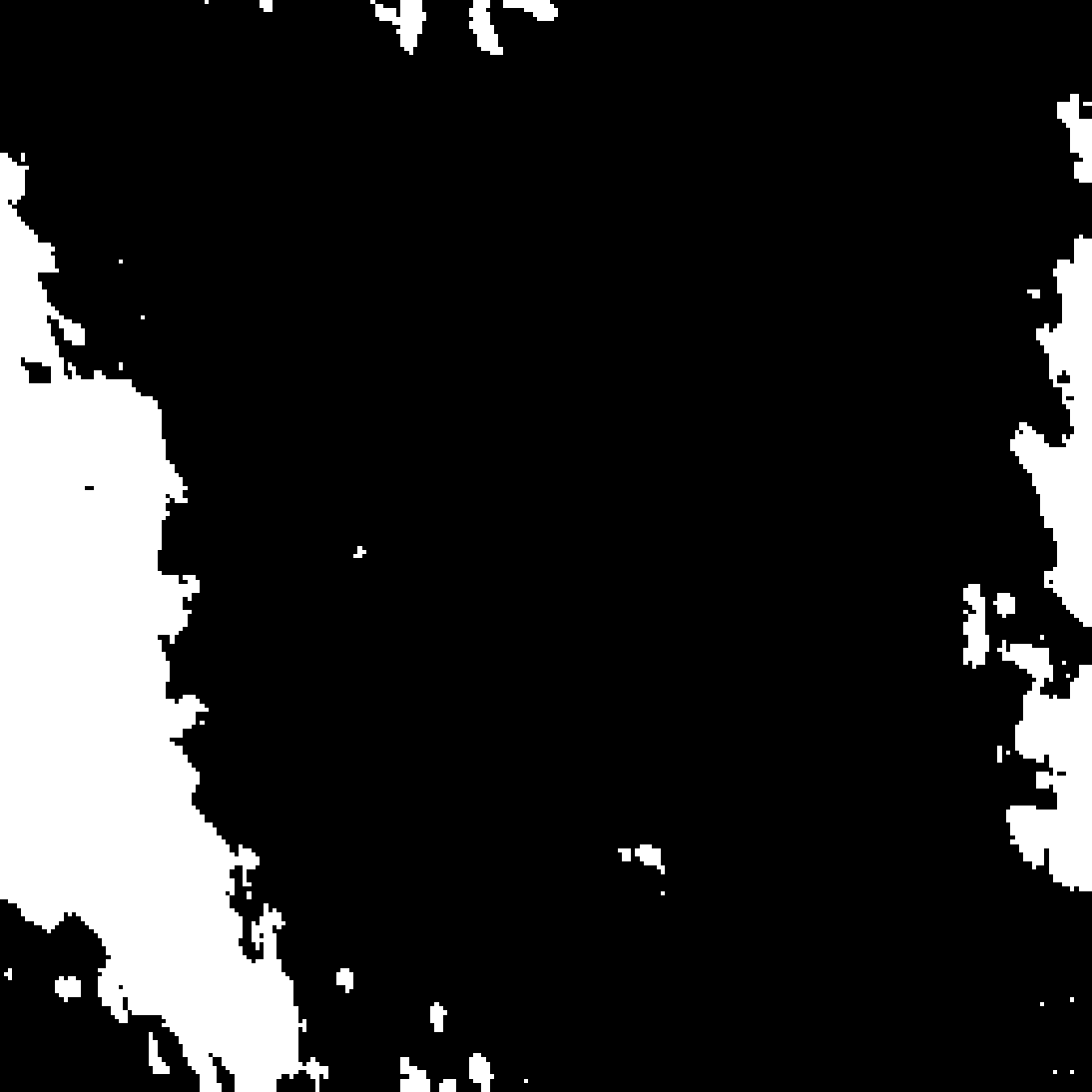} &
  \includegraphics[width=0.2\textwidth]{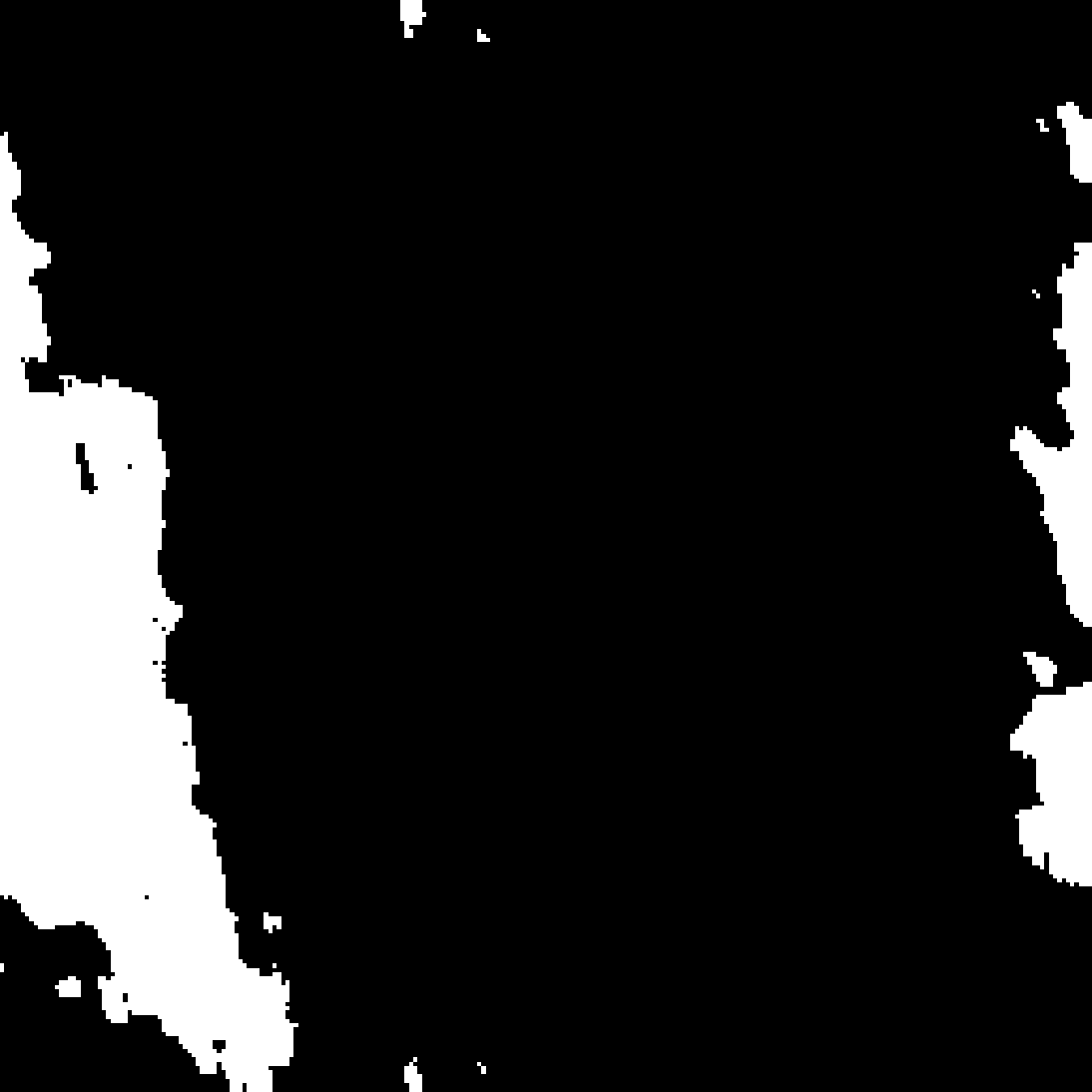} \\
\textbf{(d)}& 
  \includegraphics[width=0.2\textwidth]{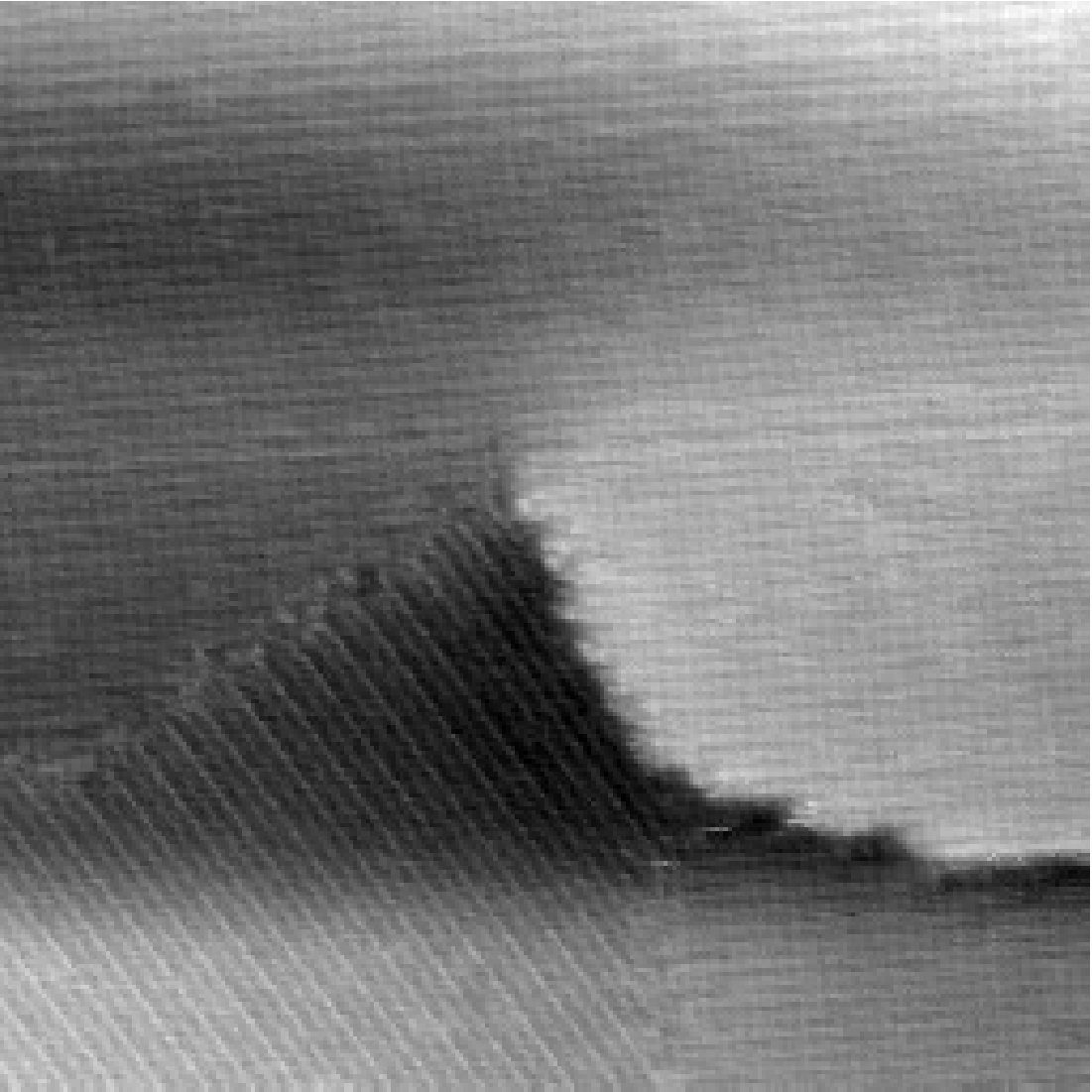} & 
  \includegraphics[width=0.2\textwidth]{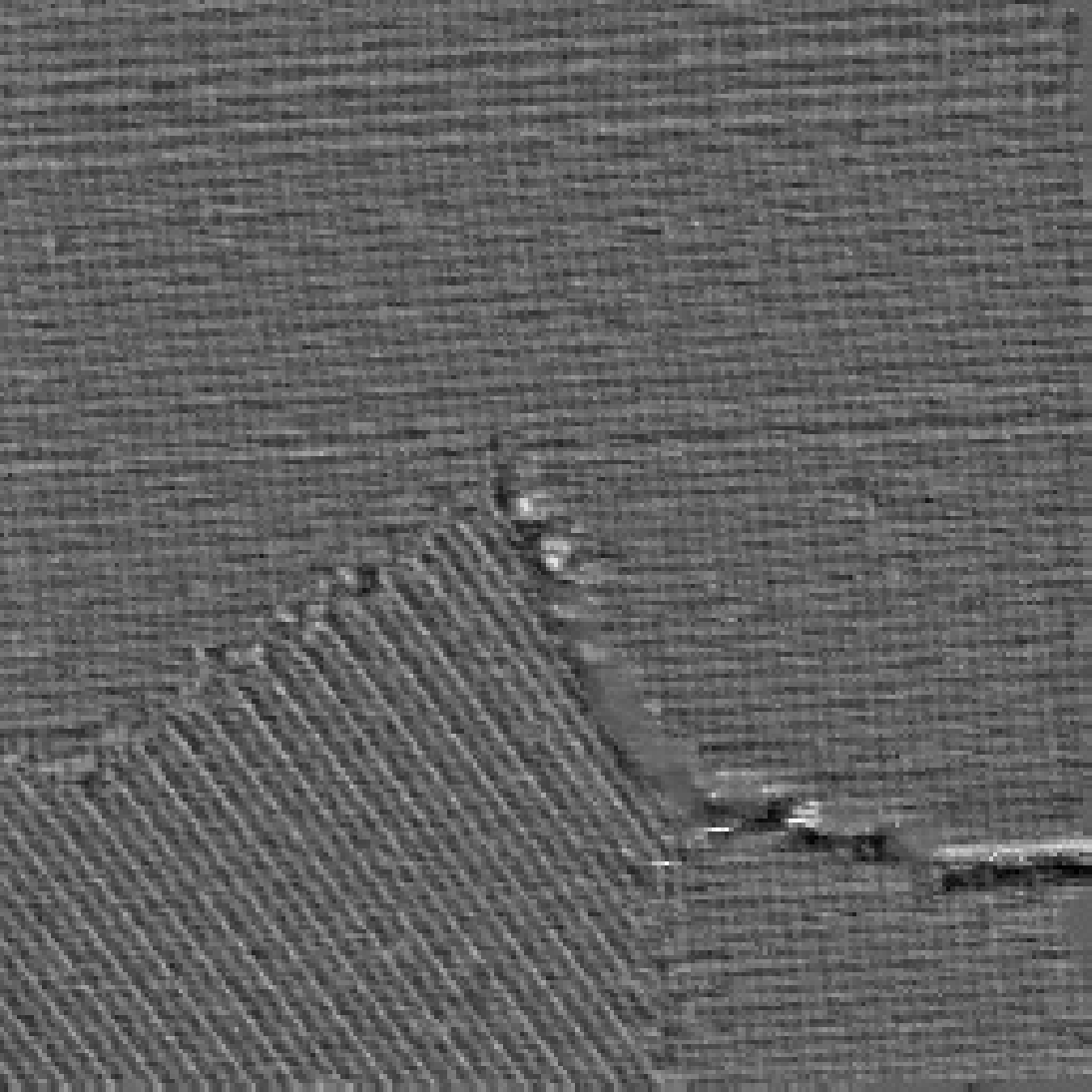}  &  
  \includegraphics[width=0.2\textwidth]{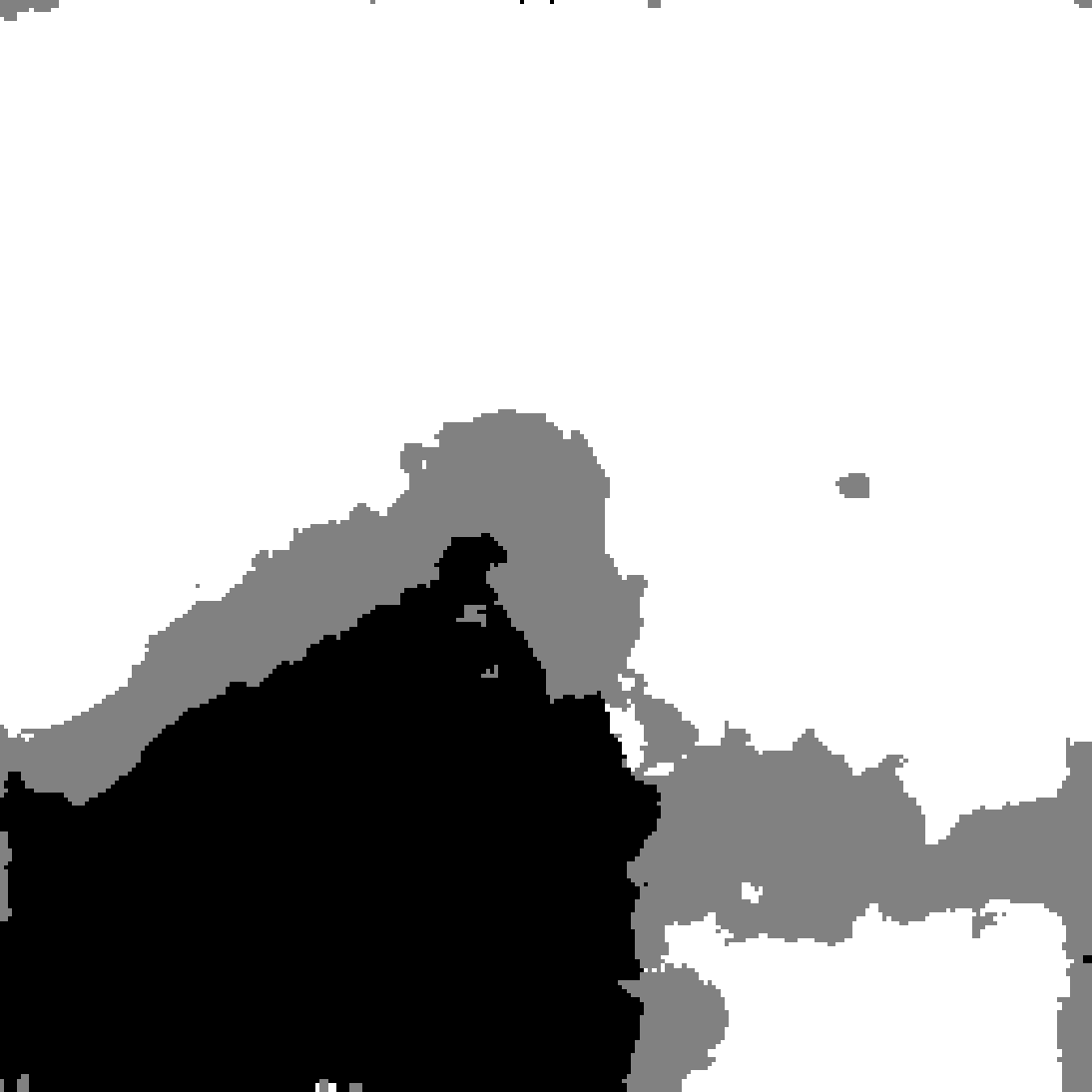} &
  \includegraphics[width=0.2\textwidth]{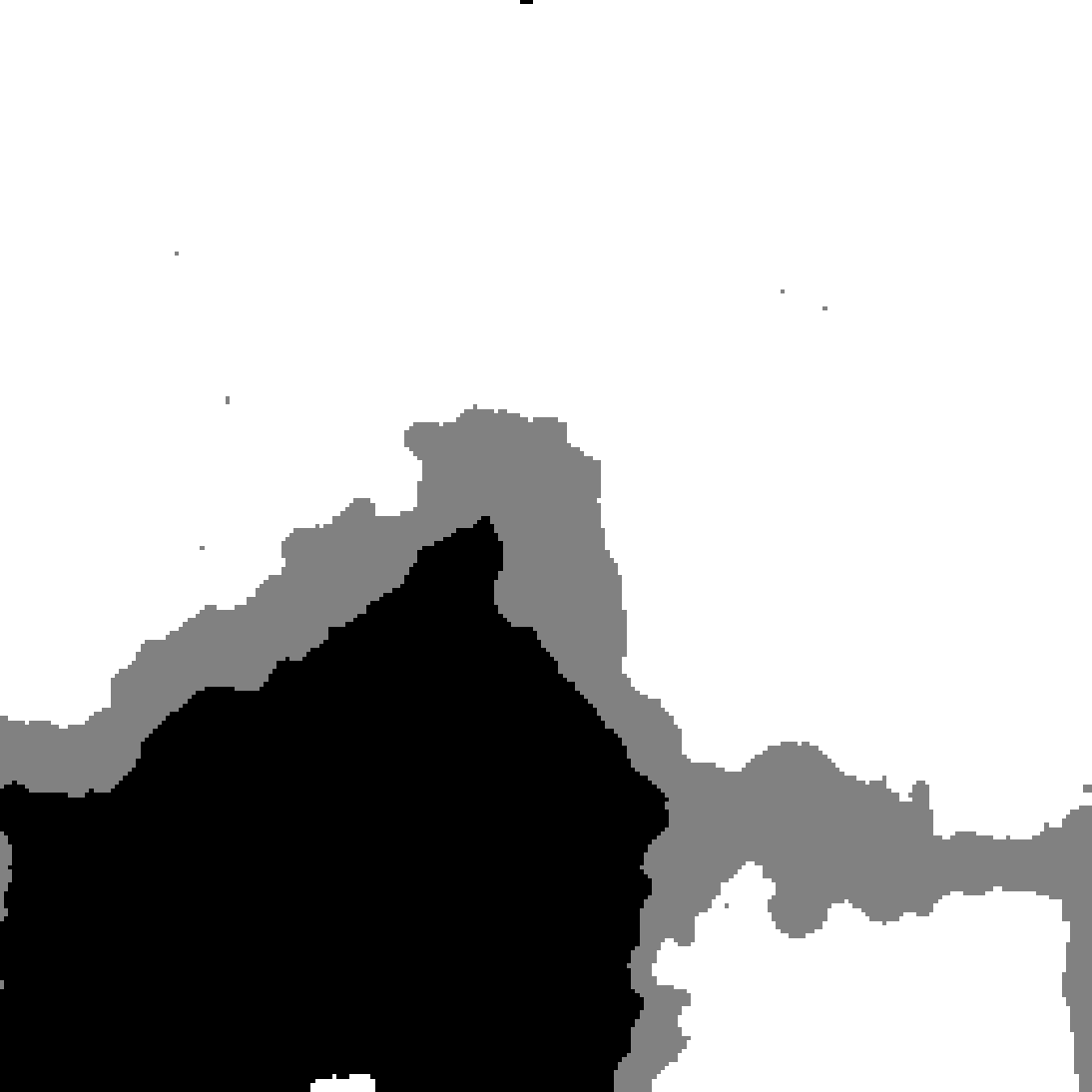} \\
\end{tabular}
\end{center}
\caption{Texture segmentation results. The used parameters are: 
(a) $\tau = 92$nd percentile, $k$ = 5, $dt = 0.03$.  
(b) $\tau = 99.50$th percentile, $k$ = 2, $dt = 0.10$. 
(c) $\tau = 98.80$th percentile, $k = 2$, $dt=0.05$. 
(d) $\tau = 85$th percentile, $k = 3$, $dt = 
0.05$. Raw scanning tunneling microscope images of cyanide on Au\{111\}, reproduced from \cite{guttentag2016hexagons} with permission. Texture segmentation results in (a) were reproduced with permission from \cite{guttentag2016hexagons}. Images and segmentation results copyright American Chemical Society}
\label{fig:result3}
\end{figure*}
\begin{figure*}[!t]
\begin{center}
\begin{tabular}{m{1mm}>{\centering\arraybackslash}>{\centering\arraybackslash}m{0.2\textwidth}>{\centering\arraybackslash}m{0.2\textwidth}>{\centering\arraybackslash}m{
0.2\textwidth}>{\centering\arraybackslash}m{0.2\textwidth}}
\centering
& \textbf{original} & \textbf{texture} & \textbf{$k$-means} & \textbf{multiclass MBO} \\
\textbf{(a)}& 
  \includegraphics[width=0.2\textwidth]{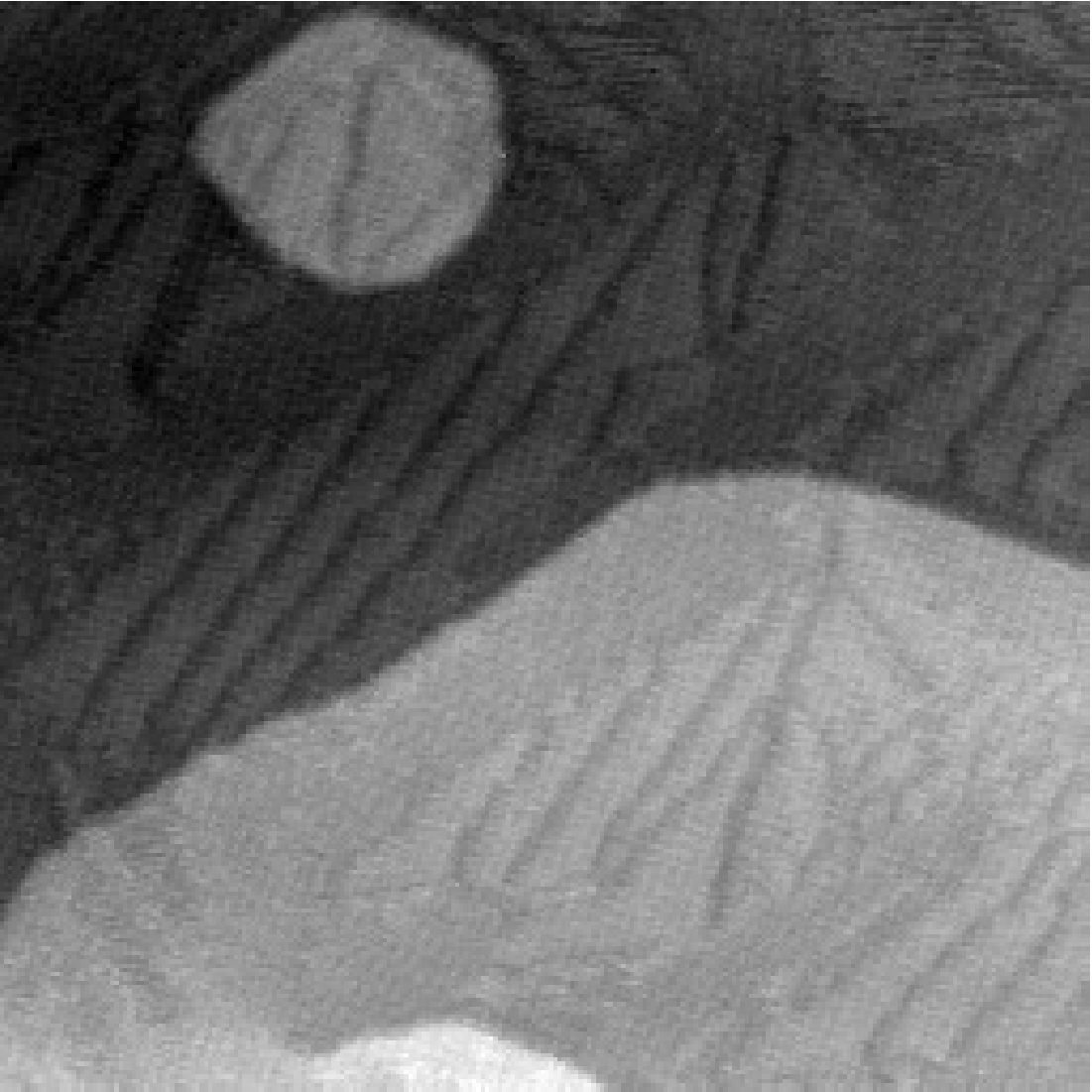} & 
  \includegraphics[width=0.2\textwidth]{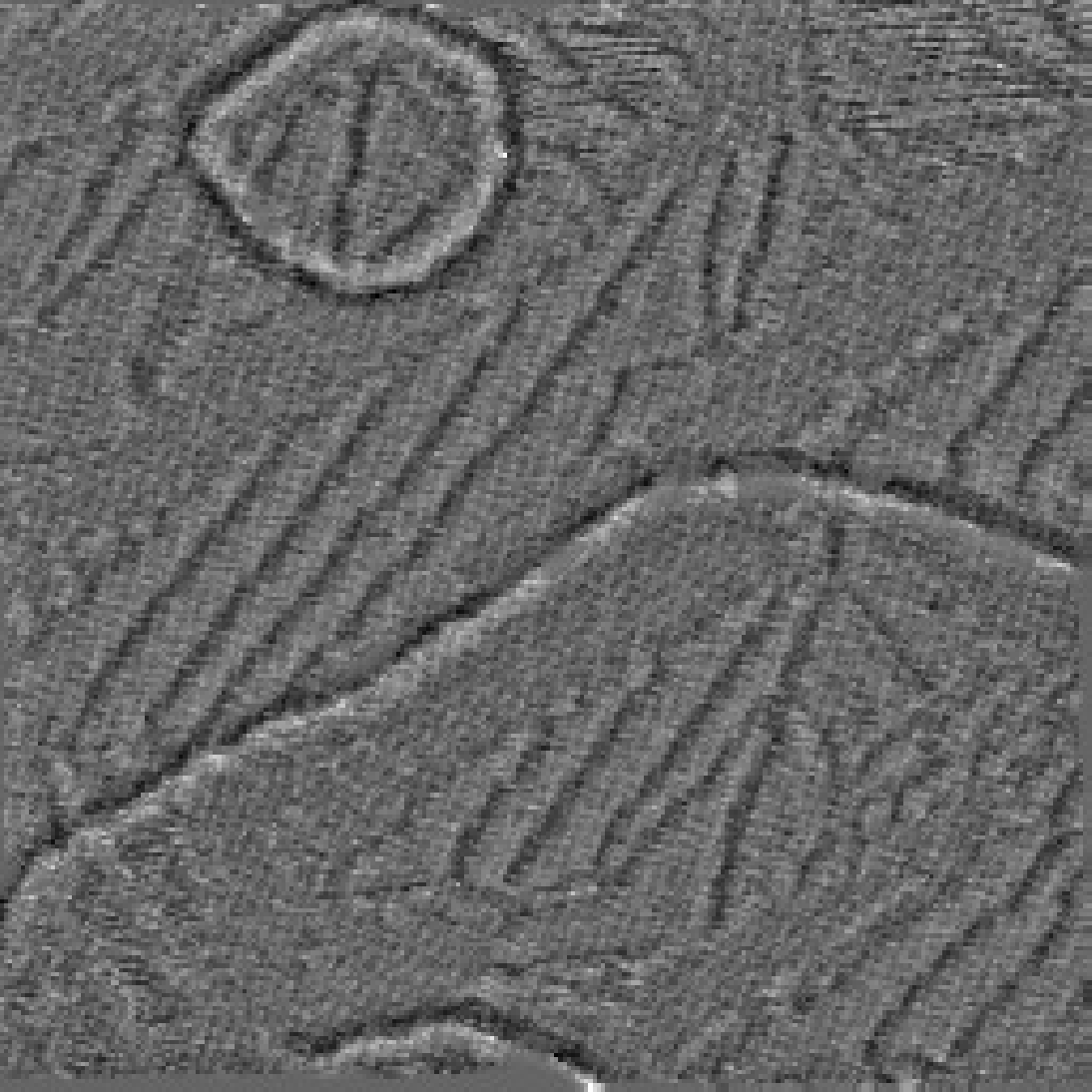}  &  
  \includegraphics[width=0.2\textwidth]{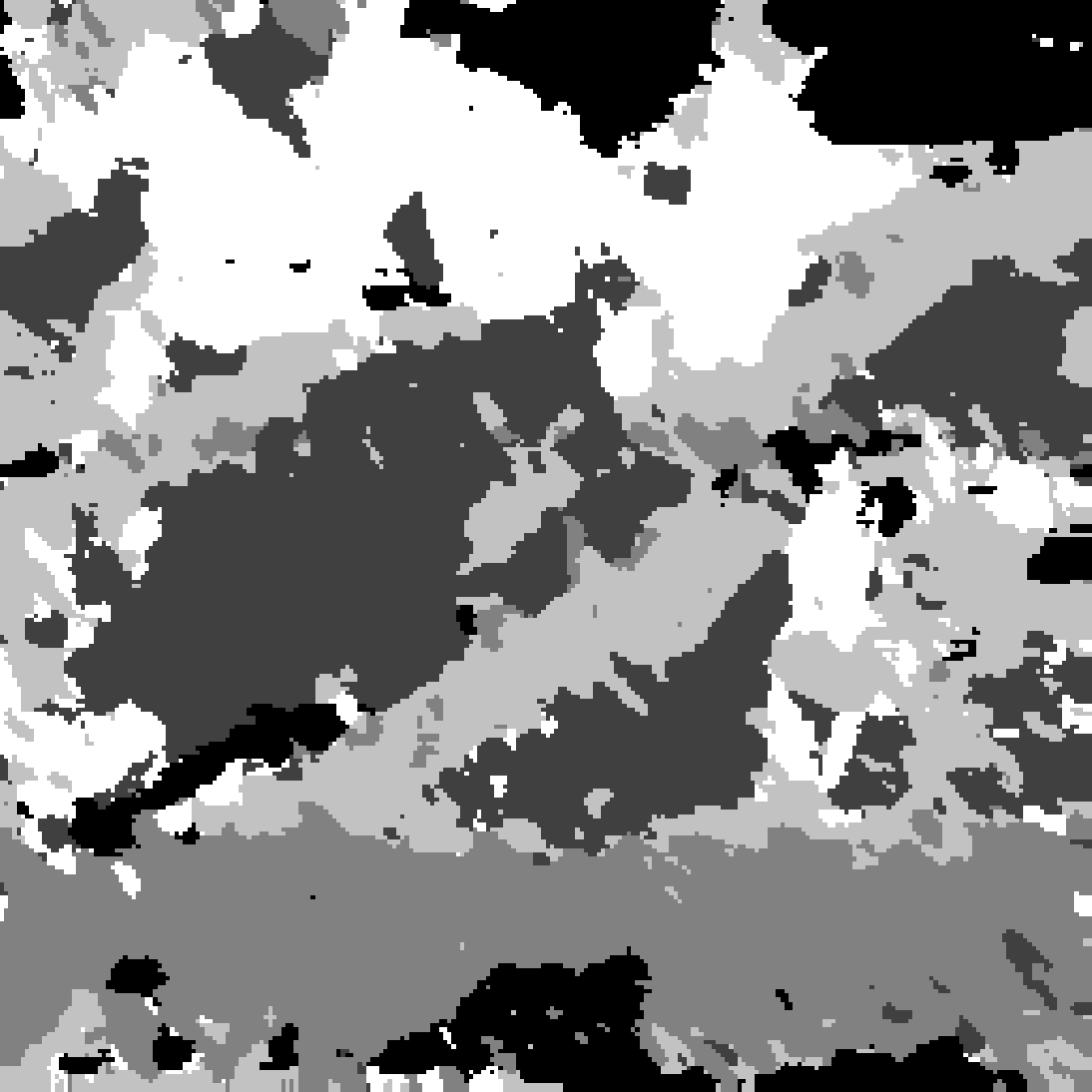} &
  \includegraphics[width=0.2\textwidth]{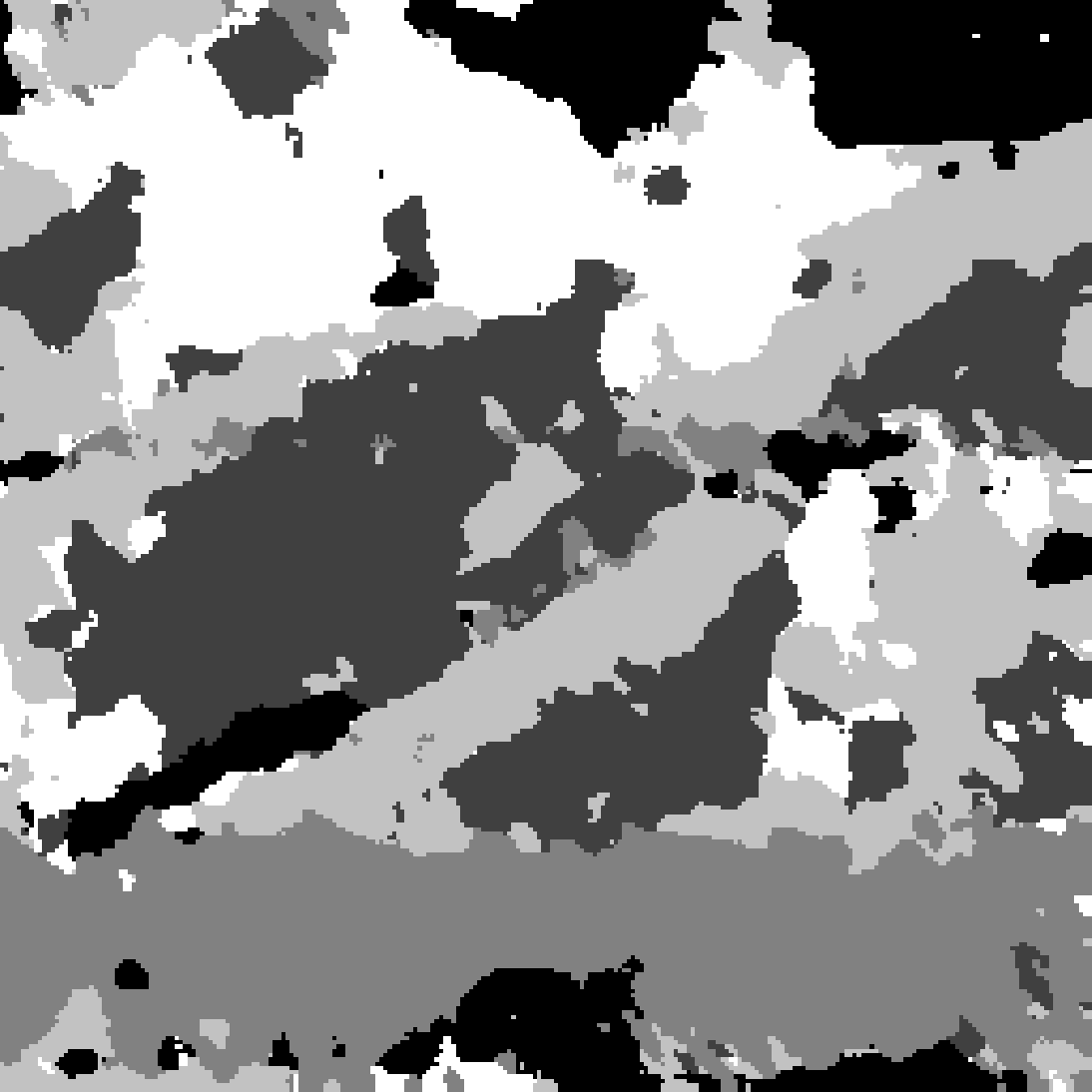} \\
\textbf{(b)}& 
  \includegraphics[width=0.2\textwidth]{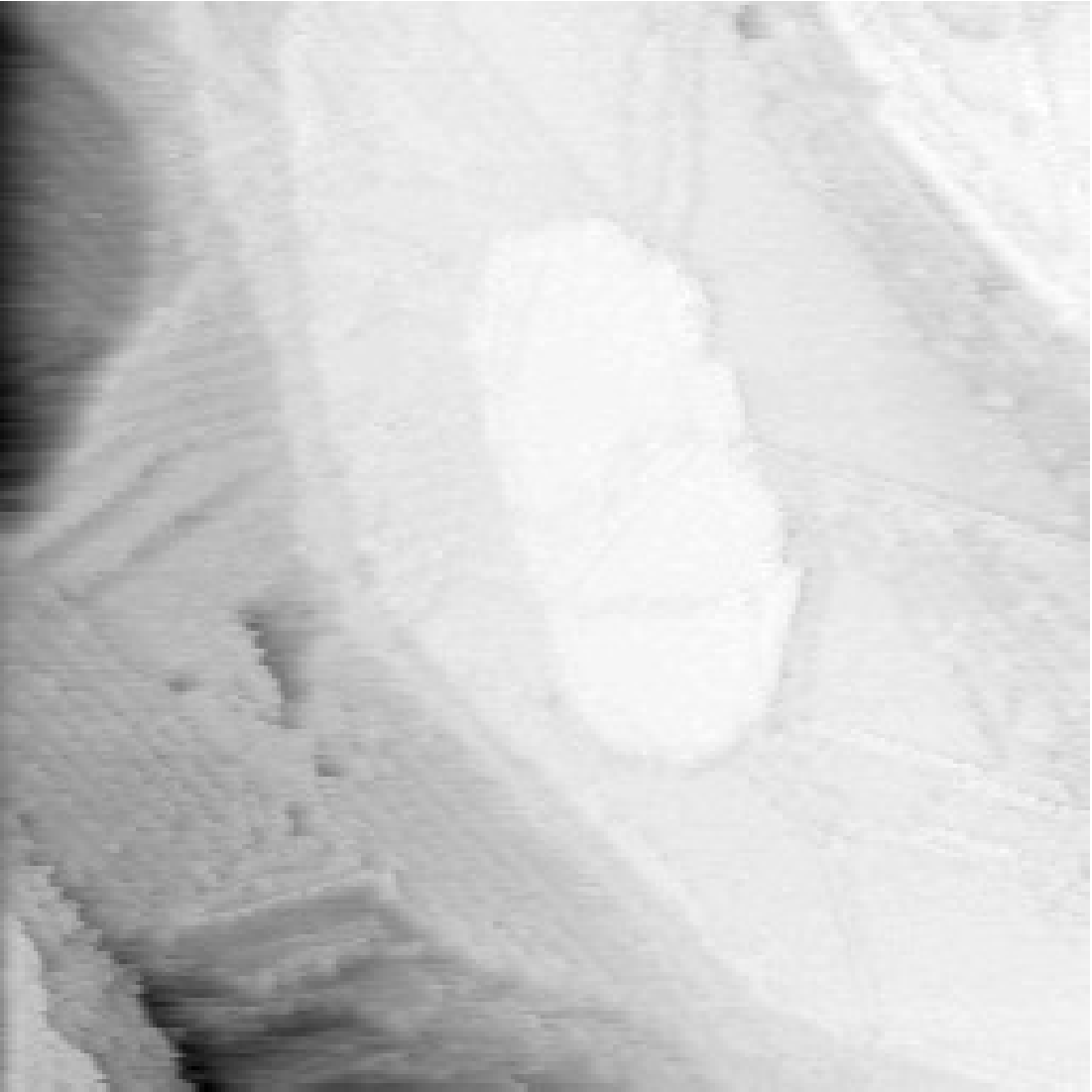} & 
  \includegraphics[width=0.2\textwidth]{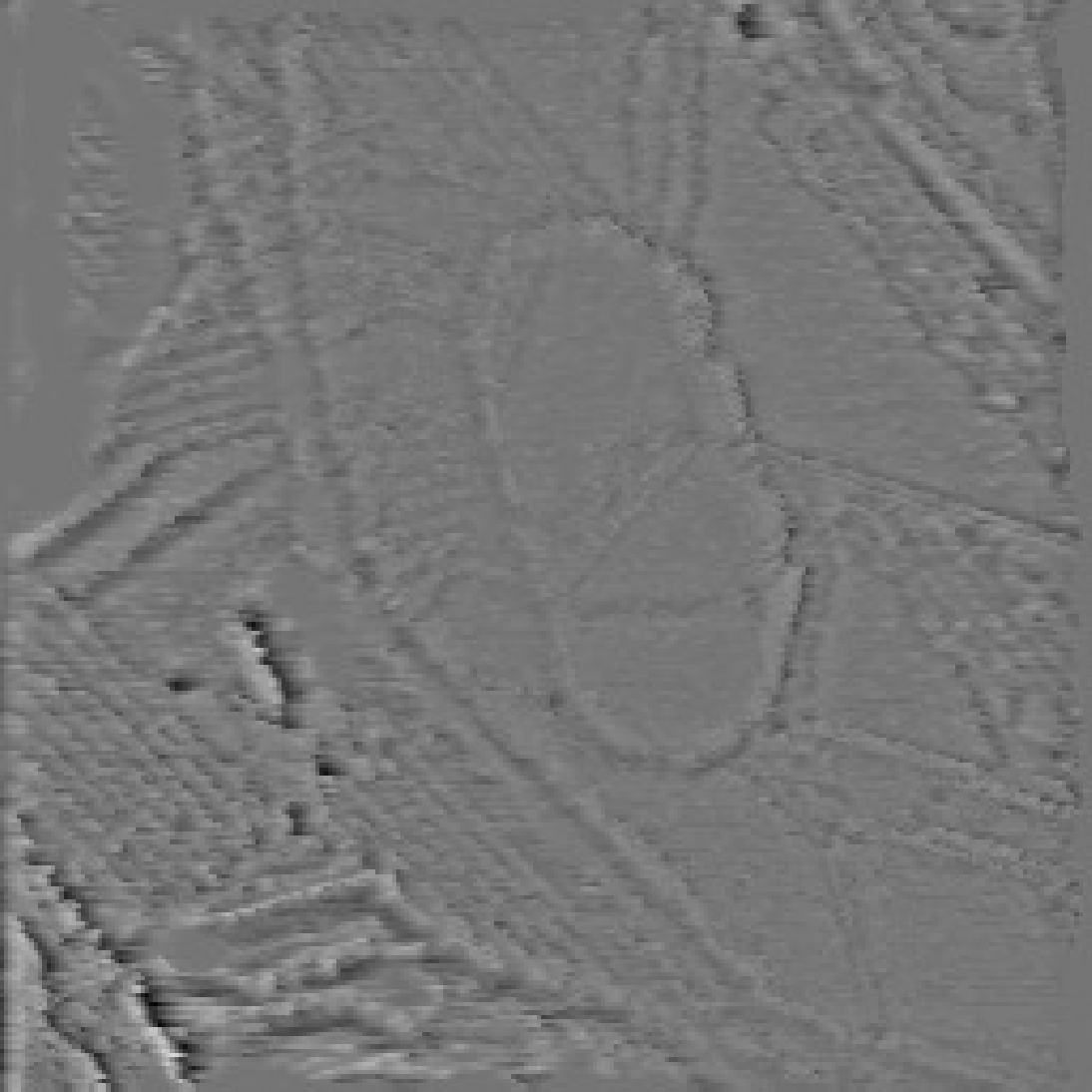}  &  
  \includegraphics[width=0.2\textwidth]{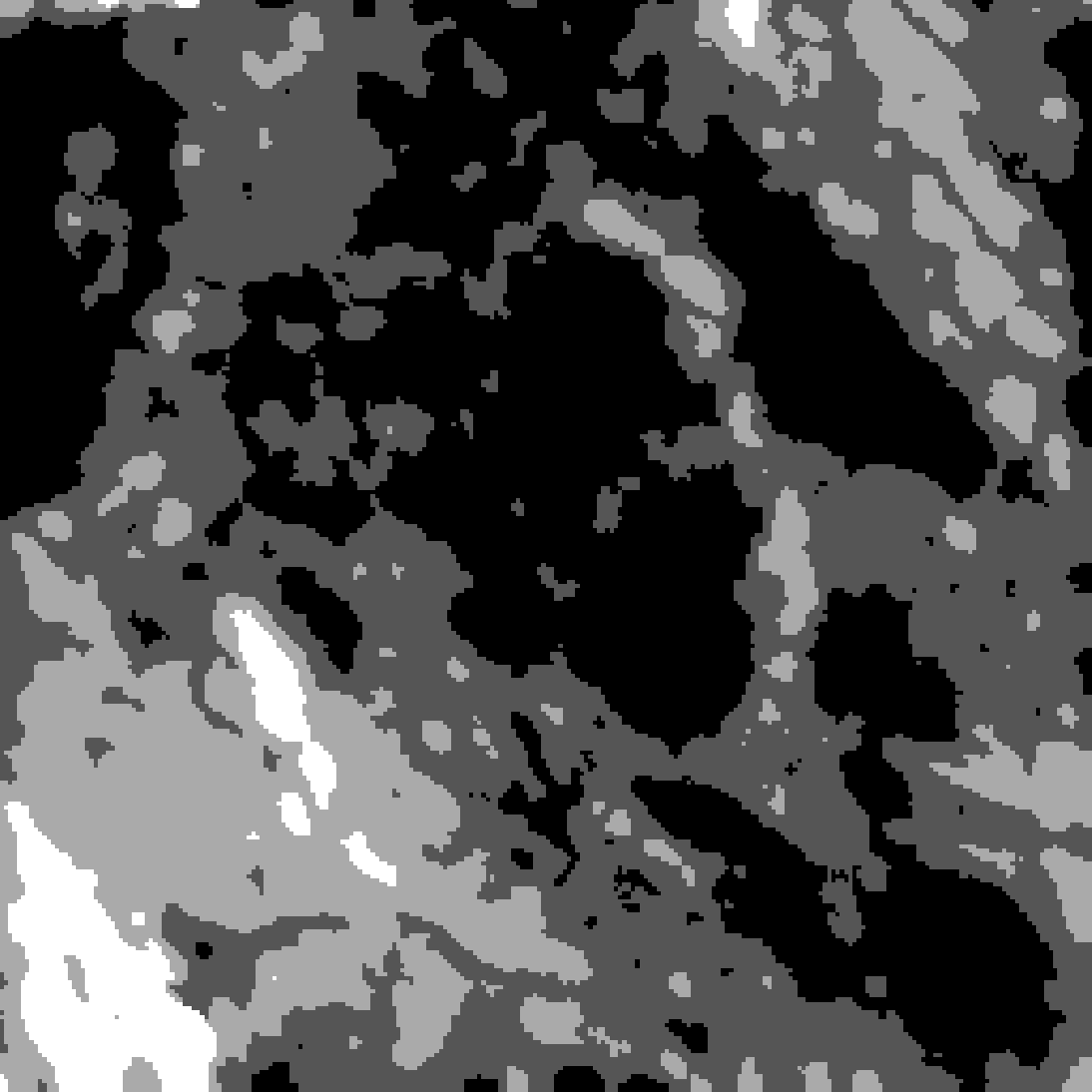} &
  \includegraphics[width=0.2\textwidth]{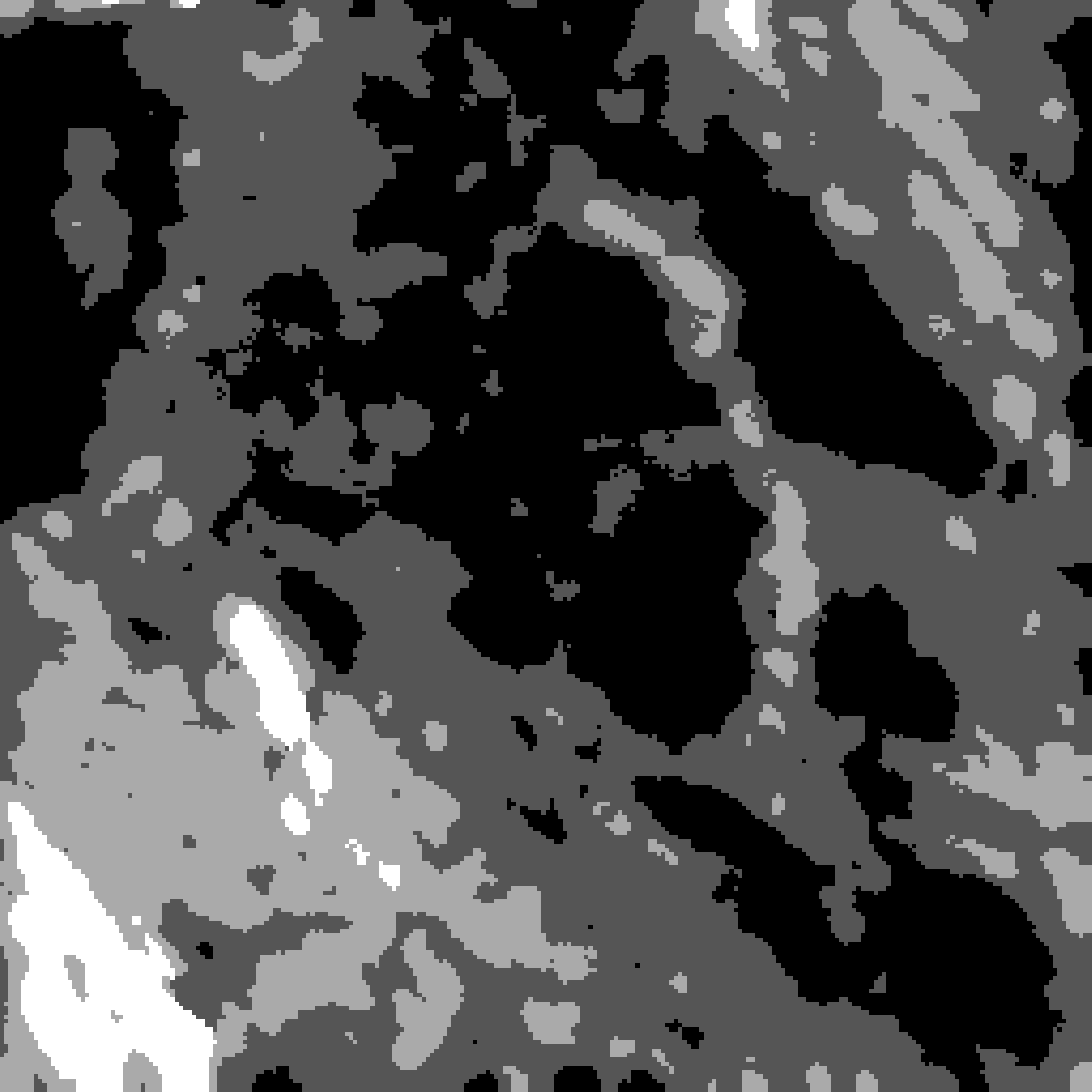} \\
\textbf{(c)}& 
  \includegraphics[width=0.2\textwidth]{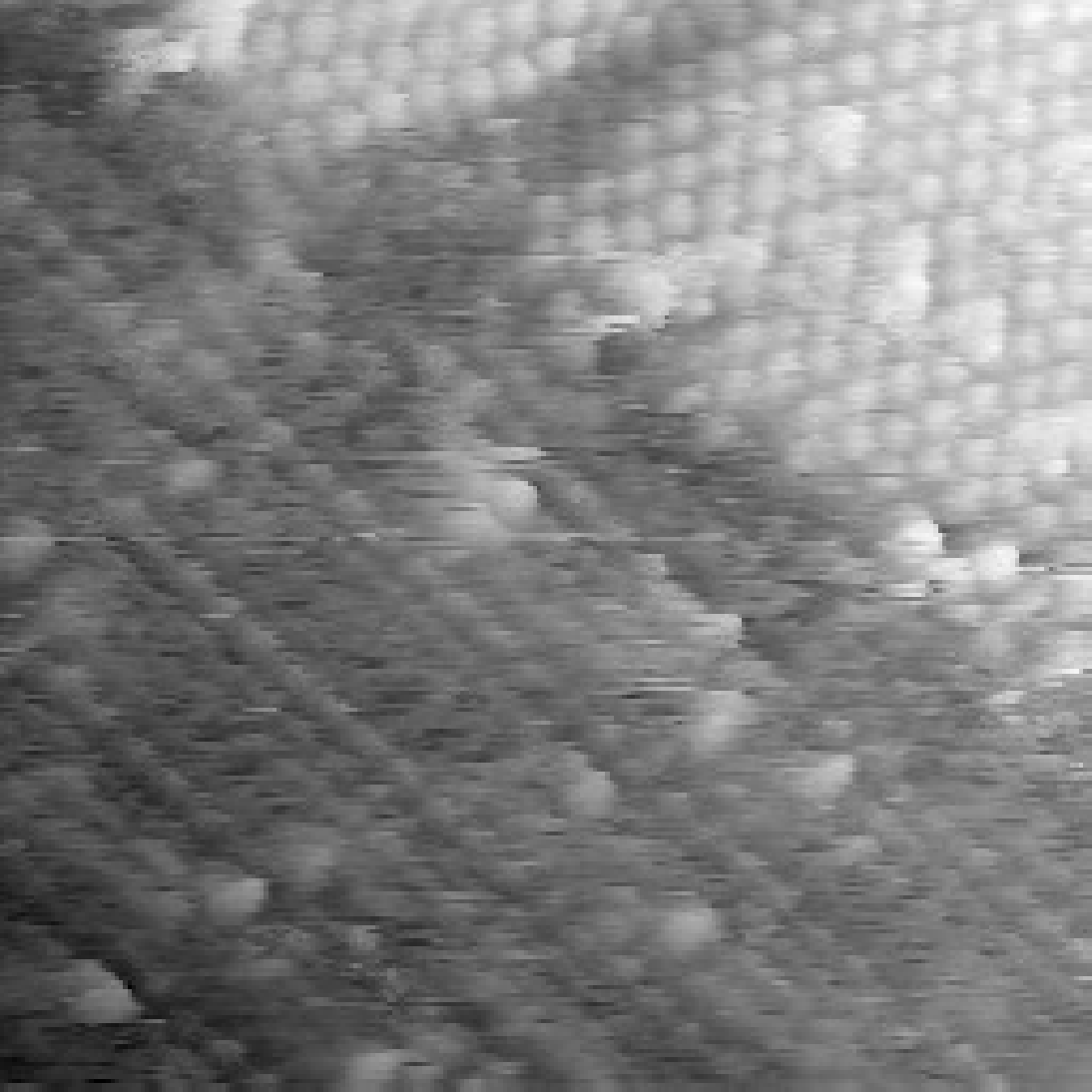} & 
  \includegraphics[width=0.2\textwidth]{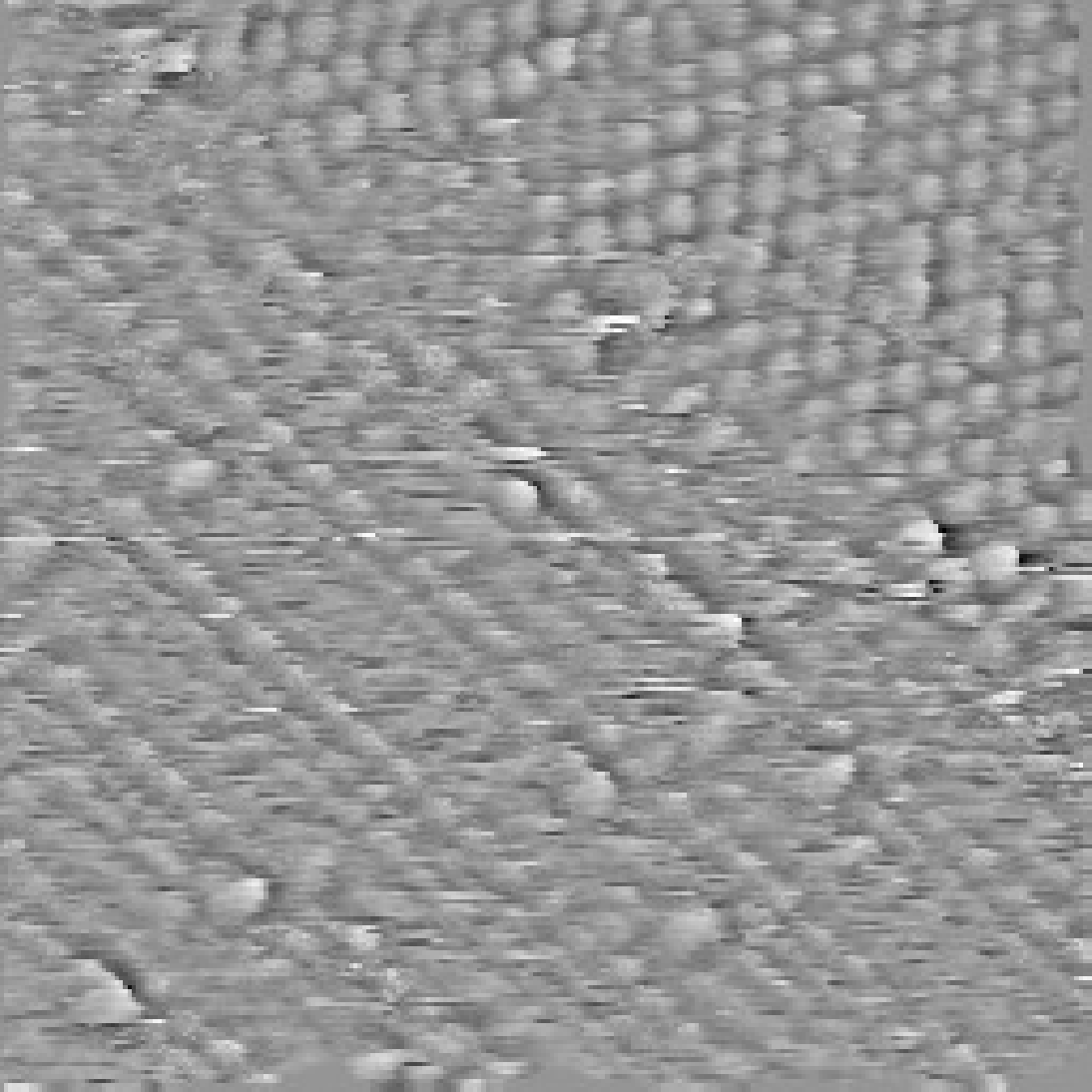}  &  
  \includegraphics[width=0.2\textwidth]{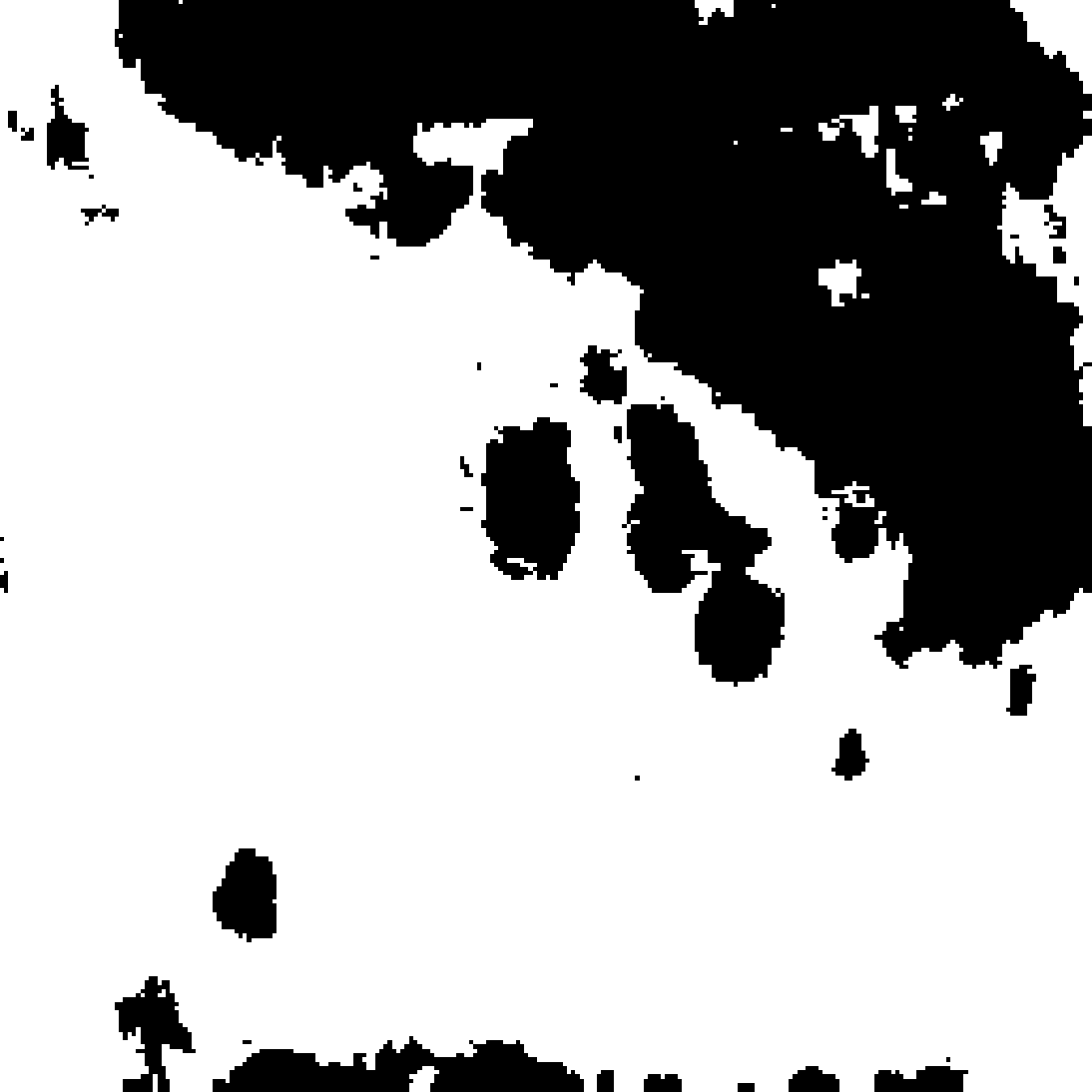} &
  \includegraphics[width=0.2\textwidth]{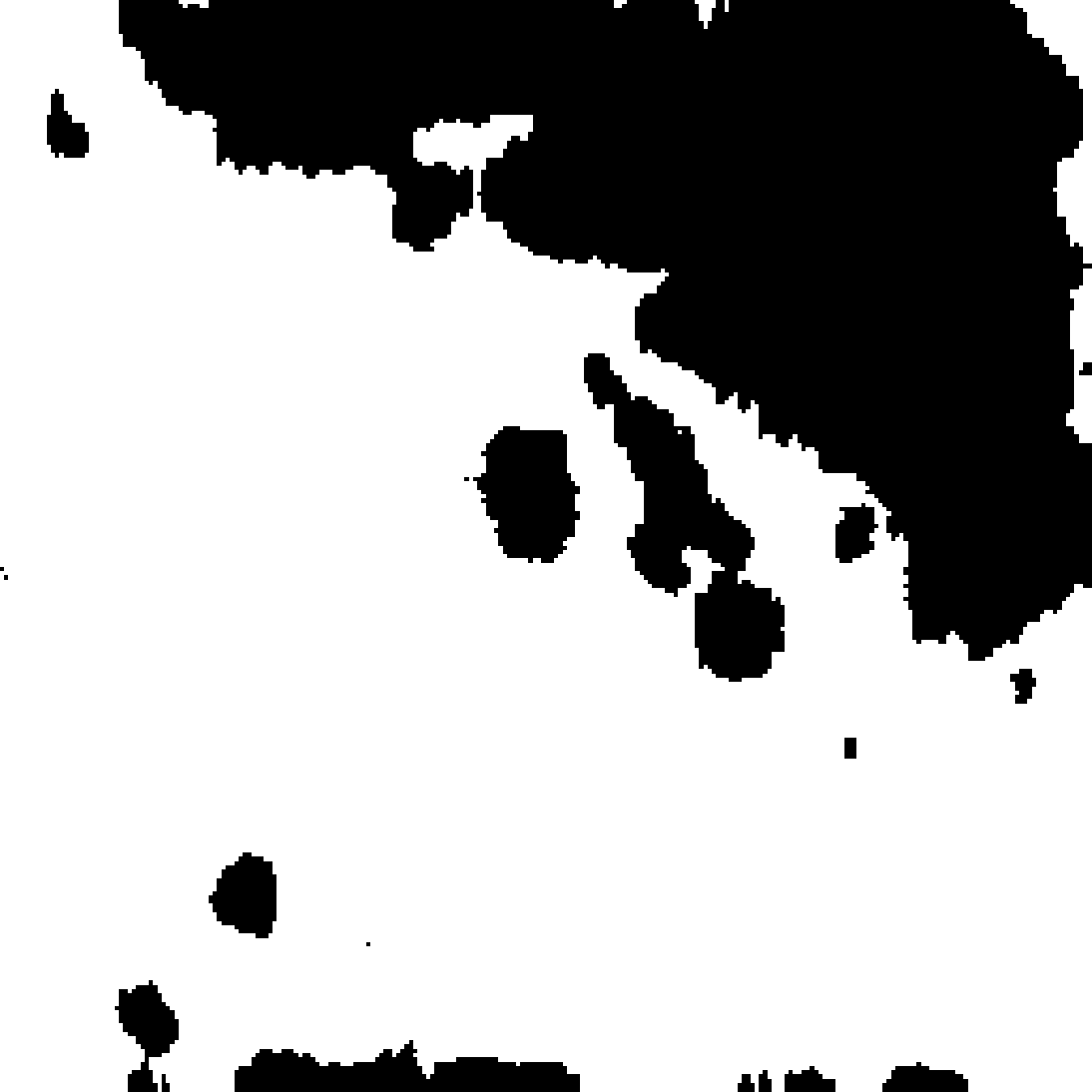} \\
\textbf{(d)}& 
  \includegraphics[width=0.2\textwidth]{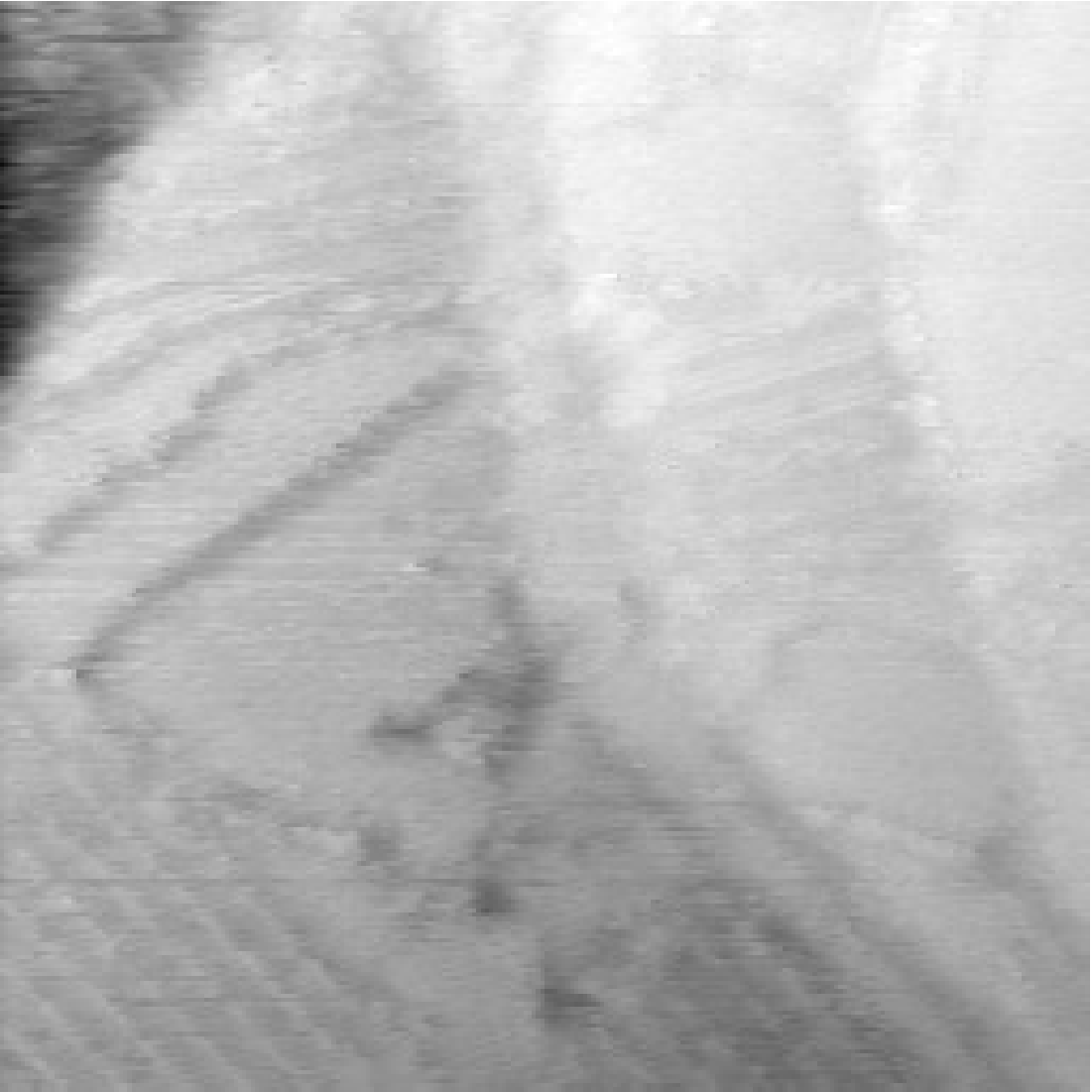} & 
  \includegraphics[width=0.2\textwidth]{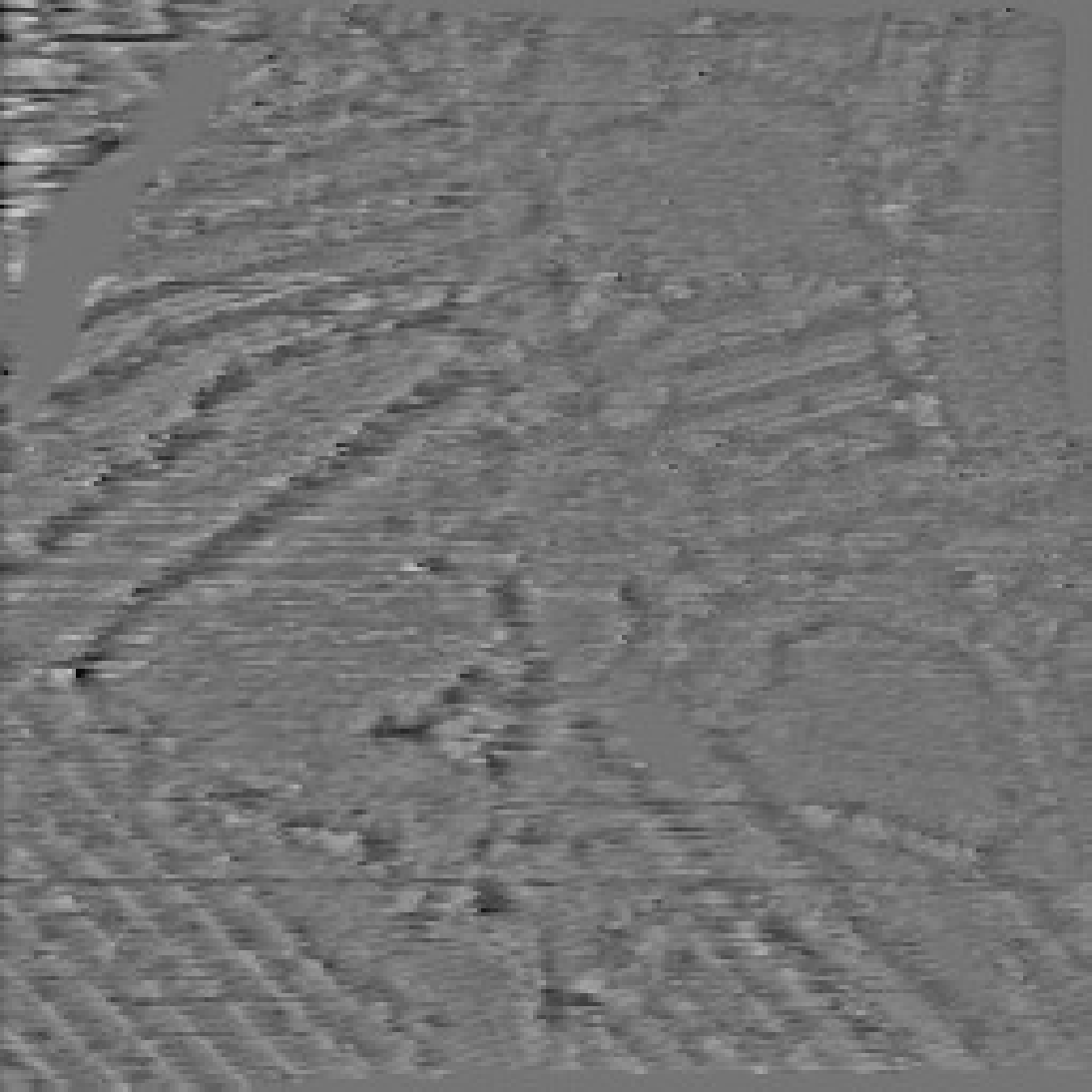}  &  
  \includegraphics[width=0.2\textwidth]{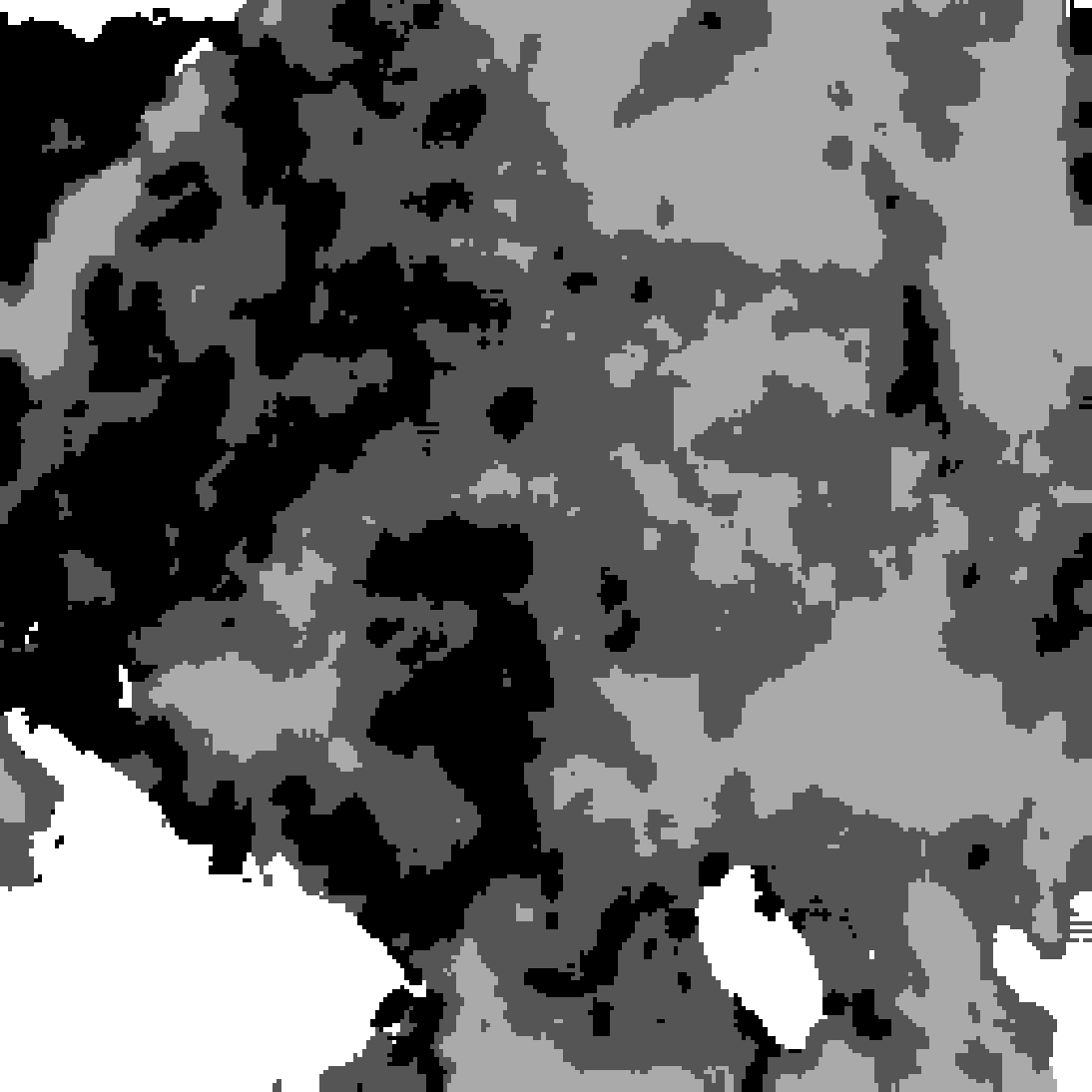} &
  \includegraphics[width=0.2\textwidth]{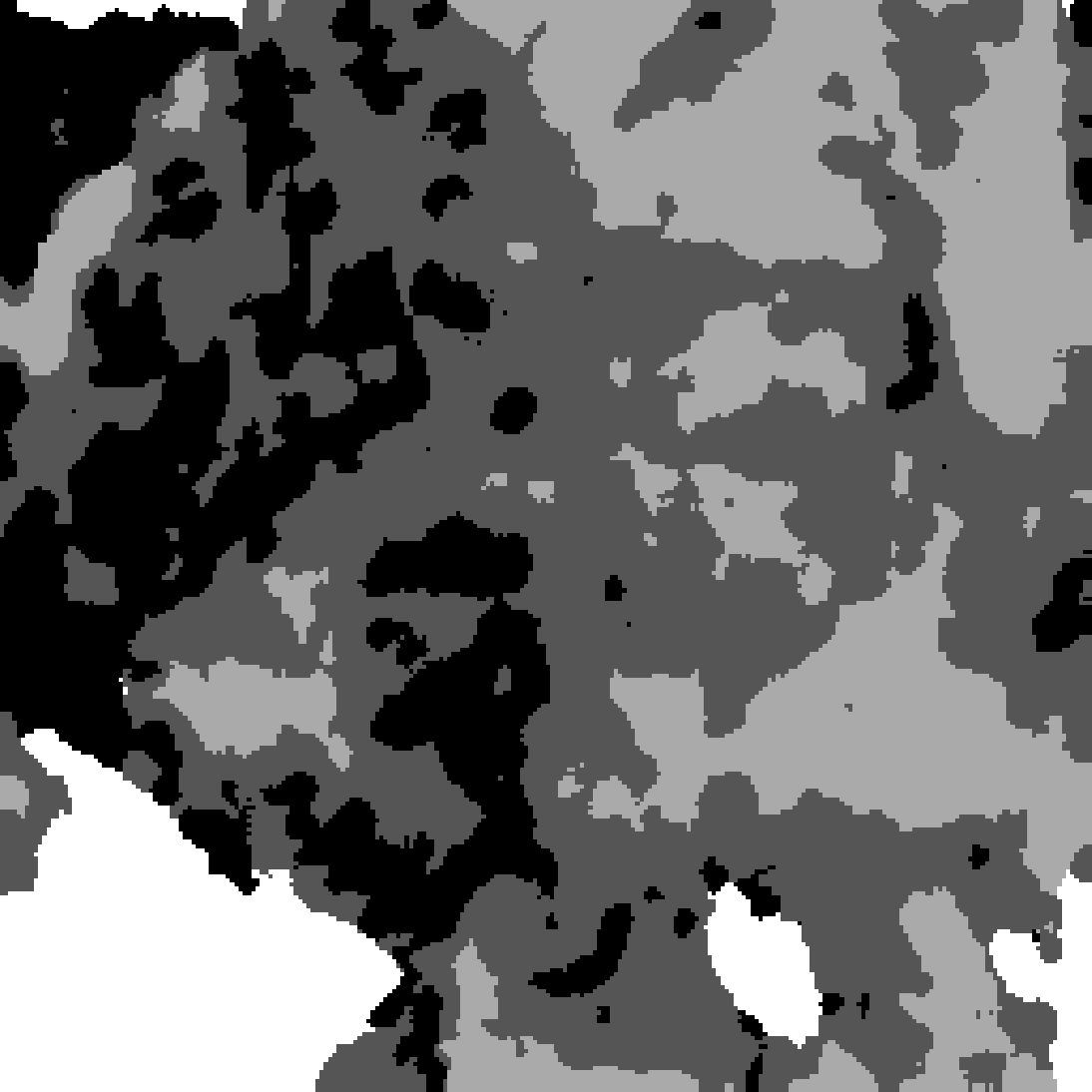} \\
\end{tabular}
\end{center}
\caption{Texture segmentation results. The parameters are 
\textbf{(a)} $\tau = 95.15$th percentile, $k=5$, $dt = 0.10$ 
\textbf{(b)} $\tau = 45$th percentile, $k=4$, $\tau =0.10$ 
\textbf{(c)} $\tau = 85$th percentile, $k=2$, $dt =0.05$ 
\textbf{(d)} $\tau = 72.50$th percentile, $k=4$, $dt = 0.05$. Raw scanning tunneling microscope images of cyanide on Au\{111\}, reproduced from \cite{guttentag2016hexagons} with permission. Images copyright American Chemical Society}
\label{fig:result4}
\end{figure*}

In order to segment the texture component of the image, we apply Algorithm \ref{alg:MECT} to obtain its ECT coefficients. Next we build the texture feature matrix accordingly to \eqref{eq:energymatrix}. Finally we apply a clustering algorithm to the energy matrix. The number of clusters is determined by the user. In our experiments, the clustering algorithms we use are $k$-means, already implemented in MATLAB, and multiclass MBO clustering \citep{garcia2014multiclass,Merkurjev}. Both methods use the cityblock metric to measure the similarity between data points of the energy matrix in order to determine the clusters. 

As an unsupervised method, $k$-means utilizes random initialization to determine the initial centroids. In order to obtain high-quality clusters, \cite{Arthur} devised a heuristic to ensure that every two initial centroids are dissimilar to each other. As a semi-supervised method, the multiclass MBO clustering  randomly selects 25\% of the labels determined by $k$-means as its initialization. Note that because of random initialization of the centroids, results may differ for every run. Hence, for each image, we run 10 replications of $k$-means and select the best result based on minimum within-cluster sums of point-to-centroid distances. The best result is only shown and is used as an  initialization to multiclass MBO clustering.

The multiclass MBO is a graph-based method, so it requires constructing the graph Laplacian matrix which is expensive both in computation and in memory. Instead of computing the matrix  exactly, we use Nystr\"om extension to approximate its eigen-decomposition \citep{fowlkes2004spectral}. The method requires from the user the number of data points to sample and the  number of eigenvectors to compute. In our experiments, we sample 300 data points and compute 30 eigenvectors. 

For all images, the multiclass MBO clustering is set with the following parameters: $\mu = 30$ (fidelity parameter) and $\eta = 10^{-7}$ (tolerance), following the notations in the work of \cite{garcia2014multiclass}. The thresholding step is performed after every three iterations of the diffusion step. The time step $dt$ is the only parameter that differs between the images.

Texture segmentation results obtained from $k$-means and multiclass MBO clustering are shown in Figures~\ref{fig:result3} and \ref{fig:result4}. We observe that both clustering  algorithms are able to identify most of the recognizable texture patterns of the images. Although both clustering results are similar, the results obtained from multiclass MBO  clustering tend to have smoother clusters and appear less noisy than the results obtained from $k$ means. In other words, the multiclass MBO results are cleaner than the $k$-means results. 

In Figure~\ref{fig:result3}a, we select $k=5$. Two of the clusters correspond to the material interfaces: one interface is a deep indentation and the other appears perforated. The three clusters correspond to the most periodic texture patterns: one that appears to have horizontal direction (the middle cluster), one that is slanted upward (the left  cluster), and one that is finer and is slanted downward (top cluster). 

In Figure~\ref{fig:result3}b, we select $k=2$. One cluster corresponds to the perforated lines while the other corresponds to the regions between the lines. However, some of the pixels appear to be misclassified. For example, pixels in the region between the perforated lines are classified as the same cluster. This result may be attributed to the noise 
or unexpected anomalies within the texture patterns. Since the clustering result from multiclass MBO appears to have less misclassified pixels than does $k$-means, it affirms that multiclass MBO is more resistant to noise, even with $25\%$ of the labels provided by $k$-means. 

In Figure~\ref{fig:result3}c, we select $k=2$ again. One cluster corresponds to the uniform texture pattern that has an upward, slanted direction while the other corresponds to the break dividing the uniform texture pattern. We observe that pixels near the break tend to be misclassified because of the break is relatively small compared to the 
uniform texture pattern.

In Figure~\ref{fig:result3}d, we select $k=3$: one corresponding to the perforated or indented edge, the other one corresponding to the slightly horizontal texture pattern (top and bottom right cluster), and the last one corresponding to the texture pattern slanted downward (bottom left cluster).\\ Again, as for the previous image discussed, pixels near the edge tend to be 
misclassified.

In Figure~\ref{fig:result4}a, the methods have difficulty clustering the texture patterns because the texture image itself has various texture patterns occurring at different sizes in various locations. For example, our results have a rectangular cluster located at the bottom of the image, but from the texture component, we observe that many types of texture patterns exist there. Furthermore, another cluster seems to correspond to the circular region in the texture pattern, but it also extends outside of it. Nevertheless, the other three clusters do correspond to identifiable texture patterns in the image. Two of the clusters correspond to texture patterns that slant upward, but one has finer texture while the other has sharper edges. The third cluster identifies with the horizontal texture pattern (top and bottom cluster). 

In Figure~\ref{fig:result4}b, we select $k=4$. One cluster corresponds to the smoothest regions of the texture component. Another corresponds to the texture pattern slanted 
downward and with rough edges (bottom left cluster). The third cluster corresponds to the finer texture pattern slanted downward (top right, bottom left, and along the right edge 
of the elliptical smooth region). However, the last cluster seems to correspond to the miscellaneous region that does not have any apparent pattern. 

In Figure~\ref{fig:result4}c, we select $k=2$, one cluster corresponding to the scaly texture pattern in the top right and the other one corresponding to the rough texture pattern 
that have some lines slanted downward. We observe that some pixels outside of the top right region belong to the same cluster, and most of them do look similar to the top 
right texture pattern as they look like scales. 

In Figure~\ref{fig:result4}d, we select $k=4$. One cluster corresponds to the texture slanted downward and it is located at the bottom left and bottom middle of the image. Another 
cluster corresponds to the smooth regions of the image, which are located at the top left corner and most of the right side of the image. The other cluster seems to correspond to 
the rough edges, which are located mostly in the left of the image. The last cluster seems to correspond to miscellaneous patterns like in Figure~\ref{fig:result4}b. 

\section{Conclusions}\label{sec:conclusion}
In this paper, we proposed a framework to segment STM images, combining variational methods and a clustering algorithm based on features extracted by the empirical wavelet transform. The expected information of microscopy images led us to first apply a cartoon+texture decomposition and then run a modified version of the multiphase CV model on the cartoon part, and a clustering of features extracted by the empirical curvelet transform on the texture part. The results in Section \ref{sec:experiments} demonstrate the proficiency of this framework to analyze STM images. \cite{guttentag2016hexagons} have already used the proposed approach to characterize patterns of cyanide molecules on Au\{111\}, complementing the results in another related work \citep{guttentag2016surface}.

There are several directions to investigate in order to improve the proposed framework. One direction is employing directional filters \citep{buades2016directional2, buades2016directional} instead of iso\-tropic filters $L_\sigma$ in the cartoon+texture  decomposition, which are designed to separate noise and  micro-texture better in the proximity of edges. As such, in the proposed local MCV model, the Gaussian filter could be upgraded to a nonlocal filter, incorporating direction to better characterize  weak edges. To gain robustness against low contrast and impulse noise, one could work with the $L^1$ fidelity term, but since the latter is non-differentiable at the origin, one would have to use  convex optimization algorithms like primal-dual methods. Finally, in order to overcome artifacts (\textit{e.g.}, scars, occluions, scratches), one could incorporate an indicator function as done by \cite{zosso2017image}. As for the segmentation of the texture component, the use of other features than the local energy of the curvelet coefficient, such as the co-ocurrence matrices \citep{haralick1973textural} constructed from the empirical curvelets, is likely to lead to more adapted clustering results. 

Overall, the proposed framework has produced remarkable segmentation results of STM images using variants of the state-of-the-art image processing algorithms. One could modify and apply this framework to other kinds of images, thus leading to more interesting contributions and applications in other scientific areas beyond nanoscience.

\begin{acknowledgements}This work was supported by the W.M. Keck Foundation Center for Leveraging Sparsity. The self-assembly and STM imaging were supported by the U.S. Department of Energy (Grant \#DE-SC-1037004).
L. Torres Mandiola was partially supported by NSF DMS-1312361.  A. Ciomaga, K. Bui, J. Fauman, and D. Kes were partially supported by NSF DMS-1045536. A. Bertozzi was supported by Simons Math + X Investigator Award number 510776. 
\end{acknowledgements}

\bibliographystyle{spmpsci}      
\bibliography{mybib}   

\end{document}